\documentclass[11pt]{article}

\usepackage[preprint]{acl}

\usepackage{times}
\usepackage{latexsym}
\usepackage{amsfonts,amsmath,amssymb,amsthm}
\usepackage{booktabs}

\usepackage[T1]{fontenc}

\usepackage[utf8]{inputenc}

\usepackage{microtype}

\usepackage{inconsolata}

\usepackage{graphicx}

\usepackage{stfloats}

\title{Representation of syntax  in LLMs through the lens of linear distance and similarity-aware entropy
}

\author{
  \textbf{Juan Pablo Vigneaux\textsuperscript{1,}\thanks{%
    Corresponding author: \href{mailto:jpvigneaux@northwestern.edu}{jpvigneaux@northwestern.edu}\\
    Code and data: \href{https://github.com/jpvigneaux/structural-probes-labelwise-analysis}{github.com/jpvigneaux/structural-probes-labelwise-analysis}
}},
  \textbf{Mary Kennedy\textsuperscript{2}},
  \textbf{Khalil Iskarous\textsuperscript{2}},
  \textbf{Robert Frank\textsuperscript{3}},
  \textbf{Matilde Marcolli\textsuperscript{4}}
\\
\\
  \textsuperscript{1}Northwestern University,
  \textsuperscript{2}University of Southern California,\\
  \textsuperscript{3}Yale University,
  \textsuperscript{4}California Institute of Technology
}

\renewcommand{\thefootnote}{\fnsymbol{footnote}}

\begin{document}
\maketitle

\renewcommand{\thefootnote}{\arabic{footnote}}
\setcounter{footnote}{0}
\begin{abstract}
Structural probes were introduced by Hewitt and Manning to reconstruct syntactic trees from a neural language model's latent representations. They are evaluated by calculating the proportion of syntactic tree edges correctly reconstructed over an annotated corpus (as measured by undirected unlabeled attachment score). Here, we disaggregate this measure, considering undirected attachment score by label (UASL), which assesses the reconstruction accuracy of each syntactic relation separately, establishing important differences among relations that overlap linguistic distinctions. Moreover, we identify two factors that predict most of UASL's variability across relations: (i) the mean and dispersion of the linear distance (on a log scale) between the related words, and (ii) the diversity (similarity-aware entropy) of the syntactic relation's head. These results, which hold across a range of model sizes and architectures, shed light on the degree of abstraction of the representation of syntax in language models and the dependence of such representation on geometric properties of the embedding space.

\end{abstract}

\section{Introduction}

\begin{figure}[h!]
\centering
\includegraphics[width=\columnwidth]{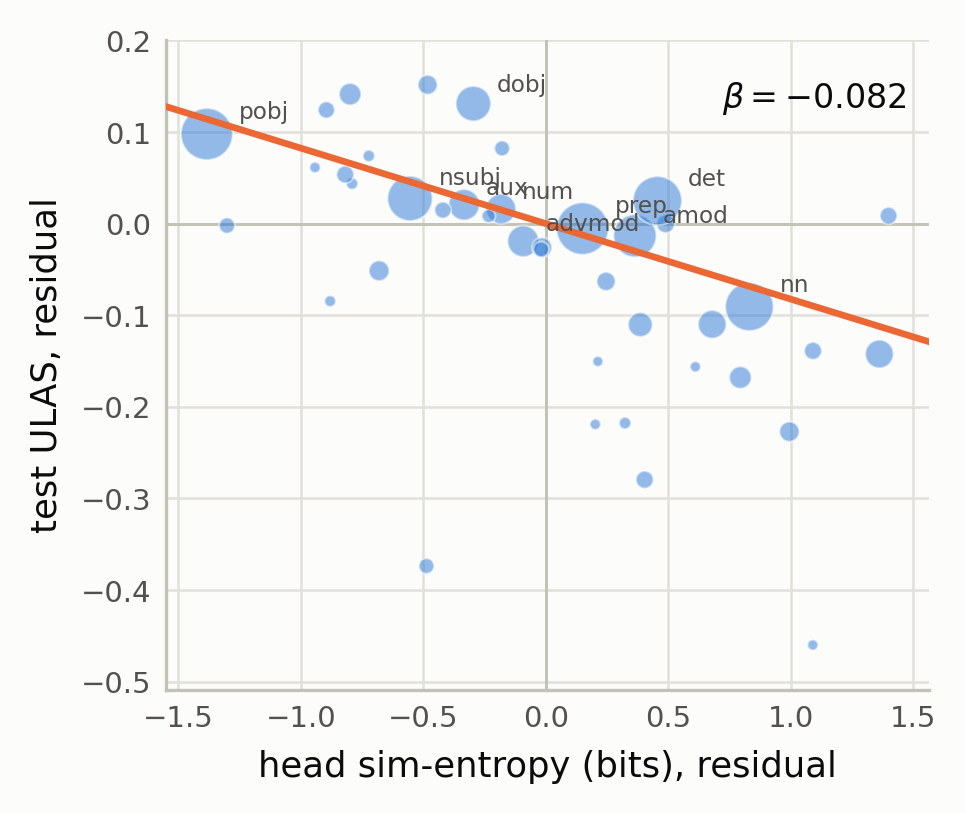}
\caption{Held-out UASL against the similarity-corrected entropy of the relation's head $H_{\rm head}^{\rm sim}$, for the optimal structural probe trained on BERT-base embeddings of the PTB. %
Both axes are residuals with respect to the other two predictors %
in the weighted linear regression of \S\ref{EntropySec}.   %
Each point is one relation, with area proportional to the number of edges used to measure its UASL; %
 the ten largest are labelled. }
\label{fig:head-sim-entropy}
\end{figure}

Probing vector representations has been widely used as a means for extracting insights into the inner workings of neural language models \cite{alain-bengio-2017-probe, conneau-etal-2018-cram}. Since the seminal work of \citet{hewitt2019structural} found that dependency trees could be recovered with probes with a reasonable degree of success, there has been a proliferation of syntactic probing methodologies \cite{chi-etal-2020-finding,davis-etal-2022-probing,tucker-etal-2022-syntax,eisape-etal-2022-probing,arps-etal-2022-probing, Diego-Simon2024a}. 
We would like to leverage these results to identify to what extent and through which mechanisms the algebraic structure of syntax is encoded and processed by LLMs in the geometry of their latent spaces. This includes not just the question of whether syntactic relations and hierarchical structures are detectable, but more precisely whether linguistic differences among dependency relations correspond to quantitative differences in performance and whether these differences can be traced back to distinct computational mechanisms.

The work we present in this paper represents the first steps in such an investigation.  It shows that individual syntactic relations behave differently with respect to probe performance, quantifying in precise mathematical terms certain significant ``limitations of abstraction" in the representation of syntactic dependencies. 

Firstly, we show that the accuracy at reconstructing a given dependency relation %
varies widely among relations. In \S\ref{UUASperfsec} we study this variability across multiple relations and looking at representations at different model depths. We then consider possible underlying explanatory variables derived from linguistic theory: nature of the syntactic-role, frequency of the relation, linear distance between head and dependent, and similarity-aware entropy of the head or dependent. %

Secondly, by performing a more detailed analysis  of the UASL dependence on distance, based both on a log-linear decay model (\S\ref{sec:logdistsec}) and on a model-free comparison between the UASL-vs-distance curves (\S\ref{sec:distances}), we find that \emph{dependency relations with similar properties from the perspective of syntactic theory exhibit similar behaviors.} Therefore, linguistic distinctions provide useful cues for the identification of likely distinct mechanisms that underlie the model's processing of these relations. 

Thirdly, we uncover another variable that modulates UASL performance across relations, largely independent of the linear distance descriptors: \emph{the similarity-aware entropy of a relation's heads}. The similarity in question is derived from neural word embeddings; we see these embeddings as an instance of a geometrically-structured semantic space  \cite{Shepard2024,Piantadosi2024,Grindrod2025}, partly shared between models \cite{huh2024position,jha2026harnessing} and somewhat comparable with human conceptualization---see, e.g. \cite{reif2019visualizing, gari-soler-apidianaki-2021-lets, gupta2022deep, Trott2023}.

Higher values of the similarity-aware entropy of a relation head  (i.e. greater diversity in heads' embeddings) hinder probe performance. We conjecture that this measures difficulties that the attention heads encounter in encoding, geometrically, the connections between dependents and heads and hence the relation as an abstract construct. It is suggestive that  the majority of the ``syntactic'' attention heads identified by \citet{clark-etal-2019-bert} are such that the dependent attends predominantly to its head; this makes us think that the problem of reconstructing the relation might be recast by the models as identifying the head given the dependent, which is more complex if the attention mechanism has to point into independent or even conflicting directions. This is the kind of understanding that we would like to motivate with this study.

Our results are stable across a range of models, which differ both in parameter counts and architecture: BERT-base, DeBERTa-v3-base, ModernBERT-base, GPT-2-base, and GPT-J-6B. This points to some underlying structure present in the language itself, or at least to the way transformers can capture that structure.

\section{Related work}
\label{sec:relatedwork}

\textbf{Probing.} Structural probes were introduced in \citet{hewitt2019structural} and  featured in \citet{Manning2020}. They have been the subject of an extensive literature; in particular,  \citet{Someya2025} established a bottom-up derivation of syntactic trees in BERT and  \citet{Simon2025} established a general dependency of UUAS on distance (but without differentiating the effect among syntactic relations).

\paragraph{Similarity-aware entropy.}  This function first appeared in ecology \citep{Leinster2012, Leinster2021a} as a way of measuring diversity. It shows up in the literature under various names, including  ``kernelized 2-complexity'' \cite{aishwarya2026}  and ``geometry-aware Shannon entropy'' \cite{posada2020}.  In NLP, this function has been used to quantify uncertainty in machine translation \cite{Cheng2024}.

\section{Background}

\subsection{Transformer embeddings}
\label{sec:embeddings}

\paragraph{Models.} In this paper we work with several transformer-based language models \cite{Vaswani2017}.  The main text utilizes BERT-base \cite{devlin2019bert}, which almost matches BERT-large at syntactic tree reconstruction by structural probes \cite{hewitt2019structural} at a fraction of the cost. BERT is one of the most impactful models in NLP research \cite{Rogers2020}. Appendices \ref{app:other-models}--\ref{app:regression} repeat every analysis on four other models: DeBERTa-v3-base \cite{he2023debertav3improvingdebertausing}, ModernBERT-base \cite{warner-etal-2025-smarter}, GPT-2-base \cite{radford-etal-2019-gpt2} and GPT-J-6B \cite{gpt-j}, which differ from BERT on various architectural choices. Moreover, in Appendix \ref{sec:shuffle-model}  we repeat all our analysis on RoBERTa-Shuffle-n1, a variant of RoBERTa \cite{liu-etal-2019-roberta} pre-trained on permuted sentences \cite{Sinha2021}, to test how much of our results can be retrieved for distributional information alone (spoiler: none). The model architectures are summarized in Table \ref{tab:models}.

\paragraph{Residual stream and checkpoints.} A transformer maintains, for each token, a running vector---the \emph{residual stream}---that every attention and MLP module reads from and adds back into \cite{Elhage2021, Ferrando2024}. At any stage of the computation, the residual stream can be seen as a latent contextualized representation of the token. With exceptions, we train structural probes on the representations that arise after each modification of the residual stream (from raw token embeddings to the final representation used for prediction) and we refer to these extraction points as \emph{residual stream checkpoints}. If the $L$ transformer blocks of a model apply attention and MLP sequentially, they contribute with $2L$ residual stream checkpoints, but if they apply attention and MLP in parallel they contribute only with $L$. Appendix \ref{app:checkpoints} introduces the formulas that characterize the checkpoints and Table \ref{tab:models} gives the number of checkpoints we introduced for each model. For BERT-base, we introduce 26 residual stream checkpoints indexed from $0$, of which the odd ones $3, 5, \ldots, 25$ (post-transformer block) are the twelve ``layers'' of \citet{hewitt2019structural}.

\subsection{Structural probes}
\label{sec:probes}

\textbf{Data.} We utilize dependency trees derived from the WSJ section of the
Penn Treebank \cite{marcus-etal-1993-building} according to
\citet{de-marneffe-etal-2006-generating}'s methodology for generation from
phrase structures, with the standard parsing split of sections 02--21 for
training, 22 for development and 23 for test \citep{collins-1997-three}.
 Each sentence yields a tree whose vertices are PTB tokens---mainly words and symbols---and whose edges are labelled by dependency relations; forty-two relations occur in the train and development sets. Table \ref{tab:edge-counts} details label counts for each split. 

\paragraph{Probes.} A structural probe reconstructs the undirected, unlabelled version of the dependency tree from the model's latent representations. It consists of a linear map from the model's latent space into a vector space of dimension $k$ (we restrict ourselves to $k=64$, following \citet{hewitt2019structural}), fitted so that the squared Euclidean distance between the images of two words approximates the number of edges separating them in the tree; the predicted tree is then the minimum spanning tree of the projected points. We fit one probe per residual stream checkpoint. Remark that labels do not enter into the training objective. We use the implementation of \citet{hewitt2019structural}, with modifications that extend it to tokenizers other than BERT's.%

\paragraph{Alignment.} It is worth remarking that, to generate the embeddings, we feed the model natural looking sentences (no PTB-specific symbols, no spurious spaces). Hence we need to align model tokens with PTB tokens. For this we generalize \citet{Hewitt2021a}'s method: each model token is assigned the natural sentence's characters that it covers, using the tokenizer's offsets, and those characters are mapped to PTB tokens via character-level Levenshtein alignment. 
 A PTB token's representation is the weighted average of the model tokens' vectors  assigned to it this way. 

\paragraph{Metrics.} A probe is ordinarily scored by the proportion of gold edges its reconstructed trees recover, the undirected unlabelled attachment score (UUAS), computed over a section of the corpus. Our main object of study is its restriction to a single relation, which we call the \emph{undirected  attachment score by label}: ${\rm UASL}(\lambda)$ is the proportion of the edges labelled $\lambda$ that the probe recovers. Since the probe is not trained on labels, UASL measures how legible a relation is in a representation, not the accuracy of a labelled parser. Both scores are defined in Appendix \ref{app:metrics}.

\section{UASL performance by dependencies}\label{UUASperfsec}

\begin{figure}  \includegraphics[width=1.07\columnwidth]{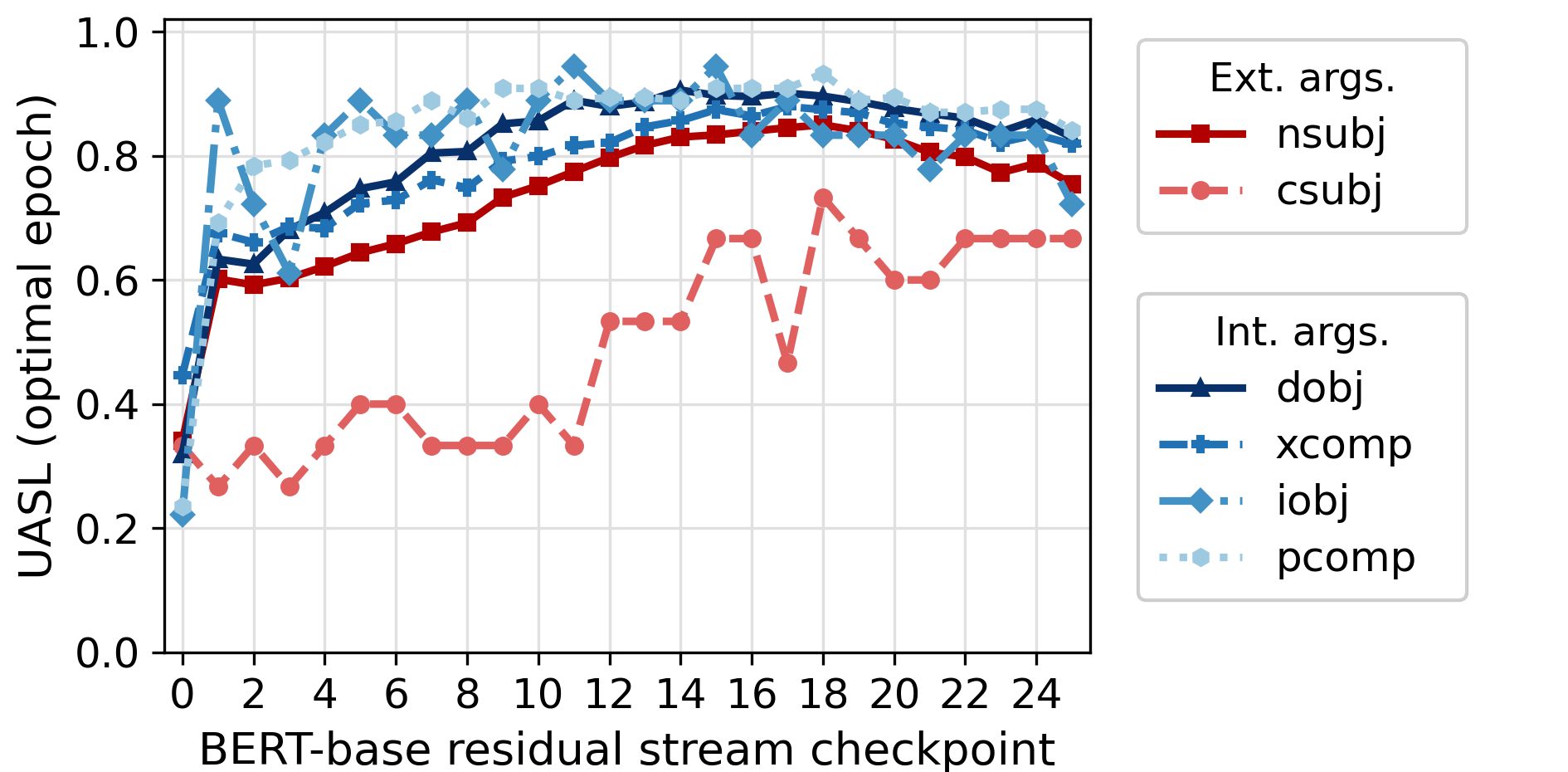}
  \caption{UASL curves for verbs' external and internal arguments. Edge counts per relation and split are in Table \ref{tab:edge-counts}.}
  \label{fig:dependencies1}
\end{figure}

\begin{figure}
\includegraphics[width=1.07\columnwidth]{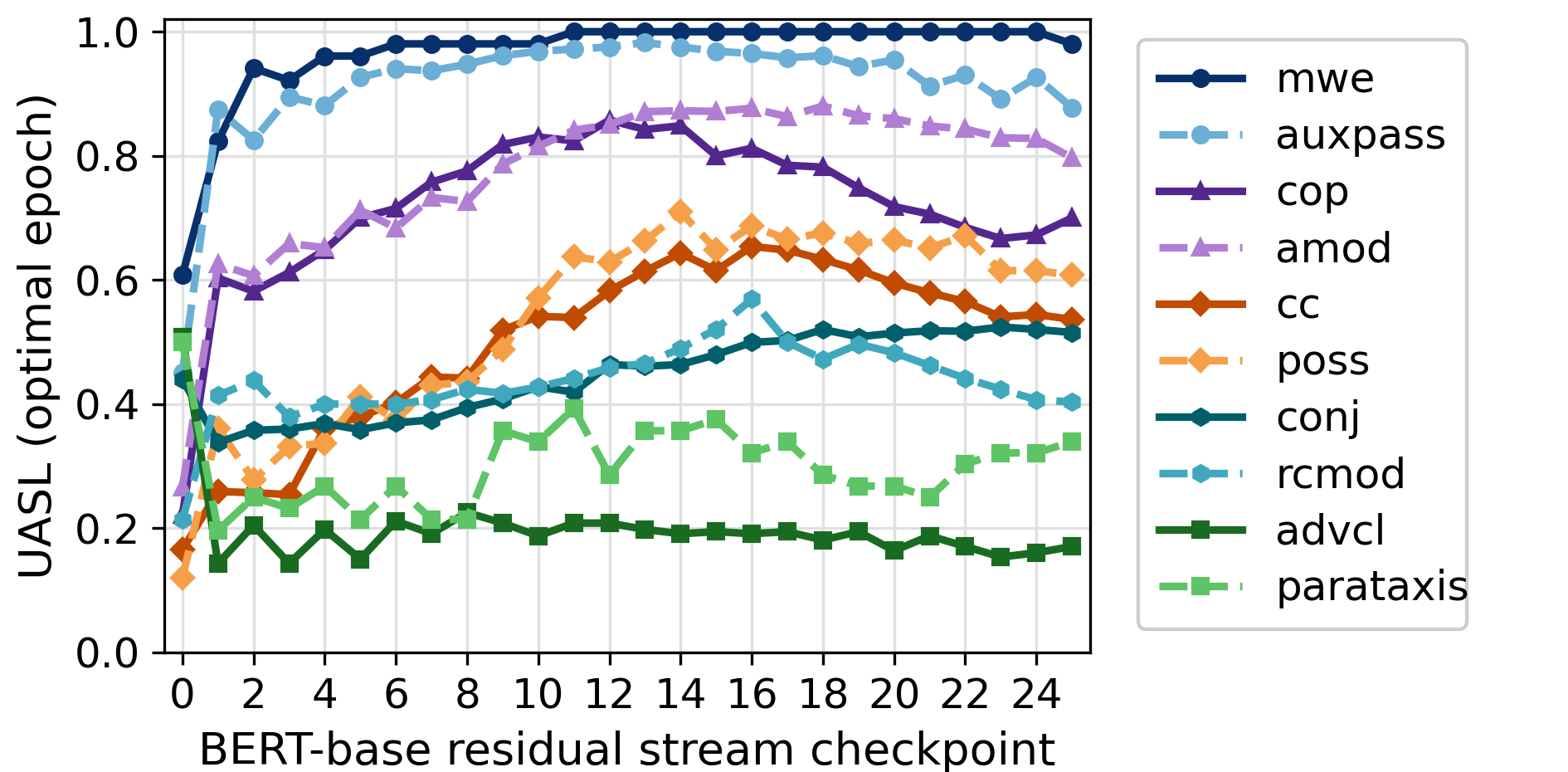}
  \caption{UASL curves of dependency relation that represent  top, intermediate, and low performance, with intermediate performance further divided among high, middle  and low range. Edge counts per relation and split are in Table \ref{tab:edge-counts}.}
  \label{fig:dependencies2}
\end{figure}

As a first step, we look at the representative dev set probe performance by dependency label through a simple qualitative analysis of different linguistic properties and general trends of the syntactic dependencies. We outline candidate theoretical explanations for the UASL performance that we quantitatively confirm in \S\ref{DistSec}  and \S\ref{EntropySec}.

\paragraph{Thematic dependencies.}
Of the 41 dependencies analyzed, some have direct relevance to the syntax of argumentation (see Figure \ref{fig:dependencies1}). Complements, also known as internal arguments \cite{compl-int-argu}, have a tendency to pattern similarly to each other. For example, {\em dobj} and {\em pobj} have nearly identical patterns and not dissimilar performances to {\em xcomp, pcomp,} and {\em iobj,} the latter of which has an even better performance for reasons we will discuss in \S\ref{EntropySec}. In this, we observe internal arguments perform better than their external counterparts.\footnote{External subjects are those taken by a vP or VoiceP head \cite{Kratzer1996, hale-argustructure}. We recognize that not all subjects (e.g. subjects of unaccusatives, see \citet{harley-vp-argu}) are external arguments. We treat {\em nsubj} as external due to it being prototypically external.}

\paragraph{Performance.} As seen in Figure \ref{fig:dependencies2}, top probe performance is held by a largely closed class of words that tend to have a set phraseology, such as {\em mwe, auxpass,} and {\em expl} (not depicted). Intermediate performance can vary from relations like copulas and adjectival modifiers in the higher range to the dependency relations between nouns and the verbs of their relative clause ({\em rcmod}) and between coordinated words ({\em conj}) in the low-intermediate range. The relation {\em conj} should not  be confused with {\em cc}, (between a coordinated head and its conjunction), which occurs in the intermediate-mid  range.

The bottom-performing relations are demonstrated with {\em advcl} (e.g. ``the cookies \textbf{burned} because I \underline{forgot} them in the oven'') and {\em parataxis} (e.g. ``I \textbf{learned} my lesson: don't \underline{forget} to set the timer''). Both denote syntactically complex sentences and serve to link two clauses, which may help to explain some of their low performance. {\em Parataxis} occurs at a similar rate to {\em expl} and {\em advcl} has a comparable presence to {\em auxpass} or {\em rcmod}, and yet both perform far lower. {\em Advcl} performs even worse than {\em parataxis} despite having a greater presence in the corpus.

Together, we may observe that probe performance is not strictly dictated by a dependency relation's frequency in the corpus. In \S\ref{DistSec} and \S\ref{EntropySec}, we explain the additional confounding linguistic factors that interact to predict probe performance.

\paragraph{Length and diversity.}
This simple qualitative linguistic analysis of the
different dependencies points to two main factors 
that drive the different UASL performance: the distance between the two terms in the relation, and a measure of diversity of the terms involved in the relation, which we discuss in \S\ref{DistSec} and \S\ref{EntropySec}, respectively. We also find the \textit{ordered} distance to have a large impact, which we confirm below with an analysis of the effect of permutations.

\section{Distance and UASL performance} 
\label{DistSec}

\begin{figure}
\centering
\includegraphics[width=\columnwidth]{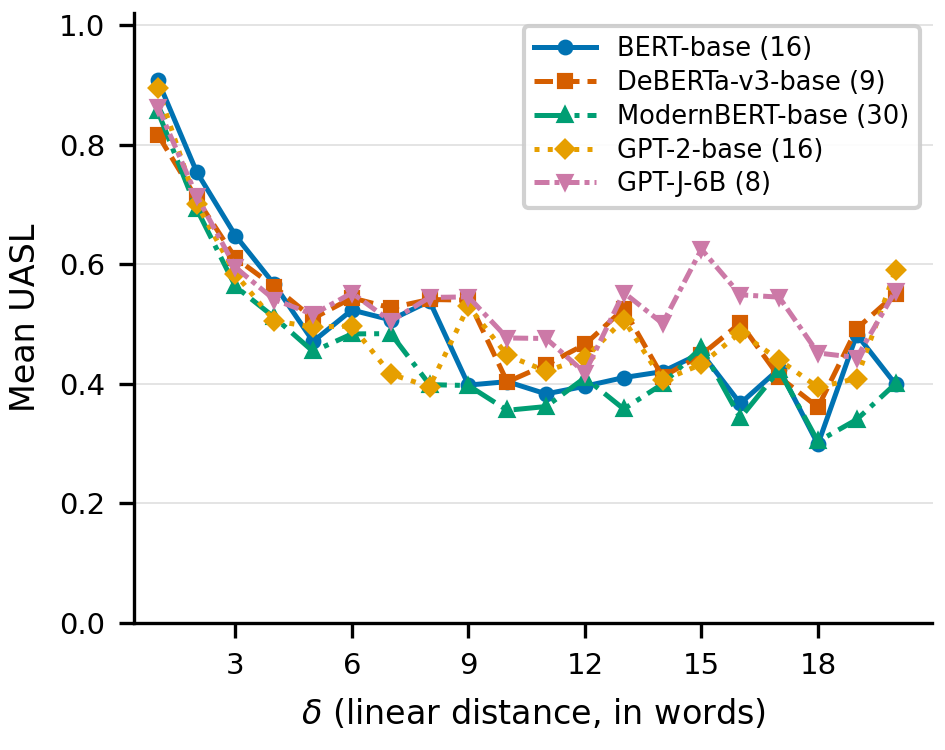}
\caption{Mean UASL against linear distance, for each model at
its own optimal residual stream checkpoint, given in parentheses. The average
is taken over the dependency relations with at least five gold edges at that
distance, and the same decay appears in all five models: steep over the first
few words, then close to flat. Figure \ref{fig:performance-as-a-function-of-distance}
gives the BERT-base curve at every post-block checkpoint.}
\label{fig:decay-across-models}
\end{figure}

In the qualitative analysis of \S\ref{UUASperfsec}, we identified that both the distribution of linear distances associated with a dependency relation  and the variability or diversity of its arguments modulate its UASL. In this section, we study the UASL dependency on linear distance across different relations in more detail; we will see that dependency relations group naturally in this respect according to syntactically motivated distinctions.  \S\ref{EntropySec} below incorporates the diversity of the arguments into the picture. 

While the results in this section confirm the findings of \citet{Simon2025},  who find that increased linear distance decreases a probe's accuracy on that word-pair, they also refine them by disaggregating the analysis over dependency relations.

\subsection{Log-distance as predictor}
\label{sec:logdistsec}

While on average, the effect of linear distance on performance is significant, distance is not necessarily a good predictor for individual dependencies. 

Based on the decay pattern in Figure \ref{fig:decay-across-models}, we fit a log-linear decay model across dependency relations and residual stream checkpoints, using a least-square regression:
\begin{equation}\label{regrUASL}
    \widehat{\mathrm{UASL}}_\delta(r, k) = a_{r,k} + b_{r,k} \, \ln \delta +  \varepsilon \, . 
\end{equation}
Let $S_\delta(r)$ be the gold edges labeled by a relation $r$ and associated with a linear distance $\delta$; then  $\mathrm{UASL}_\delta(r,k)$ is the proportion of elements of $S_\delta(r)$ recovered by the structural probe at residual stream checkpoint $k$ . We restrict the regression to the $\delta$s for which $S_\delta(r)$ on the PTB test set is at least 5. 

\begin{figure}[b]
  \includegraphics[width=\columnwidth]{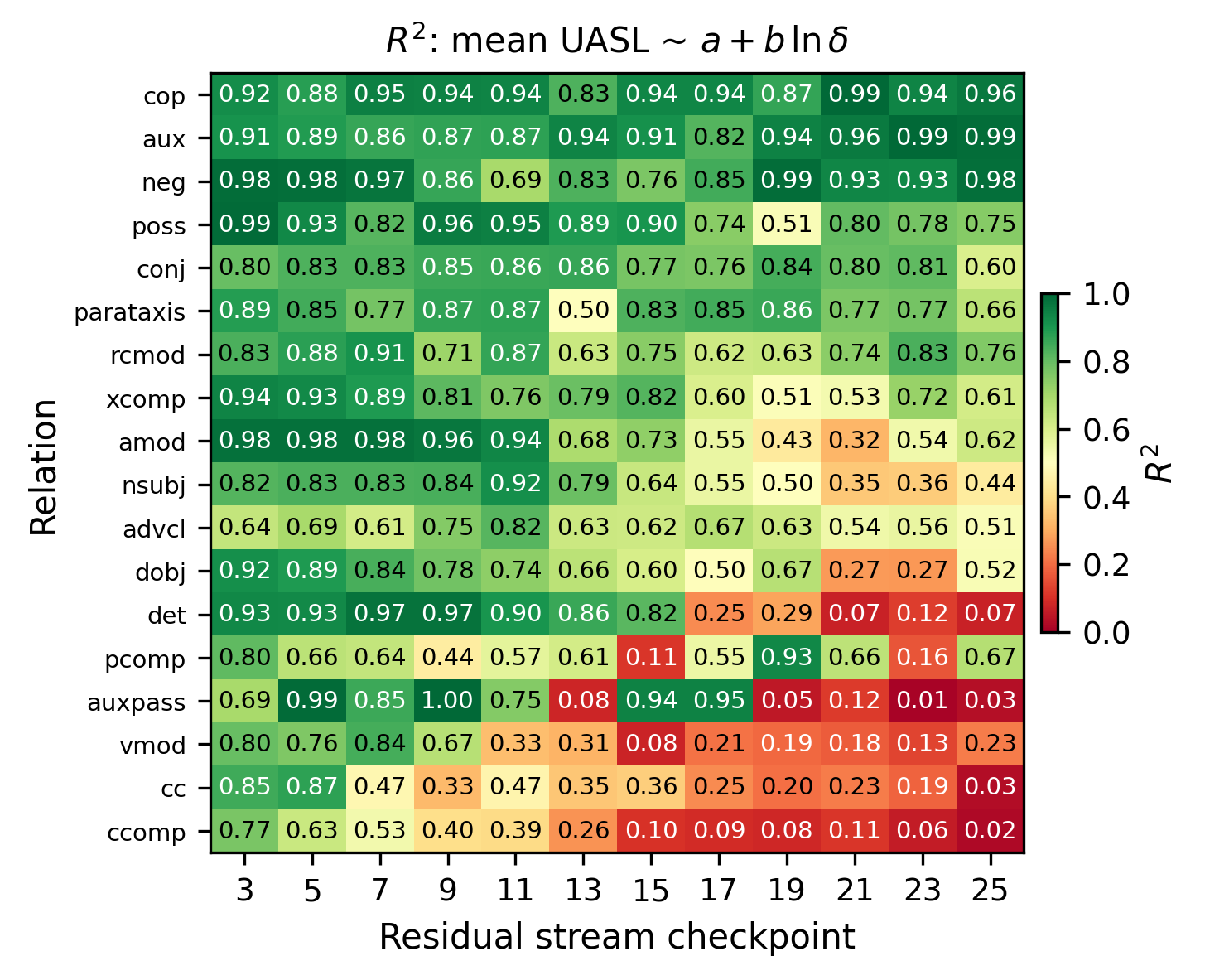}
  \caption{Fitness of the log-linear decay model across selected dependency relations, with the residual stream evaluated after each full transformer block. Relations taken from Figures \ref{fig:dependencies1} and \ref{fig:dependencies2}, and \S\ref{sec:logdistsec}; \textit{csubj}, \textit{iobj} and \textit{mwe} omitted because they vary too little in linear distance. Relations are ordered by mean $R^{2}$ across residual stream checkpoints.  \textit{Auxpass} is fitted on only three points. See Figure \ref{fig:ulas-vs-log-distance} in the Appendix for elaboration.}
  \label{fig:R2-log-distance-model}
\end{figure}

In Figures \ref{fig:R2-log-distance-model} and  \ref{fig:R2-all-relations}, we display the coefficient of determination $R^2$ of the regression for each pair $(r,k)$. Ten relations had fewer than three observed distance values and  were excluded from analysis. Among the best matches, we find dependencies like \textit{neg}, \textit{cop}, \textit{aux} (occurring in the top and high-intermediate performance range in \S\ref{UUASperfsec}), which exhibit near-perfect log-linear decay at all residual stream checkpoints. 
At the opposite end, near the bottom, we find relations
such as \textit{vmod}, \textit{ccomp}, \textit{cc} (occurring in the middle-intermediate performance range in \S\ref{UUASperfsec}) that deviate very significantly from a log-linear decay. Meanwhile, \S\ref{UUASperfsec}'s worst-performer, {\em advcl}, appears here in the middle. 

\subsection{Clustering of dependencies by impact of distance on UASL}\label{sec:distances}

We want to group the 42 dependency relations that we extracted from the PTB by similar dependence of their UASL on distance, without assuming log-linearity.  Here, we introduce a measure $d_{\rm UASL}$ that compares UASL associated with two relations over their shared range of observed distances. 
In Appendix \ref{app:wasserstein}, we incorporate a measurement of the difference of the ranges, based on optimal transport, as a complementary and a priori independent point of comparison. 

For each pair $(r,k)$ of relation and residual stream checkpoint, we introduce its \emph{range} $\mathcal{I}(r,k)$, an interval with endpoints given by the minimal and maximal linear distance present in the data set, and the function $\mathrm{UASL}_\delta(r_1,k):\mathcal{I}(r,k) \to [0,1]$  that we obtain by interpolating the values
$ \{ (\delta, \, \mathrm{UASL}_\delta(r,k)) \, : \, |S_\delta(r,k)| \ge 5 \}.$
The \emph{mean squared error per residual stream checkpoint} is 
\begin{multline*}
    MSE_k(r_1,r_2)=\\\sum_{\delta\,\in\,\mathcal{I}(r_1,r_2,k)}
  \frac{\left(
    \widehat{\mathrm{UASL}}_\delta(r_1,k) - \widehat{\mathrm{UASL}}_\delta(r_2,k)\right)^{2}}{|\mathcal{I}(r_1,r_2,k)|}
\end{multline*}
 where $\mathcal{I}(r_1,r_2,k)$ is the common range $\mathcal{I}(r_1,k) \cap \mathcal{I}(r_2,k)$, and the \emph{UASL distance} defined as 
\begin{multline*}
    d_{\rm UASL}(r_1,r_2)=\\\sqrt{
    \frac{1}{K_{r_1,r_2}}
    \sum_{\substack{k\,:\\ |\mathcal{I}(r_1,r_2,k)|\,\ge\,1}}
    \!\mathrm{MSE}_k(r_1,r_2)
  }
\end{multline*}
where $K_{r_1,r_2}$ is the number of residual stream checkpoints for which the intersection
is non-empty.

Averaging first over distances, for a fixed checkpoint, and then over residual stream checkpoints, ensures uniform contribution of residual stream checkpoints regardless of the size of the $\mathcal{I}(r_1,r_2,k)$, and of distances at each checkpoint.  
Pairs with $\mathcal{I}(r_1,r_2,k)=\emptyset$ at all $k$ are assigned
$d_{\mathrm{UASL}} = 1$.

\begin{figure}
\centering\includegraphics[width=\columnwidth]{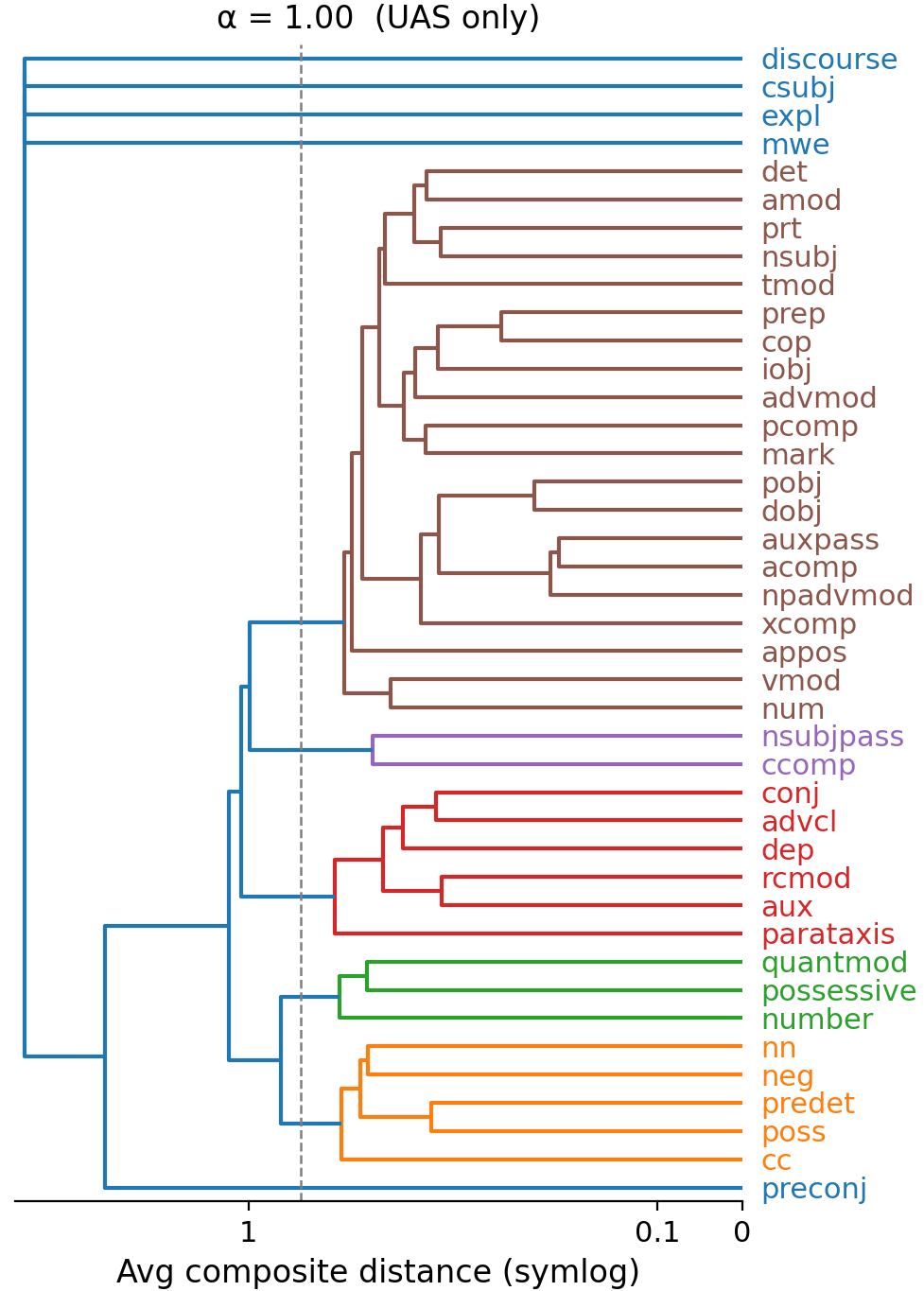}
    \caption{Dendrogram of dependency relations, clustered by UASL performance as a function of linear distance, with symmetric log-scale on the horizontal axis (linear below 0.3, logarithmic above) and a vertical line marking 25\% of the tallest merge.}
    \label{fig:UASL-only-dendrogram} 
\end{figure}

 We use the average linkage method (UPGMA),  to produce dendrograms of the agglomerative clustering of dependency relations with respect $d_{\rm UASL}$ (and with respect to a composite metric that incorporates range differences in Appendix \ref{sec:appendix-all-dendros}). The results for  $d_{\rm UASL}$ are depicted in Figure  \ref{fig:UASL-only-dendrogram}. At the highest level,  dependency relations are divided into two main clusters, colored green-yellow and red-purple-brown (we ignore the singletons for now). The green-yellow cluster consists predominantly of relations involving function words (excepting {\em nn} and {\em poss}) that are largely nominal modifiers. The relations in the red-purple-brown cluster are largely headed either uniquely by verbs or by verbs  as well other parts of speech in addition to verb, such as {\em cop/prep/npadvmod/conj}; exceptions are {\em det/amod/appos/num/rcmod}.\footnote{For an in-depth explanation of the dependency roles, see \citet{stanford-dep-manual}.}

More specifically, the green cluster ({\em possessive/number/quantmod}) involves a largely closed class of words operating as dependents of nominal heads.%
The yellow cluster, while also  involving largely nominal modification, has  greater variation in its set of dependent words. {\em Predet, poss} and {\em nn} all modify nouns in the sense that they specify or pick out the reference of their nominal head. %
In contrast, {\em neg} and {\em cc} involve closed class dependents and play functional syntactic roles, though their heads have greater variability. {\em Neg} negates its head (mostly verbs but also possibly nominals and adjectives)%
while {\em cc} indicates its head (noun/adj/prep/verb/etc) is coordinated.

The red cluster is notable for its tendency to involve clause-level dependents:  from relative clause modification {\em rcmod} to {\em parataxis/advcl} (see %
\S\ref{UUASperfsec}). It is perhaps for this reason that {\em conj} clusters into the red group rather than the yellow group where {\em cc} occurs; though {\em conj} and {\em cc} dependencies share the same class of heads, only the dependents of  {\em conj} are highly variable as opposed to  the closed-class dependents in {\em cc}. In contrast, {\em conj} dependents must be the same part of speech as their heads, and this relation can  coordinate clauses (``I \textbf{baked} cookies and John \underline{ate} them''). Such  verb dependent to verb head linkages may be a characteristic feature of the red cluster ({\em dep} excluded) and may explain why {\em ccomp} (``I \textbf{think} he \underline{needs} cookies'') is in the next nearest cluster.

The large brown cluster is defined by largely open-class dependents  (excepting {\em det/cop/prep/mark/auxpass/num}) and has a tendency towards being verbally headed (exceptions being nominally-headed {\em det/amod/appos/num} and {\em prep/npadvmod/cop}, which can be headed by a variety of parts of speech). Overall, it contains the argument relations ({\em nsubj/dobj/iobj/pobj/xcomp}) likely as a by-product of the cluster's proclivity towards verbal heads.

In sum, there is a broad tendency for the green-yellow clusters to exhibit more functionally modificational roles to nominals, while the brown-purple-red cluster shows proclivity towards verbal heads and a greater variety in the classes of words of the dependents, with red being additionally notable for its emphasis on clausal-clausal linkage. For an analysis on clustering behavior when calculated with $\tilde{d}_{\mathrm{range}}$ rather than $\tilde{d}_{\mathrm{UASL}}$, see Appendix \ref{sec:appendix-all-dendros}.

\section{Entropy analysis} \label{EntropySec}

In \S\ref{UUASperfsec}, we considered the amount of variability in the words that can play the roles of head and dependent as an explanation of the UASL of each relation type; here we make that argument quantitative. One way to measure this variability is through the entropy of the distribution of words in the head and dependent position for each relation $r$. For the head, this would be $ H_{\mathrm{head}}(r) = \mathbb E_{p^h_r}(-\log_2 p^h_r), $
where $p^h_r$ is the relative frequency estimate of the probability of word $x$ occurring as the head of dependency relation $r$. While potentially informative \citep{Simon2025},  this measure  does not account for a crucial aspect of the distribution: lexical similarity. Because the structural probes are trained over word embeddings, which reflect similarity via proximity, we would expect a relation that has many similar words as a head, say, will behave differently from a relation whose heads are more distinct from one another. We therefore use the similarity-aware entropy motivated in \S\ref{sec:relatedwork}. For a given relation $r$, similarity entropy of the head is  defined as 
\begin{multline}\label{eq:diversity}
H^{\mathrm{sim}}_{\mathrm{head}}(r) =\\-\sum_{x\in\mathrm{heads}(r)}  p^h_r(x)\log_2 \sum_{y\in \mathrm{heads}(r)} Z_{x,y} p^h_r(y). 
\end{multline}
Here $Z_{x,y}$ is a similarity score between words $x$ and $y$, which we compute as $Z_{x,y} = \max(\mathrm{cos}([\![x]\!],[\![y]\!]),0)$
where $[\![ \cdot ]\!]$ denotes a word embedding. For word embeddings, we use a 50-dimensional vector derived via PCA from the fasttext 300-dimensional wikinews word embeddings \cite{bojanowski-etal-2017-enriching}.%

To measure the impact of similarity-aware entropy, we perform a weighted least squares regression to predict UASL scores for each of the 42 dependency relations. Because UASL is a proportion, the sampling noise of each relation's score depends on how many edges it was computed from; we therefore weight each relation by that count, which also makes the abundance of a relation part of the estimator rather than a separate predictor.

In addition to similarity-aware entropy, we also take as predictors the distribution of distances between the head and dependent for each relation. Here we represent distance via the mean and standard deviation of the distribution of $\log(\text{linear distance})$ for each relation (as found in the dependency version of the Penn Treebank). The standard deviation was introduced as a way of factoring in the spread of the linear distance; our previous discussion gives us to consider this spread. Indeed, if $f_r(x)$ denotes the accuracy a probe attains on an edge of type $r$ whose log length is $x$, and  $X$ is the log length of a randomly drawn edge with label $r$, then the relation's UASL is $\mathbb{E}[f_r(X)]$. Were $f$ exactly linear---as it would be under the log-linear decay model of \S\ref{sec:logdistsec}---we would have $\mathbb{E}[f_r(X)] = f_r(\mathbb{E}[X])$. Since we find that the fit of this model varies considerably across relations, a measure of the spread is likely informative.\footnote{We choose standard deviation rather than variance as a measure of spread on an empirical basis: the variance fits worse in all five models tested and leaves the term short of significance in each of them (Appendix \ref{app:moments}). We use moments of $\log$ length rather than the log of the mean length because UASL decays in $\log$ distance rather than in distance; empirically $\mathrm{mean}(\log \delta)$ outperforms $\log(\mathrm{mean}\ \delta)$ by about $0.09$ in $R^{2}$ for every model we tested (Appendix \ref{app:moments}).}

The results of the regression are shown in  Table \ref{tab:regression_results}.
\begin{table}[t]
     \begin{tabular}{lrrr}
        \toprule
        \textbf{Predictor} & \textbf{Coef} & \textbf{SE} & \textbf{p}  \\
        \midrule
        const               &  1.045 & 0.038 & $<$.001 \\
        head sim-entropy  & -0.082 & 0.014 &   $<$.001 \\
        mean log linear dist.   & -0.252 & 0.030 & $<$.001 \\
        sd log dep length   & +0.163 & 0.066 & .019 \\
        \bottomrule
    \end{tabular}
    \caption{WLS regression predicting UASL from properties of dependency relations, all computed on our dev set (from the PTB). Each relation is weighted by the number of dev set edges its UASL was computed from. Predictors are on their natural scales ($R^{2}=0.736$, adjusted $R^{2}=0.715$, $F(3,38)=35.2$, $p<10^{-10}$, $n=42$).}
\label{tab:regression_results}
\end{table}
The large negative coefficient for mean log dependency length confirms what we found earlier: performance on UASL for a specific dependency relation \textit{decreases} as its mean log length \textit{increases}. The standard deviation of $\log(\text{linear distance})$ carries a \textit{positive} coefficient: this tells us that, holding the mean fixed, relations whose linear distances are more spread out are recovered \textit{better}: in fact, the medians are generally smaller than the means, so the spread translates into shorter edges that are easier to recover. 
Most relevantly to the current discussion, we find a \textit{strong negative} effect of the similarity-corrected entropy of the head of the relation (see Figure \ref{fig:head-sim-entropy}). All three relationships are shown visually, for every model, in Figure \ref{fig:predictor-grid} in Appendix \ref{app:regression}.

We did not find a similarly significant effect of similarity-corrected entropy of the dependent when it was added  as a predictor variable in the regression. Curiously, only dependent entropy has been explored as a factor in human sentence processing, with mixed results depending on how dependent distributions are characterized \citep{linzen2013}, though  the characterizations that were explored do not make use of the similarity correction (but see \citealp{meister2024}). Our results suggest that human experimental work could explore similarity-aware head entropy.

One possible objection to measuring lexical diversity through fastText embeddings is that these representations are not the ones that the probe actually operates on. We therefore repeated the analysis with $Z_{x,y}$ computed using the model's own contextualised representations at the probed residual stream checkpoint. Specifically, each word type is represented by the mean of its contextual vectors over its corpus occurrences.   Strikingly, the results of the WLS regression are essentially unchanged: head similarity-corrected entropy $\beta = -0.081$, $p<.001$; mean log length $\beta = -0.251$, $p<.001$; sd log length $\beta = +0.143$, $p = .046$; $R^{2} = 0.707$. 
We report the fastText-based figures in Table \ref{tab:regression_results} because that predictor is independent of the probed model, which makes the relationship an out-of-sample one rather than a statement about a geometry the probe and the predictor share.

The regression in Table \ref{tab:regression_results} supports our earlier observations: greater lexical diversity impedes performance of the structural probe. This suggests that the model's ability to identify syntactic relations depends not on abstract syntactic properties but rather on the specifics of the lexical items it is trained on. In this regard, the language itself and its associated processing difficulties may serve to predict a probe's performance in linguistically predictable ways. For example, the negative correlation between the dependency length and the UASL has parallels in cognitive science, where processing difficulty is positively associated with the increasing length between heads and dependents \cite{gibson1998,lewis2008,jaeger2011}.

\section{Conclusions} \label{ConclSec}
We have demonstrated several ways in which the complexity and variability of the dependency relations themselves can affect probe performance. 

 Long linear distances lower probe performance, an effect  also observed in humans. We also find that dependency relations with greater similarity entropy of the head are likewise negatively correlated with probe performance. Additionally, results reveal the word order itself is impactful for some of the top-performing relations that rely on word order as these relations suffer when the probe is trained on shuffled transformers, suggesting that the more templatic relations may be learned by some sort of $n$-gram detection, making them easier to acquire than other more complex relations.

 This work furthermore demonstrates the evolution within the probe. Supporting the conclusions of \citet{Someya2025}, who use derivational probes to establish that BERT builds syntactic trees in a bottom-up fashion, we analyze UASL decay to find that deeper layers of the transformer have a greater likelihood of reconstructing dependencies with longer linear distances (see Fig. \ref{fig:performance-as-a-function-of-distance}). However, not all UASL curves decay polynomially in linear distance. 

UASL is in fact better explained by three indicators: the mean of the logarithm of the linear distance associated to a syntactic dependency, the dispersion of that log length distribution (measured as standard deviation), and the similarity-aware entropy of the head of the relation.  The abundance of a relation contributes in the form of precision with which the associated UASL is estimated. Hierarchical clustering by probe performance as a function of linear distance additionally reveals linguistic clustering behavior.

We close with some remarks on abstraction.  Syntax of human languages is in essence an algebraic structure, amenable to a theoretical description in terms of precise relations. This has advantages in terms of studying its representation in both human and artificial systems, since one has a unifying abstract algebraic level to compare with. A key to abstraction, in the modeling of any kind of system, is that it can group together under a single rule a very broad variety of empirical phenomena. In fact, abstraction is all the more successful the more it is robust against a background of high individual variability. We find the notion of similarity-aware entropy especially suitable for capturing the kind of variability underlying abstraction. The fact that structural probe performance is negatively affected by higher values of the similarity-aware entropy of a syntactic relation's heads provides a precise quantitative measurement of the limitations (or at least difficulties) of abstraction in LLMs. The effect, as we show, is robust across different  models. 

Our research helps to shine light into the blackbox problem of LLMs and helps to better understand what elements are revealed by a probe. We find that a probe is not affected by structural, statistical, or linguistic elements alone; instead, these factors appear to interact and modulate each other.

\section*{Limitations}

Our main limitation is that we could not quantify the distinction between internal variation among syntactic relations and external. Additionally, we trained only one probe per model checkpoint, which meant that we did not evaluate probe stability under different corpus splits and training runs.

However, we believe these initial findings warrant future research to test similar structures in other languages in order to discover whether and how probe's performance and behaviors may further vary across model architecture and language(s). Particularly of interest would be languages with more permutational symmetry where distance may be less relevant; English shows strict adherence to word order in its syntactic construction, and languages with greater word-order variation may demonstrate interesting trade-offs between the factors our study identified as affecting probe performance. 

Additionally, we have not yet linked the structural probe performance analysis with the computational mechanisms of the transformer. It is not yet proven that the model indeed exploits the syntactic encoding detected by the structural probes for its own linguistic competence. However, we speculate that the internal variation among syntactic relations that we have detected is a valuable input for future work directed at uncovering different attentional mechanisms that might be at play in building different clusters of relations. 

\section*{Acknowledgments}
M.M. is supported by NSF grants DMS-2104330 and DMS-2506176.
Claude Code (mainly Opus) was used throughout the project for coding assistance
and for the generation of the figures; all code has been reviewed by the
authors. Claude was also used for some literature search, and for proof-reading
the final draft.

\bibliography{anthology-1, anthology-2, custom}

\begin{thebibliography}{53}
\providecommand{\natexlab}[1]{#1}

\bibitem[{Aishwarya et~al.(2026)Aishwarya, Li, Madiman, and
  Meckes}]{aishwarya2026}
Gautam Aishwarya, Dongbin Li, Mokshay Madiman, and Mark Meckes. 2026.
\newblock \href {https://doi.org/10.1090/proc/17752} {Metric complexity is a
  {Bryant--Tupper} diversity}.
\newblock \emph{Proc. Amer. Math. Soc.}, 154(10):4439--4451.

\bibitem[{Alain and Bengio(2017)}]{alain-bengio-2017-probe}
Guillaume Alain and Yoshua Bengio. 2017.
\newblock \href {https://arxiv.org/abs/1610.01644} {Understanding intermediate
  layers using linear classifier probes}.
\newblock In \emph{The 5th International Conference on Learning Representations
  (Workshop Track)}.

\bibitem[{Arps et~al.(2022)Arps, Samih, Kallmeyer, and
  Sajjad}]{arps-etal-2022-probing}
David Arps, Younes Samih, Laura Kallmeyer, and Hassan Sajjad. 2022.
\newblock \href {https://doi.org/10.18653/v1/2022.findings-emnlp.502} {Probing
  for constituency structure in neural language models}.
\newblock In \emph{Findings of the Association for Computational Linguistics:
  EMNLP 2022}, pages 6738--6757, Abu Dhabi, United Arab Emirates. Association
  for Computational Linguistics.

\bibitem[{Bojanowski et~al.(2017)Bojanowski, Grave, Joulin, and
  Mikolov}]{bojanowski-etal-2017-enriching}
Piotr Bojanowski, Edouard Grave, Armand Joulin, and Tomas Mikolov. 2017.
\newblock \href {https://doi.org/10.1162/tacl_a_00051} {Enriching word vectors
  with subword information}.
\newblock \emph{Transactions of the Association for Computational Linguistics},
  5:135--146.

\bibitem[{Cheng and Vlachos(2024)}]{Cheng2024}
Julius Cheng and Andreas Vlachos. 2024.
\newblock Measuring {{Uncertainty}} in {{Neural Machine Translation}} with
  {{Similarity-Sensitive Entropy}}.
\newblock In \emph{Proceedings of the 18th {{Conference}} of the {{European
  Chapter}} of the {{Association}} for {{Computational Linguistics}}
  ({{Volume}} 1: {{Long Papers}})}, pages 2115--2128, St. Julian's, Malta.
  Association for Computational Linguistics.

\bibitem[{Chi et~al.(2020)Chi, Hewitt, and Manning}]{chi-etal-2020-finding}
Ethan~A. Chi, John Hewitt, and Christopher~D. Manning. 2020.
\newblock \href {https://doi.org/10.18653/v1/2020.acl-main.493} {Finding
  universal grammatical relations in multilingual {BERT}}.
\newblock In \emph{Proceedings of the 58th Annual Meeting of the Association
  for Computational Linguistics}, pages 5564--5577, Online. Association for
  Computational Linguistics.

\bibitem[{Clark et~al.(2019)Clark, Khandelwal, Levy, and
  Manning}]{clark-etal-2019-bert}
Kevin Clark, Urvashi Khandelwal, Omer Levy, and Christopher~D. Manning. 2019.
\newblock \href {https://doi.org/10.18653/v1/W19-4828} {What does {BERT} look
  at? an analysis of {BERT}{'}s attention}.
\newblock In \emph{Proceedings of the 2019 ACL Workshop BlackboxNLP: Analyzing
  and Interpreting Neural Networks for NLP}, pages 276--286, Florence, Italy.
  Association for Computational Linguistics.

\bibitem[{Collins(1997)}]{collins-1997-three}
Michael Collins. 1997.
\newblock \href {https://doi.org/10.3115/976909.979620} {Three generative,
  lexicalised models for statistical parsing}.
\newblock In \emph{35th Annual Meeting of the Association for Computational
  Linguistics and 8th Conference of the {E}uropean Chapter of the Association
  for Computational Linguistics}, pages 16--23, Madrid, Spain. Association for
  Computational Linguistics.

\bibitem[{Conneau et~al.(2018)Conneau, Kruszewski, Lample, Barrault, and
  Baroni}]{conneau-etal-2018-cram}
Alexis Conneau, German Kruszewski, Guillaume Lample, Lo{\"i}c Barrault, and
  Marco Baroni. 2018.
\newblock \href {https://doi.org/10.18653/v1/P18-1198} {What you can cram into
  a single {\$}{\&}!{\#}* vector: Probing sentence embeddings for linguistic
  properties}.
\newblock In \emph{Proceedings of the 56th Annual Meeting of the Association
  for Computational Linguistics (Volume 1: Long Papers)}, pages 2126--2136,
  Melbourne, Australia. Association for Computational Linguistics.

\bibitem[{Davis et~al.(2022)Davis, Bryant, Caines, Rei, and
  Buttery}]{davis-etal-2022-probing}
Christopher Davis, Christopher Bryant, Andrew Caines, Marek Rei, and Paula
  Buttery. 2022.
\newblock \href {https://doi.org/10.18653/v1/2022.conll-1.25} {Probing for
  targeted syntactic knowledge through grammatical error detection}.
\newblock In \emph{Proceedings of the 26th Conference on Computational Natural
  Language Learning (CoNLL)}, pages 360--373, Abu Dhabi, United Arab Emirates
  (Hybrid). Association for Computational Linguistics.

\bibitem[{de~Marneffe et~al.(2006)de~Marneffe, MacCartney, and
  Manning}]{de-marneffe-etal-2006-generating}
Marie-Catherine de~Marneffe, Bill MacCartney, and Christopher~D. Manning. 2006.
\newblock \href {https://aclanthology.org/L06-1260/} {Generating typed
  dependency parses from phrase structure parses}.
\newblock In \emph{Proceedings of the Fifth International Conference on
  Language Resources and Evaluation ({LREC}{'}06)}, Genoa, Italy. European
  Language Resources Association (ELRA).

\bibitem[{de~Marneffe and Manning(2008)}]{stanford-dep-manual}
Marie-Catherine de~Marneffe and Christopher~D. Manning. 2008.
\newblock \href
  {https://downloads.cs.stanford.edu/nlp/software/dependencies_manual.pdf}
  {\emph{Stanford typed dependencies manual}}.

\bibitem[{Devlin et~al.(2019)Devlin, Chang, Lee, and
  Toutanova}]{devlin2019bert}
Jacob Devlin, Ming-Wei Chang, Kenton Lee, and Kristina Toutanova. 2019.
\newblock {BERT}: Pre-training of deep bidirectional transformers for language
  understanding.
\newblock In \emph{Proceedings of the 2019 conference of the North American
  chapter of the Association for Computational Linguistics: Human Language
  Technologies, volume 1 (long and short papers)}, pages 4171--4186.

\bibitem[{{Diego-Sim{\'o}n} et~al.(2024){Diego-Sim{\'o}n}, D'Ascoli, Chemla,
  Lakretz, and King}]{Diego-Simon2024a}
Pablo {Diego-Sim{\'o}n}, St{\'e}phane D'Ascoli, Emmanuel Chemla, Yair Lakretz,
  and Jean-R{\'e}mi King. 2024.
\newblock \href {https://arxiv.org/abs/2412.05571} {A polar coordinate system
  represents syntax in large language models}.
\newblock In \emph{Advances in Neural Information Processing Systems},
  volume~37.

\bibitem[{{Diego-Sim\'{o}n} et~al.(2025){Diego-Sim\'{o}n}, Chemla, King, and
  Lakretz}]{Simon2025}
Pablo~J. {Diego-Sim\'{o}n}, Emmanuel Chemla, Jean-Remi King, and Yair Lakretz.
  2025.
\newblock \href {https://arxiv.org/abs/2508.03211} {Probing {{Syntax}} in
  {{Large Language Models}}: {{Successes}} and {{Remaining Challenges}}}.
\newblock In \emph{Second {{Conference}} on {{Language Modeling}}}.

\bibitem[{Eisape et~al.(2022)Eisape, Gangireddy, Levy, and
  Kim}]{eisape-etal-2022-probing}
Tiwalayo Eisape, Vineet Gangireddy, Roger Levy, and Yoon Kim. 2022.
\newblock \href {https://doi.org/10.18653/v1/2022.findings-emnlp.203} {Probing
  for incremental parse states in autoregressive language models}.
\newblock In \emph{Findings of the Association for Computational Linguistics:
  EMNLP 2022}, pages 2801--2813, Abu Dhabi, United Arab Emirates. Association
  for Computational Linguistics.

\bibitem[{Elhage et~al.(2021)Elhage, Nanda, Olsson, Henighan, Joseph, Mann,
  Askell, Bai, Chen, Conerly, DasSarma, Drain, Ganguli, {Hatfield-Dodds},
  Hernandez, Jones, Kernion, Lovitt, Ndousse, Amodei, Brown, Clark, Kaplan,
  McCandlish, and Olah}]{Elhage2021}
Nelson Elhage, Neel Nanda, Catherine Olsson, Tom Henighan, Nicholas Joseph, Ben
  Mann, Amanda Askell, Yuntao Bai, Anna Chen, Tom Conerly, Nova DasSarma, Dawn
  Drain, Deep Ganguli, Zac {Hatfield-Dodds}, Danny Hernandez, Andy Jones,
  Jackson Kernion, Liane Lovitt, Kamal Ndousse, and 6 others. 2021.
\newblock \href {https://transformer-circuits.pub/2021/framework/index.html} {A
  mathematical framework for transformer circuits}.
\newblock \emph{Transformer Circuits Thread}.

\bibitem[{Ferrando et~al.(2024)Ferrando, Sarti, Bisazza, and
  {Costa-juss{\`a}}}]{Ferrando2024}
Javier Ferrando, Gabriele Sarti, Arianna Bisazza, and Marta~R.
  {Costa-juss{\`a}}. 2024.
\newblock \href {https://doi.org/10.48550/arXiv.2405.00208} {A primer on the
  inner workings of transformer-based language models}.
\newblock arXiv preprint \texttt{arXiv:2405.00208}.
\newblock \emph{Preprint}, arXiv:2405.00208.

\bibitem[{Gar{\'i}~Soler and
  Apidianaki(2021)}]{gari-soler-apidianaki-2021-lets}
Aina Gar{\'i}~Soler and Marianna Apidianaki. 2021.
\newblock \href {https://doi.org/10.1162/tacl_a_00400} {Let{'}s play mono-poly:
  {BERT} can reveal words' polysemy level and partitionability into senses}.
\newblock \emph{Transactions of the Association for Computational Linguistics},
  9:825--844.

\bibitem[{Gibson(1998)}]{gibson1998}
Edward Gibson. 1998.
\newblock Linguistic complexity: locality of syntactic dependencies.
\newblock \emph{Cognition}, 68:1--76.

\bibitem[{Grindrod(2025)}]{Grindrod2025}
Jumbly Grindrod. 2025.
\newblock \href {https://doi.org/10.3998/ergo.9261} {Transformers,
  contextualism, and polysemy}.
\newblock \emph{Ergo: An Open Access Journal of Philosophy}.

\bibitem[{Gupta et~al.(2022)Gupta, Shi, Gimpel, and Sachan}]{gupta2022deep}
Vikram Gupta, Haoyue Shi, Kevin Gimpel, and Mrinmaya Sachan. 2022.
\newblock Deep clustering of text representations for supervision-free probing
  of syntax.
\newblock In \emph{Proceedings of the AAAI Conference on Artificial
  Intelligence}, volume~36, pages 10720--10728.

\bibitem[{Hale and Keyser(2002)}]{hale-argustructure}
Ken Hale and Samuel~Jay Keyser. 2002.
\newblock \emph{The Basic Elements of Argument Structure}.
\newblock Number~39 in Linguistic Inquiry Monographs. MIT Press, Cambridge, MA.

\bibitem[{Harley(2017)}]{harley-vp-argu}
Heidi Harley. 2017.
\newblock \href {https://doi.org/10.1093/oso/9780198767886.003.0001} {The
  “bundling” hypothesis and the disparate functions of little v}.
\newblock In Roberta D'Alessandro, Irene Franco, and Ángel J.~Gallego,
  editors, \emph{The Verbal Domain}, Oxford Studies in Theoretical Linguistics,
  pages 3--28. Oxford University Press, Oxford.

\bibitem[{He et~al.(2023)He, Gao, and
  Chen}]{he2023debertav3improvingdebertausing}
Pengcheng He, Jianfeng Gao, and Weizhu Chen. 2023.
\newblock \href {https://arxiv.org/abs/2111.09543} {{DeBERTaV3}: Improving
  {DeBERTa} using {ELECTRA}-style pre-training with gradient-disentangled
  embedding sharing}.
\newblock In \emph{The Eleventh International Conference on Learning
  Representations}.

\bibitem[{Hewitt et~al.(2021)Hewitt, Ethayarajh, Liang, and
  Manning}]{Hewitt2021a}
John Hewitt, Kawin Ethayarajh, Percy Liang, and Christopher Manning. 2021.
\newblock \href {https://doi.org/10.18653/v1/2021.emnlp-main.122} {Conditional
  probing: Measuring usable information beyond a baseline}.
\newblock In \emph{Proceedings of the 2021 {{Conference}} on {{Empirical
  Methods}} in {{Natural Language Processing}}}, pages 1626--1639, Online and
  Punta Cana, Dominican Republic. Association for Computational Linguistics.

\bibitem[{Hewitt and Manning(2019)}]{hewitt2019structural}
John Hewitt and Christopher~D Manning. 2019.
\newblock A structural probe for finding syntax in word representations.
\newblock In \emph{Proceedings of the 2019 Conference of the North American
  Chapter of the Association for Computational Linguistics: {{Human}} Language
  Technologies, Volume 1 (Long and Short Papers)}, pages 4129--4138.

\bibitem[{Hornstein and Pietroski(2009)}]{compl-int-argu}
Norbert Hornstein and Paul Pietroski. 2009.
\newblock \href {https://doi.org/10.5565/rev/catjl.157} {Basic operations:
  Minimal syntax-semantics}.
\newblock \emph{Catalan Journal of Linguistics}, 8(1):113--139.

\bibitem[{Huh et~al.(2024)Huh, Cheung, Wang, and Isola}]{huh2024position}
Minyoung Huh, Brian Cheung, Tongzhou Wang, and Phillip Isola. 2024.
\newblock Position: {{The}} platonic representation hypothesis.
\newblock In \emph{Forty-First International Conference on Machine Learning}.

\bibitem[{Jaeger and Tily(2011)}]{jaeger2011}
T.~Florian Jaeger and Harry Tily. 2011.
\newblock \href {https://doi.org/10.1002/wcs.126} {On language ‘utility’:
  processing complexity and communicative efficiency}.
\newblock \emph{WIREs Cognitive Science}, 2(3):323--335.

\bibitem[{Jha et~al.(2025)Jha, Zhang, Shmatikov, and
  Morris}]{jha2026harnessing}
Rishi Jha, Collin Zhang, Vitaly Shmatikov, and John Morris. 2025.
\newblock \href {https://arxiv.org/abs/2505.12540} {Harnessing the universal
  geometry of embeddings}.
\newblock \emph{Advances in Neural Information Processing Systems},
  38:45963--45987.

\bibitem[{Kratzer(1996)}]{Kratzer1996}
Angelika Kratzer. 1996.
\newblock \href {https://doi.org/10.1007/978-94-015-8617-7_5} {Severing the
  external argument from its verb}.
\newblock In Johan Rooryck and Laurie Zaring, editors, \emph{Phrase Structure
  and the Lexicon}, pages 109--137. Springer, Dordrecht.

\bibitem[{Leinster(2021)}]{Leinster2021a}
Tom Leinster. 2021.
\newblock \emph{Entropy and Diversity: {{The}} Axiomatic Approach}.
\newblock Cambridge University Press.

\bibitem[{Leinster and Cobbold(2012)}]{Leinster2012}
Tom Leinster and Christina~A. Cobbold. 2012.
\newblock \href {https://doi.org/10.1890/10-2402.1} {Measuring diversity: The
  importance of species similarity}.
\newblock \emph{Ecology}, 93(3):477--489.

\bibitem[{Lewis et~al.(2006)Lewis, Vasishth, and Dyke}]{lewis2008}
Richard~L. Lewis, Shravan Vasishth, and Julie A.~Van Dyke. 2006.
\newblock \href {https://doi.org/10.1016/j.tics.2006.08.007} {Computational
  principles of working memory in sentence comprehension}.
\newblock \emph{Trends in Cognitive Sciences}, 10(10):447--454.

\bibitem[{Linzen et~al.(2013)Linzen, Marantz, and Pylkk{\"a}nen}]{linzen2013}
Tal Linzen, Alec Marantz, and Liina Pylkk{\"a}nen. 2013.
\newblock \href {https://doi.org/10.1075/ml.8.2.01lin} {Syntactic context
  effects in visual word recognition: An {MEG} study}.
\newblock \emph{The Mental Lexicon}, 8(2):117--139.

\bibitem[{Liu et~al.(2019)Liu, Ott, Goyal, Du, Joshi, Chen, Levy, Lewis,
  Zettlemoyer, and Stoyanov}]{liu-etal-2019-roberta}
Yinhan Liu, Myle Ott, Naman Goyal, Jingfei Du, Mandar Joshi, Danqi Chen, Omer
  Levy, Mike Lewis, Luke Zettlemoyer, and Veselin Stoyanov. 2019.
\newblock \href {https://arxiv.org/abs/1907.11692} {{RoBERTa}: A robustly
  optimized {BERT} pretraining approach}.
\newblock \emph{Preprint}, arXiv:1907.11692.

\bibitem[{Manning et~al.(2020)Manning, Clark, Hewitt, Khandelwal, and
  Levy}]{Manning2020}
Christopher~D. Manning, Kevin Clark, John Hewitt, Urvashi Khandelwal, and Omer
  Levy. 2020.
\newblock \href {https://doi.org/10.1073/pnas.1907367117} {Emergent linguistic
  structure in artificial neural networks trained by self-supervision}.
\newblock \emph{Proceedings of the National Academy of Sciences},
  117(48):30046--30054.

\bibitem[{Marcus et~al.(1993)Marcus, Santorini, and
  Marcinkiewicz}]{marcus-etal-1993-building}
Mitchell~P. Marcus, Beatrice Santorini, and Mary~Ann Marcinkiewicz. 1993.
\newblock \href {https://aclanthology.org/J93-2004/} {Building a large
  annotated corpus of {E}nglish: The {P}enn {T}reebank}.
\newblock \emph{Computational Linguistics}, 19(2):313--330.

\bibitem[{Meister et~al.(2024)Meister, Giulianelli, and Pimentel}]{meister2024}
Clara Meister, Mario Giulianelli, and Tiago Pimentel. 2024.
\newblock Towards a similarity-adjusted surprisal theory.
\newblock In \emph{Proceedings of the 2024 conference on Empirical Methods in
  Natural Language Processing}, pages 16485--16498.

\bibitem[{Piantadosi et~al.(2024)Piantadosi, Muller, Rule, Kaushik, Gorenstein,
  Leib, and Sanford}]{Piantadosi2024}
Steven~T. Piantadosi, Dyana~C.Y. Muller, Joshua~S. Rule, Karthikeya Kaushik,
  Mark Gorenstein, Elena~R. Leib, and Emily Sanford. 2024.
\newblock \href {https://doi.org/10.1016/j.tics.2024.06.011} {Why concepts are
  (probably) vectors}.
\newblock \emph{Trends in Cognitive Sciences}, 28(9):844--856.

\bibitem[{Posada et~al.(2020)Posada, Vani, Schwarzer, and
  Lacoste-Julien}]{posada2020}
Jose~Gallego Posada, Ankit Vani, Max Schwarzer, and Simon Lacoste-Julien. 2020.
\newblock \href {https://proceedings.mlr.press/v108/posada20a.html} {{GAIT}: A
  geometric approach to information theory}.
\newblock In \emph{Proceedings of the Twenty Third International Conference on
  Artificial Intelligence and Statistics}, volume 108 of \emph{Proceedings of
  Machine Learning Research}, pages 2601--2611. PMLR.

\bibitem[{Radford et~al.(2019)Radford, Wu, Child, Luan, Amodei, and
  Sutskever}]{radford-etal-2019-gpt2}
Alec Radford, Jeff Wu, Rewon Child, David Luan, Dario Amodei, and Ilya
  Sutskever. 2019.
\newblock \href
  {https://cdn.openai.com/better-language-models/language_models_are_unsupervised_multitask_learners.pdf}
  {Language models are unsupervised multitask learners}.
\newblock Technical report, OpenAI.

\bibitem[{Reif et~al.(2019)Reif, Yuan, Wattenberg, Viegas, Coenen, Pearce, and
  Kim}]{reif2019visualizing}
Emily Reif, Ann Yuan, Martin Wattenberg, Fernanda~B Viegas, Andy Coenen, Adam
  Pearce, and Been Kim. 2019.
\newblock Visualizing and measuring the geometry of {BERT}.
\newblock \emph{Advances in neural information processing systems}, 32.

\bibitem[{Rogers et~al.(2020)Rogers, Kovaleva, and Rumshisky}]{Rogers2020}
Anna Rogers, Olga Kovaleva, and Anna Rumshisky. 2020.
\newblock \href {https://doi.org/10.1162/tacl_a_00349} {A {{Primer}} in
  {{BERTology}}: {{What We Know About How BERT Works}}}.
\newblock \emph{Transactions of the Association for Computational Linguistics},
  8:842--866.

\bibitem[{Shepard(2024)}]{Shepard2024}
Roger~N. Shepard. 2024.
\newblock \href {https://doi.org/10.4324/9781032722450-5} {\emph{Form,
  {{Formation}}, and {{Transformation}} of {{Internal Representations}}}}, 1
  edition, pages 87--122.
\newblock Routledge, London.

\bibitem[{Sinha et~al.(2021)Sinha, Jia, Hupkes, Pineau, Williams, and
  Kiela}]{Sinha2021}
Koustuv Sinha, Robin Jia, Dieuwke Hupkes, Joelle Pineau, Adina Williams, and
  Douwe Kiela. 2021.
\newblock \href {https://doi.org/10.18653/v1/2021.emnlp-main.230} {Masked
  {{Language Modeling}} and the {{Distributional Hypothesis}}: {{Order Word
  Matters Pre-training}} for {{Little}}}.
\newblock In \emph{Proceedings of the 2021 {{Conference}} on {{Empirical
  Methods}} in {{Natural Language Processing}}}, pages 2888--2913, Online and
  Punta Cana, Dominican Republic. Association for Computational Linguistics.

\bibitem[{Someya et~al.(2025)Someya, Yoshida, Yanaka, and Oseki}]{Someya2025}
Taiga Someya, Ryo Yoshida, Hitomi Yanaka, and Yohei Oseki. 2025.
\newblock \href {https://doi.org/10.18653/v1/2025.conll-1.7} {Derivational
  probing: Unveiling the layer-wise derivation of syntactic structures in
  neural language models}.
\newblock In \emph{Proceedings of the 29th Conference on Computational Natural
  Language Learning}, pages 93--104, Vienna, Austria. Association for
  Computational Linguistics.

\bibitem[{Trott and Bergen(2023)}]{Trott2023}
Sean Trott and Benjamin Bergen. 2023.
\newblock \href {https://doi.org/10.1037/rev0000420} {Word meaning is both
  categorical and continuous.}
\newblock \emph{Psychological Review}, 130(5):1239--1261.

\bibitem[{Tucker et~al.(2022)Tucker, Eisape, Qian, Levy, and
  Shah}]{tucker-etal-2022-syntax}
Mycal Tucker, Tiwalayo Eisape, Peng Qian, Roger Levy, and Julie Shah. 2022.
\newblock \href {https://doi.org/10.18653/v1/2022.naacl-main.394} {When does
  syntax mediate neural language model performance? evidence from dropout
  probes}.
\newblock In \emph{Proceedings of the 2022 Conference of the North American
  Chapter of the Association for Computational Linguistics: Human Language
  Technologies}, pages 5393--5408, Seattle, United States. Association for
  Computational Linguistics.

\bibitem[{Vaswani et~al.(2017)Vaswani, Shazeer, Parmar, Uszkoreit, Jones,
  Gomez, Kaiser, and Polosukhin}]{Vaswani2017}
Ashish Vaswani, Noam Shazeer, Niki Parmar, Jakob Uszkoreit, Llion Jones,
  Aidan~N Gomez, {\L}ukasz Kaiser, and Illia Polosukhin. 2017.
\newblock Attention is all you need.
\newblock \emph{Advances in neural information processing systems}, 30.

\bibitem[{Wang and Komatsuzaki(2021)}]{gpt-j}
Ben Wang and Aran Komatsuzaki. 2021.
\newblock {GPT-J-6B: A 6 Billion Parameter Autoregressive Language Model}.
\newblock \url{https://github.com/kingoflolz/mesh-transformer-jax}.

\bibitem[{Warner et~al.(2025)Warner, Chaffin, Clavi{\'e}, Weller,
  Hallstr{\"o}m, Taghadouini, Gallagher, Biswas, Ladhak, Aarsen, Adams, Howard,
  and Poli}]{warner-etal-2025-smarter}
Benjamin Warner, Antoine Chaffin, Benjamin Clavi{\'e}, Orion Weller, Oskar
  Hallstr{\"o}m, Said Taghadouini, Alexis Gallagher, Raja Biswas, Faisal
  Ladhak, Tom Aarsen, Griffin~Thomas Adams, Jeremy Howard, and Iacopo Poli.
  2025.
\newblock \href {https://doi.org/10.18653/v1/2025.acl-long.127} {Smarter,
  better, faster, longer: A modern bidirectional encoder for fast, memory
  efficient, and long context finetuning and inference}.
\newblock In \emph{Proceedings of the 63rd Annual Meeting of the Association
  for Computational Linguistics (Volume 1: Long Papers)}, pages 2526--2547,
  Vienna, Austria. Association for Computational Linguistics.

\end{thebibliography}

\appendix

\counterwithin{figure}{section}
\counterwithin{table}{section}

\clearpage
\section{Embeddings, alignment and probes in detail}
\label{app:embeddings}

This appendix states precisely the choices summarized in \S\ref{sec:embeddings} and \S\ref{sec:probes}.

\subsection{Residual stream checkpoints}
\label{app:checkpoints}

\begin{table*}[b]
\centering
\small
\setlength{\tabcolsep}{4pt}
\begin{tabular}{l|l|l|l|l|r|r|r|r}
\toprule
\textbf{Model} & \textbf{Type} & \textbf{Layer norm.} & \textbf{Attn./MLP} & \textbf{Position encoding} & $L$ & $d$ & \textbf{ckpts} & \textbf{best} \\
\midrule
BERT-base          & encoder & post-LN & sequential & learned absolute, in the stream     & 12 &  768 & 26 & 16 \\
DeBERTa-v3-base    & encoder & post-LN & sequential & disentangled relative, in attention & 12 &  768 & 13 &  9 \\
ModernBERT-base    & encoder & pre-LN  & sequential & rotary, in attention                & 22 &  768 & 46 & 30 \\
GPT-2-base         & decoder & pre-LN  & sequential & learned absolute, in the stream     & 12 &  768 & 26 & 16 \\
GPT-J-6B           & decoder & pre-LN  & parallel   & rotary, in attention                & 28 & 4096 & 29 &  8 \\
RoBERTa-Shuffle-n1 & encoder & post-LN & sequential & learned absolute, in the stream     & 12 &  768 & 13 &  9 \\
\bottomrule
\end{tabular}
\caption{The six models probed in this work. \emph{Type} specifies the attention pattern and pre-training objective, \emph{Attn./MLP} whether a block applies attention and MLP in sequence or in parallel, and \emph{Position encoding} where positional information enters the computation; $L$ is the number of blocks, $d$ the model dimension, \textbf{ckpts} the number of residual stream checkpoints we introduced and \textbf{best} the one at which development UUAS peaks. In general, \textbf{ckpts} follows from the architecture, with the exception of DeBERTa-v3-base and RoBERTa-Shuffle-n1 for which we only extracted uncontextualized embeddings with positional information and post-transformer-block checkpoints.}
\label{tab:models}
\end{table*}

We study transformer-based language models that follow different architectural choices, summarized in Table \ref{tab:models}. Crucially, these include  pre-layer or post-layer normalization, and  sequential or parallel placement of attention and MLP modules. The exact formulas that describe the residual stream checkpoints depend on these choices. 

 In all cases, the model first tokenizes the text and then embeds each token in a space of dimension $d = d_{\rm model}$, which yields a sequence of vectors $x_{1:n}^{(0)}=(x_1^{(0)}, \ldots, x_n^{(0)})$. An encoding $z$ of the position is  added to these raw embeddings, which yields $x_i^{(1)} = L\big(x_i^{(0)} + z(i)\big)$ for a post-layer-normalization model and $x_i^{(1)} = x_i^{(0)} + z(i)$ otherwise.

  The embedding $x^{(1)}$ is modified by $L$ subsequent ``attention blocks'' made of an attention module and an MLP module. 
 For each block, indexed by $k=0,..., L-1$, we denote by $A_k$  its  multi-head attention module, by $F_k$ its MLP module (a feed-forward network with one hidden layer) and by $L_\bullet^{(k)}$ its layer normalizations. We remark that while attention acts on the whole sequence, MLP acts separately on each token. 

For encoder-only sequential models with post-layer normalization (BERT, RoBERTa, DeBERTa), the application of an attention block $k=0,...,L-1$ induces new embeddings:
\begin{align}
x_i^{(2k+2)} &= L_1^{(k)}\big(x_i^{(2k+1)} + A_k(x_{1:n}^{(2k+1)})_i\big), \label{eq:postln-attn}\\
x_i^{(2k+3)} &= L_2^{(k)}\big(x_i^{(2k+2)} + F_k(x_i^{(2k+2)})\big), \label{eq:postln-ffn}
\end{align}
Similarly, in the case of encoder-only sequential models with pre-layer normalization (ModernBERT):
\begin{align}
x_i^{(2k+2)} &= x_i^{(2k+1)} + A_k\big(L_1^{(k)}(x_{1:n}^{(2k+1)})\big)_i, \label{eq:preln-attn}\\
x_i^{(2k+3)} &= x_i^{(2k+2)} + F_k\big(L_2^{(k)}(x_i^{(2k+2)})\big), \label{eq:preln-ffn}
\end{align} 
The decoder-only version (GPT-2) has $A_k\big(L_1^{(k)}(x_{1:i}^{(2k+1)})\big)_i$ instead of $A_k\big(L_1^{(k)}(x_{1:n}^{(2k+1)})\big)_i$.

Finally, a decoder-only parallel model with pre-layer normalization (GPT-J) is such that a single normalization feeds both the attention and MLP module, which are applied  in parallel, yielding
\begin{equation}
x_i^{(k+1)} = x_i^{(k)} + A_k(u_{1:i}^{(k)})_i + F_k(u_i^{(k)}), \label{eq:parallel}
\end{equation}
where $u^{(k)} = L^{(k)}(x^{(k)})$ is the single normalized copy both branches read.

We refer to the indices of these generated representations as  \emph{residual stream checkpoints}.  In general, sequential models have $2L+2$ checkpoints, indexed by   $r= 0, 1, ..., 2k+2, 2k +3, ..., 2L, 2l +1$. GPT-J's rotary embeddings are applied inside attention and never enter the stream, and its embedding layer applies no normalization, which yields $1+L=29$ checkpoints. ModernBERT-base is rotary as well (no $z$ is added to the residual stream) but its embedding layer still applies a layer normalization, which yields $2L+2$.

Language models generate probabilistic predictions over the vocabulary by applying  an affine transformation followed by a softmax  to the embeddings associated with the last residual stream checkpoint. For probing, we retrieve  $x^{(r)}$ at the checkpoints specified in Table \ref{tab:models} and its caption.

\begin{table*}
\centering\small
\begin{tabular}{lrrr@{\hspace{3em}}lrrr}
\toprule
\textbf{Relation} & \textbf{Train} & \textbf{Dev} & \textbf{Test} & \textbf{Relation} & \textbf{Train} & \textbf{Dev} & \textbf{Test} \\
\midrule
\textit{acomp}     &  1,163 &    48 &    99 & \textit{neg}        &  5,370 &   199 &   371 \\
\textit{advcl}     &  6,210 &   293 &   372 & \textit{nn}         & 74,082 & 3,228 & 4,364 \\
\textit{advmod}    & 29,282 & 1,276 & 1,839 & \textit{npadvmod}   &  3,451 &   189 &   231 \\
\textit{amod}      & 60,276 & 2,506 & 3,481 & \textit{nsubj}      & 66,179 & 2,780 & 3,991 \\
\textit{appos}     &  6,926 &   276 &   466 & \textit{nsubjpass}  &  6,625 &   253 &   386 \\
\textit{aux}       & 33,875 & 1,240 & 2,148 & \textit{num}        & 22,606 & 1,148 & 1,238 \\
\textit{auxpass}   &  7,448 &   286 &   435 & \textit{number}     &  8,798 &   448 &   454 \\
\textit{cc}        & 24,003 & 1,003 & 1,383 & \textit{parataxis}  &  1,643 &    56 &    95 \\
\textit{ccomp}     & 15,297 &   553 & 1,003 & \textit{pcomp}      &  5,123 &   208 &   240 \\
\textit{conj}      & 24,378 & 1,009 & 1,350 & \textit{pobj}       & 89,665 & 3,745 & 5,282 \\
\textit{cop}       &  9,161 &   330 &   578 & \textit{poss}       & 16,805 &   709 & 1,056 \\
\textit{csubj}     &    379 &    15 &    20 & \textit{possessive} &  8,719 &   425 &   547 \\
\textit{csubjpass} &     10 &     1 &     0 & \textit{preconj}    &    345 &    11 &    18 \\
\textit{dep}       & 10,903 &   564 &   657 & \textit{predet}     &    362 &    14 &    18 \\
\textit{det}       & 77,699 & 3,327 & 4,548 & \textit{prep}       & 91,893 & 3,783 & 5,376 \\
\textit{discourse} &    106 &     7 &    11 & \textit{prt}        &  2,628 &   116 &   159 \\
\textit{dobj}      & 40,199 & 1,637 & 2,172 & \textit{quantmod}   &  3,442 &   192 &   202 \\
\textit{expl}      &    855 &    32 &    57 & \textit{rcmod}      &  7,686 &   290 &   455 \\
\textit{iobj}      &    554 &    18 &    23 & \textit{tmod}       &  4,281 &   244 &   274 \\
\textit{mark}      & 10,326 &   421 &   655 & \textit{vmod}       &  9,319 &   357 &   524 \\
\textit{mwe}       &  1,393 &    51 &    88 & \textit{xcomp}      & 11,001 &   398 &   666 \\
\midrule
\textbf{All} & 800,466 & 33,686 & 47,332 & & & & \\
\bottomrule
\end{tabular}
\caption{Gold edge counts per dependency relation, for each Penn Treebank WSJ split. A relation's count is the number of edges the probe evaluation scores under it: edges of the gold dependency tree whose two endpoints are both non-punctuation tokens. Root arcs are excluded, having only one endpoint among the tokens, as are the edges the evaluation inserts to bridge removed punctuation, which carry no single relation. The dev counts are the weights of the weighted regression of \S\ref{EntropySec}, and the test counts the denominators of the held-out UASL reported there. \textit{csubjpass} occurs once in dev and not at all in test, so it drops out of every test-set analysis.}
\label{tab:edge-counts}
\end{table*}

\subsection{The probe and its training}
\label{app:probe}

Let $d_{T_s}(w_i, w_j)$ be the number of edges on the path between $w_i$ and $w_j$ in the gold tree $T_s$. The probe is a matrix $B \in \mathbb{R}^{k \times d}$ minimizing
\begin{equation}
\sum_{s} \frac{1}{|V_s|^{2}} \sum_{i,j} \Big|  d_{T_s}(w_i, w_j) - \big\lVert B(x_i - x_j) \big\rVert^{2} \Big| ,
\end{equation}
the inner sum running over all pairs of words of the sentence and the normalization by $|V_s|^2$ preventing long sentences from dominating. We optimize with Adam at a learning rate of $10^{-3}$, reduced by a factor of ten when the development loss stops improving, in batches of 20 sentences and for at most 40 epochs, retaining the probe with the best development loss. Rank $k=64$ is the value used by \citet{hewitt2019structural}. Nothing here varies across models or residual stream checkpoints except the model dimension $d$.

\subsection{UUAS and UASL}
\label{app:metrics}

The tree associated with a sentence $s$ has vertices $V_s$ (its words) and edges $E_s$ labelled by a function $\ell_s : E_s \to \Lambda$, where $\Lambda$ is the set of dependency labels; for each $\lambda \in \Lambda$ we write $E_s(\lambda) = \ell_s^{-1}(\lambda)$. The probe's reconstruction is $\hat T_s = (V_s, \hat E_s)$, the minimum spanning tree of the projected points $Bx_\bullet$ under the Euclidean distance of $\mathbb{R}^{64}$. Edges are compared as unordered pairs, the reconstruction being undirected, and the two scores are
\begin{equation}
{\rm UUAS} = \frac{\sum_{s} |E_s \cap \hat E_s|}{\sum_{s} |E_s|} ,
\end{equation}
\begin{equation}
{\rm UASL}(\lambda) = \frac{\sum_{s} |E_s(\lambda) \cap \hat E_s|}{\sum_{s} |E_s(\lambda)|} .
\end{equation}
Both are pooled over the data set rather than averaged sentence by sentence, so that every edge counts once and a relation's score is not dominated by short sentences in which it happens to occur. It follows that UUAS is the average of ${\rm UASL}(\lambda)$ over relations weighted by their edge counts, which is why those counts appear as the regression weights in \S\ref{EntropySec}.

\subsection{Another look at layer normalization}
\label{app:convention}

We remark that in the case of models with post-layer normalization, the residual stream is bounded in norm, whereas the models with pre-layer normalization have unbounded residual streams. 

It is conceivable to train structural probes on layer-normalized versions of the pre-LN models' checkpoints; these are precisely the representations that are manipulated by the attention or MLP block located just after the checkpoint. 
In this section, we will refer the extraction of residual stream vectors as described in the previous subsection as ``convention A'', and to the layer-normalized versions of them as ``convention B''. 

A priori, probes are not indifferent to the convention. Remember that layer normalization is given by 
\begin{equation}
L(u) = \gamma \odot \frac{u - \mu(u)\mathbf{1}}{\sigma(u)} + \beta,
\end{equation}
where $\mu$ and $\sigma$ are the mean and standard deviation over the $d$ coordinates of $u$, and $\gamma, \beta \in \mathbb{R}^d$ are learned. Of the four operations composing $L$, three can be absorbed by the probe matrix $B$: the shift $\beta$ cancels in the difference $x_i - x_j$ on which the probe is evaluated, the gain $\gamma$ is diagonal, and the centering is the fixed projection $I - \mathbf{1}\mathbf{1}^{\!\top}\!/d$. What remains is the rescaling by $1/\sigma(x_i)$, which depends on the token and so is representable by no fixed linear map. The A-versus-B contrast isolates exactly this.

That said, the performance of structural probes trained on GPT-2-base representations barely distinguishes the two conventions. The probes attain a peak development UUAS of $0.7755$ at checkpoint 16 under A and $0.7741$ at checkpoint 15 under B; across checkpoints the mean difference $\mathrm{B} - \mathrm{A}$ is $-0.0050$, exceeding $0.015$ in magnitude only at the deepest checkpoints, where residual-stream norms grow sharply under A.

\clearpage

\section{UASL by relation across residual stream checkpoints}
\label{app:other-models}

\begin{figure*}[t]
\centering
\includegraphics[width=0.45\textwidth]{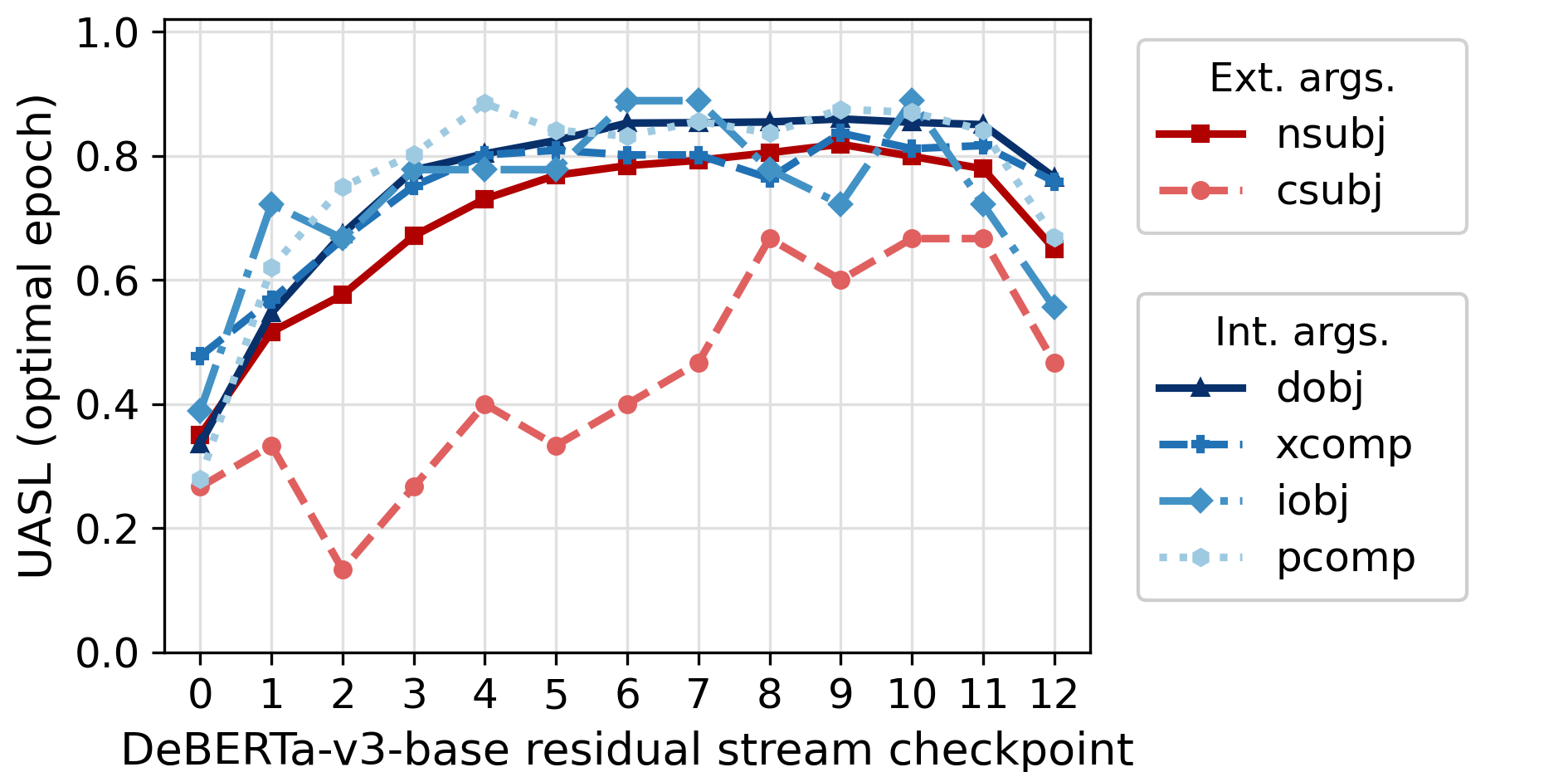}\hfill
\includegraphics[width=0.45\textwidth]{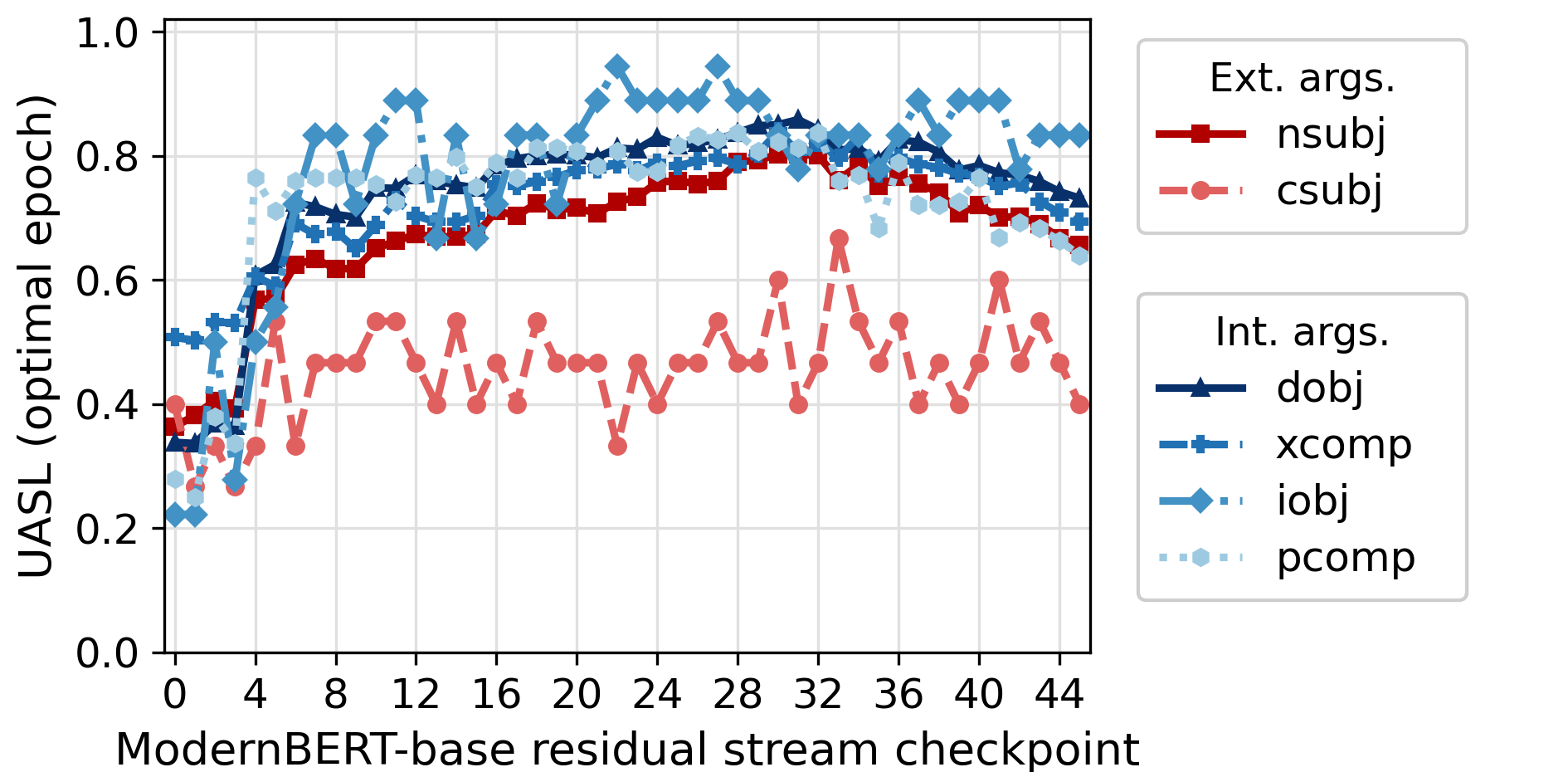}\\[0.6em]
\includegraphics[width=0.45\textwidth]{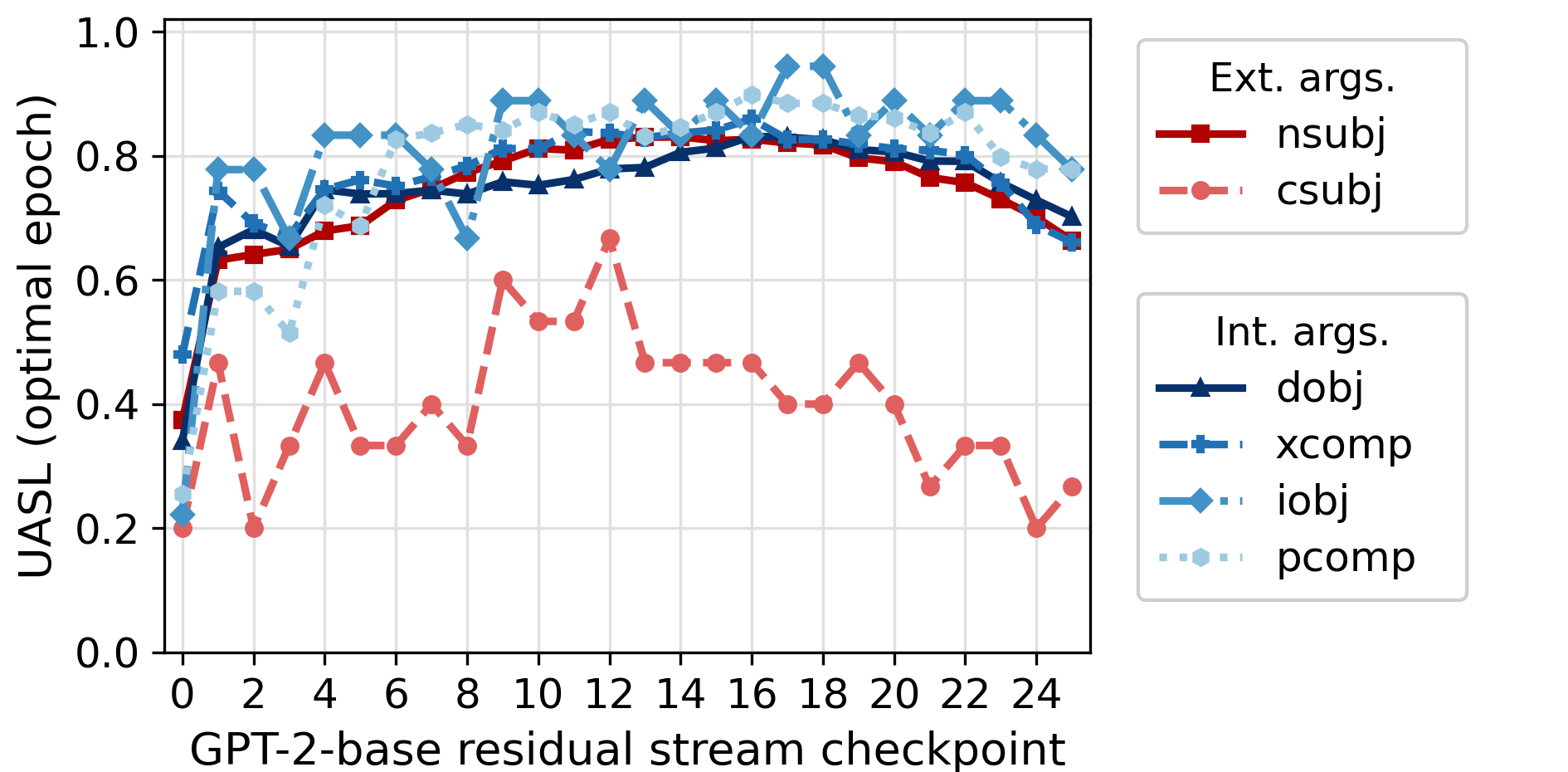}\hfill
\includegraphics[width=0.45\textwidth]{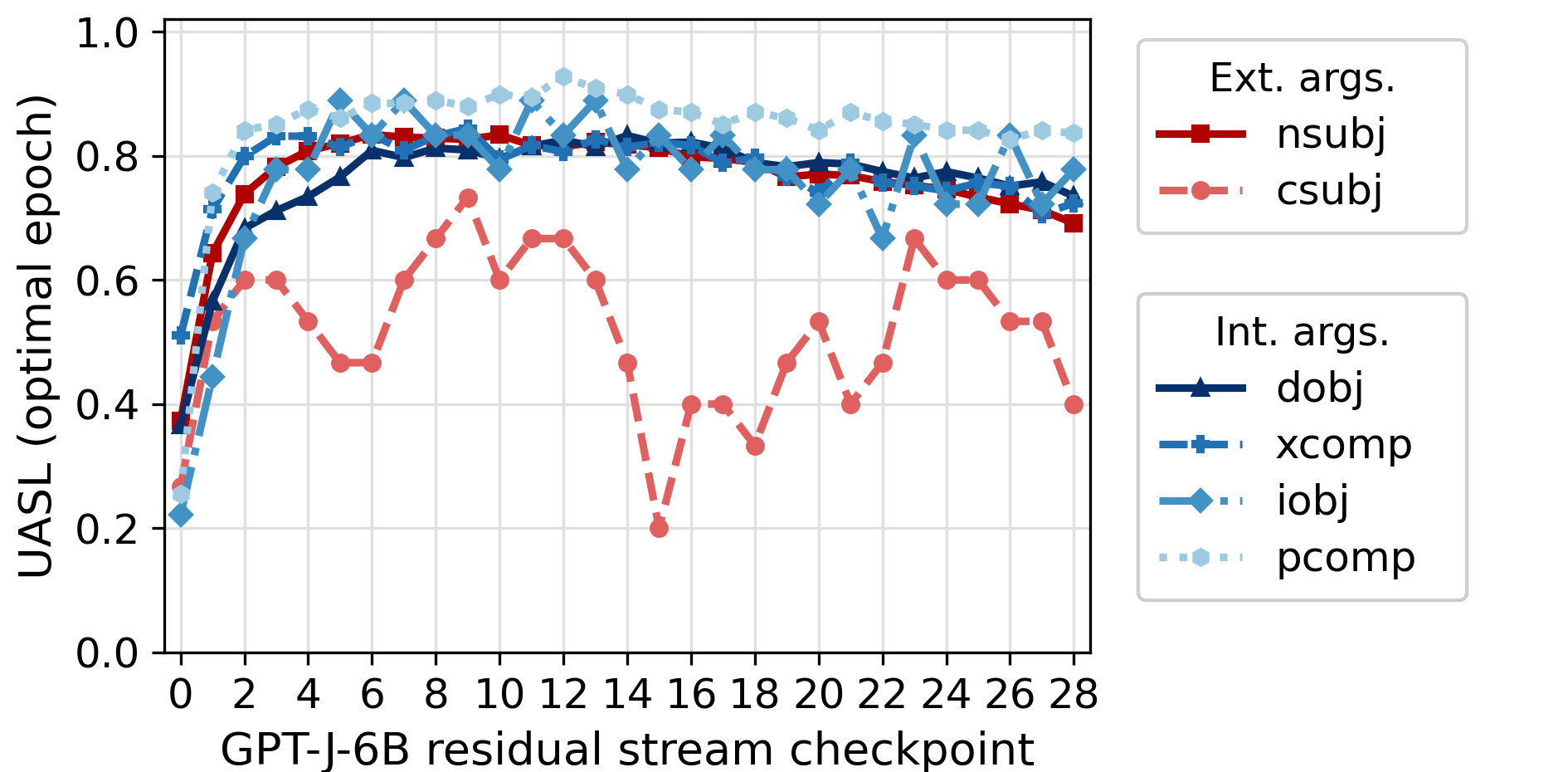}
\caption{UASL by relation at residual stream checkpoints in different models. Clockwise from the top left: DeBERTa-v3-base, ModernBERT-base, GPT-J-6B, GPT-2-base. Compare Figure \ref{fig:dependencies1}.}
\label{fig:app-selected}
\end{figure*}

The analyses discussed in the main body of the text largely focus on the canonical BERT-base model. However, in order to verify that our results are not specific to this one older, smaller model, we repeat our analyses on DeBERTa-v3-base, ModernBERT-base, GPT-2-base \citep{radford-etal-2019-gpt2}, and GPT-J-6B \citep{gpt-j}; see Table \ref{tab:models}. We reran all analyses using the same alignments, training/dev/test splits, hyperparameters, and residual stream checkpoints.

Figure \ref{fig:app-selected} is the analogue of Figure \ref{fig:dependencies1}; it charts UASL evolution over checkpoints for verbs' external and internal arguments. Remarkably, we can see that the general order is preserved: \textit{pcomp} is the most reliably parsed dependency relation, and \textit{csubj} is the least. \textit{Nsubj}, \textit{dobj}, and \textit{xcomp} track closely together, just as they had in Figure \ref{fig:dependencies1}.

\clearpage
\section{Linear distance: decay of UASL and clustering of relations}

\subsection{Models' UASL as a function of linear distance}

\begin{figure}[b]
\includegraphics[width=\columnwidth]{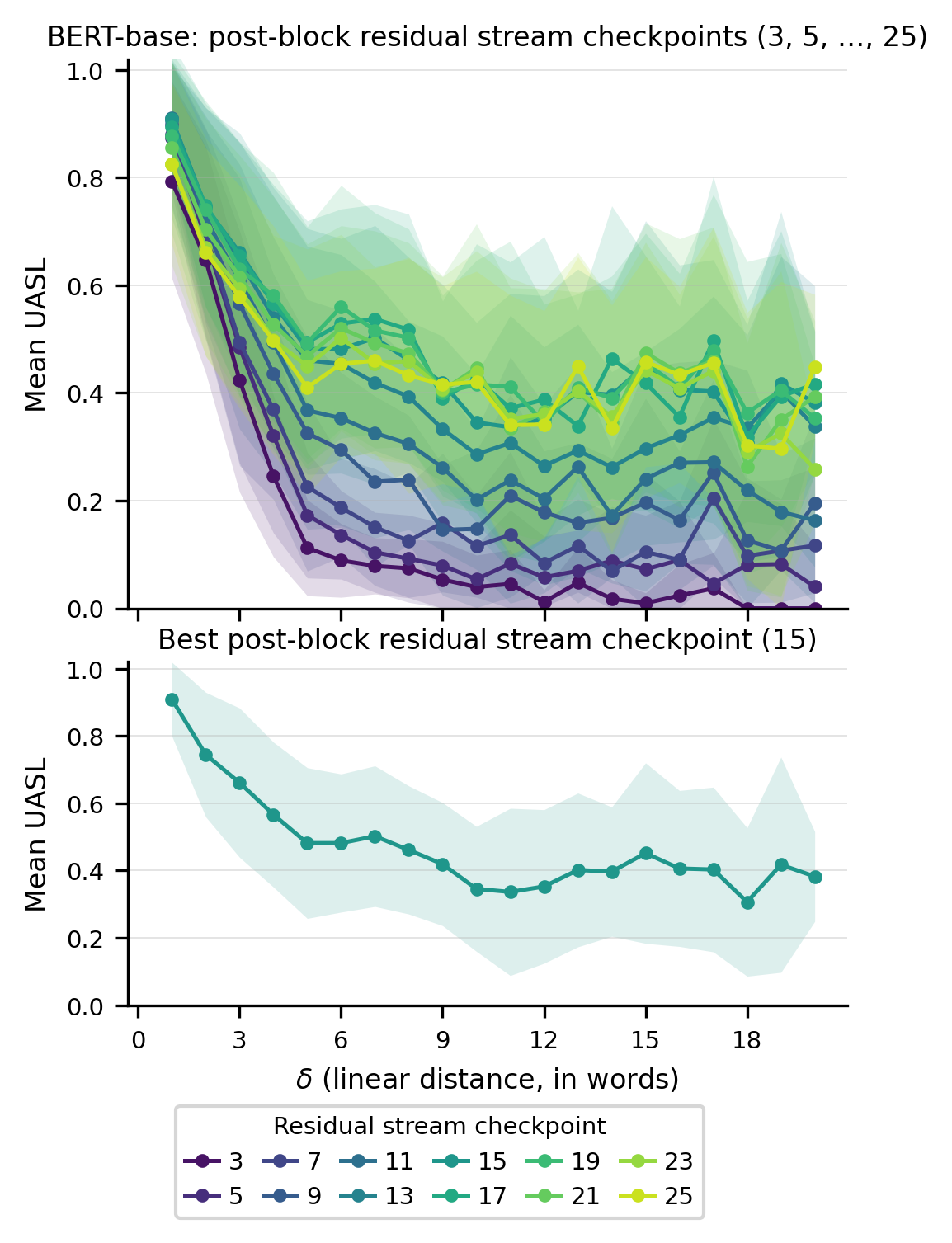}
  \caption{Average UASL and standard deviation of per-relation averages, as a function of linear distance between endpoints of a dependency edge, at all odd residual stream checkpoints (after each attention block) and at residual stream checkpoint 16 with optimal UUAS.}
  \label{fig:performance-as-a-function-of-distance}
\end{figure}

\begin{figure*}[!htbp]
\centering
\includegraphics[width=0.455\textwidth]{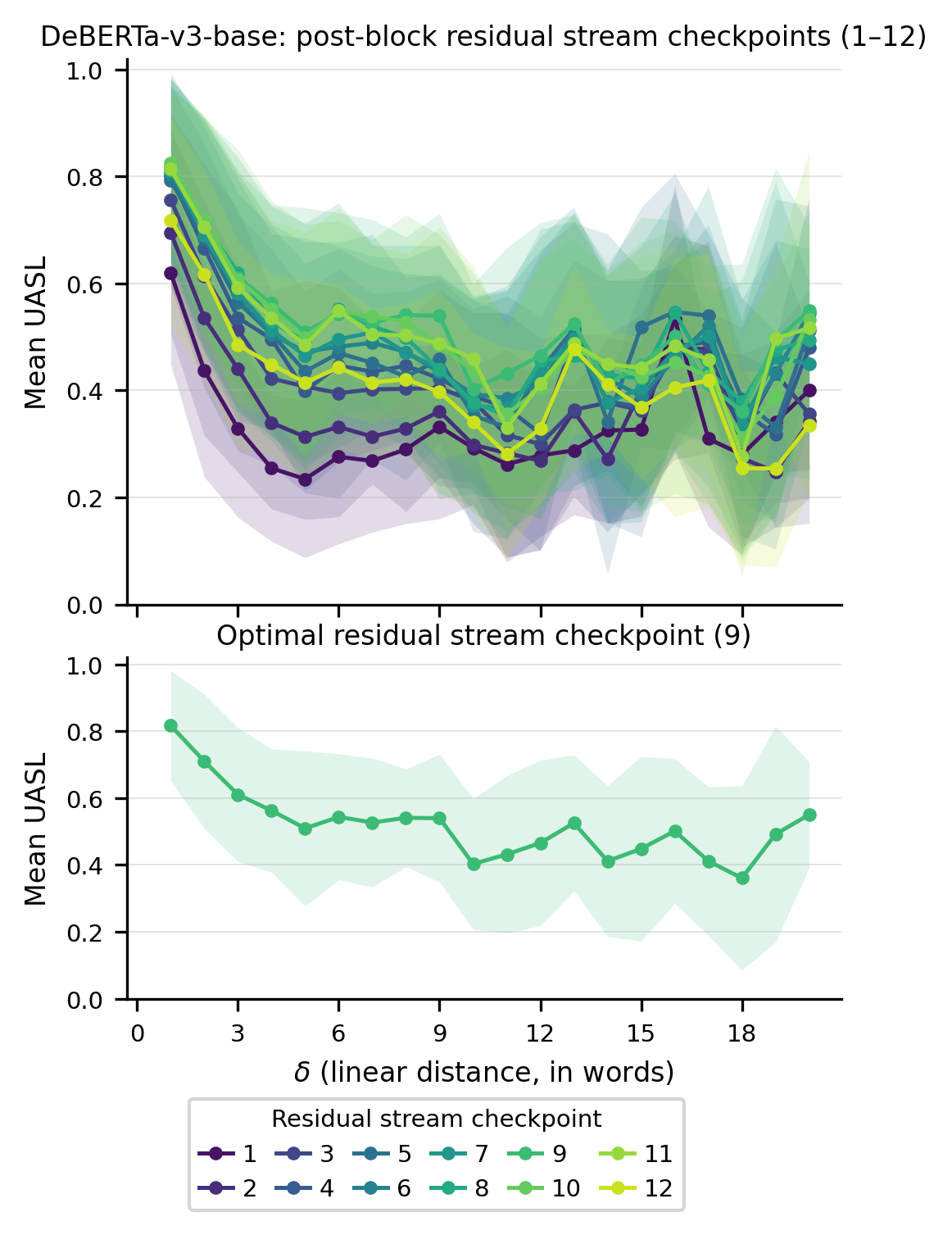}\hfill
\includegraphics[width=0.455\textwidth]{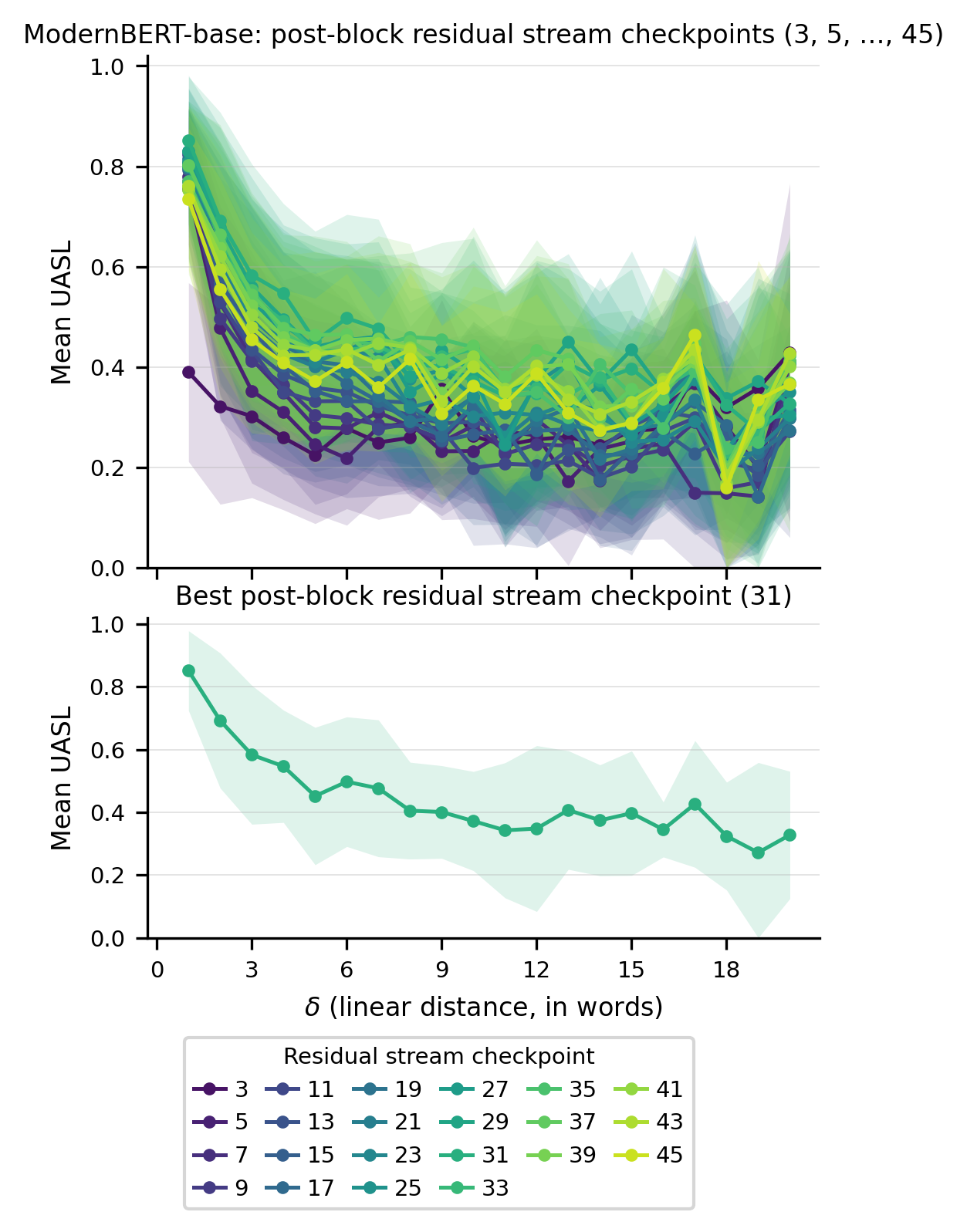}\\[0.7em]
\includegraphics[width=0.455\textwidth]{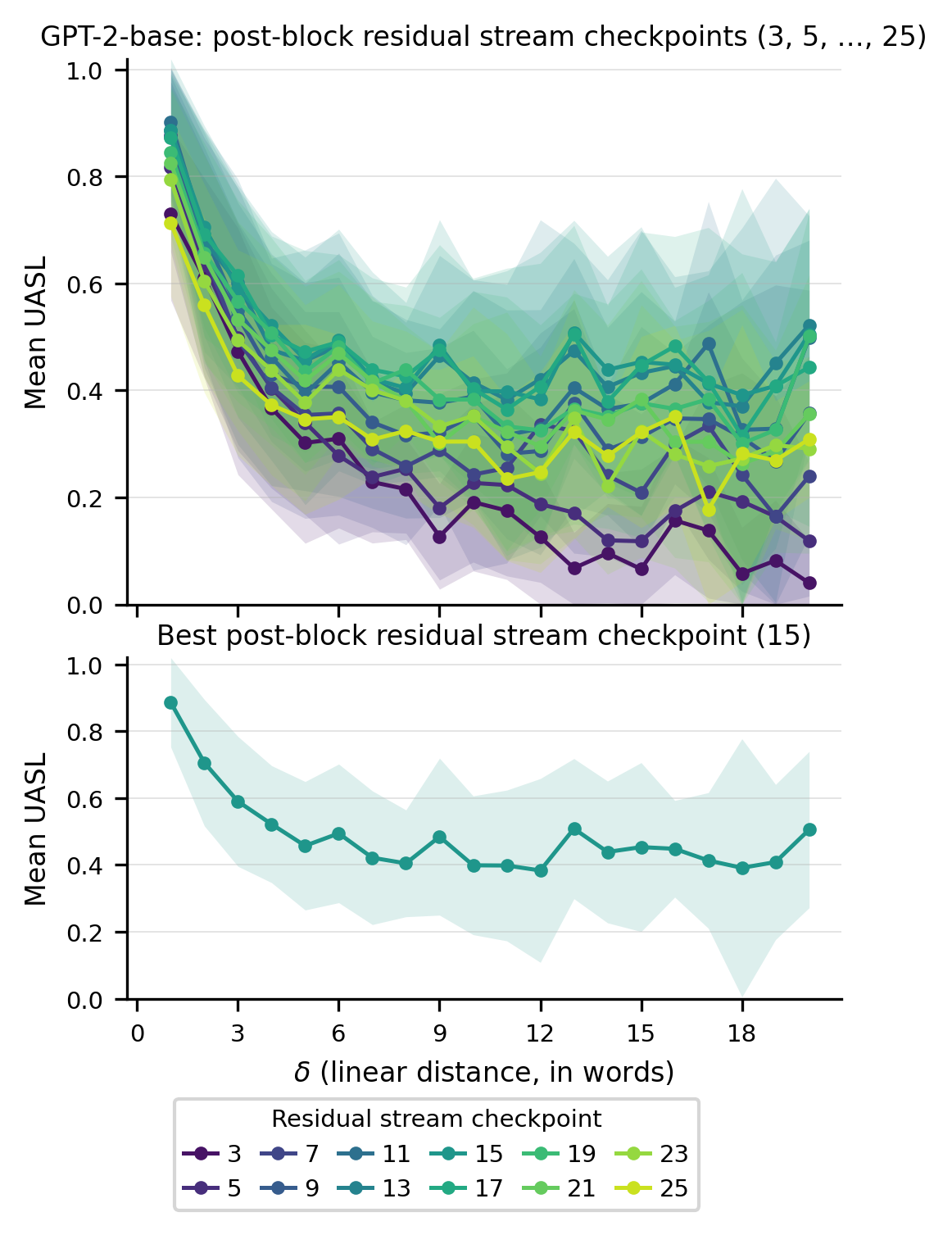}\hfill
\includegraphics[width=0.455\textwidth]{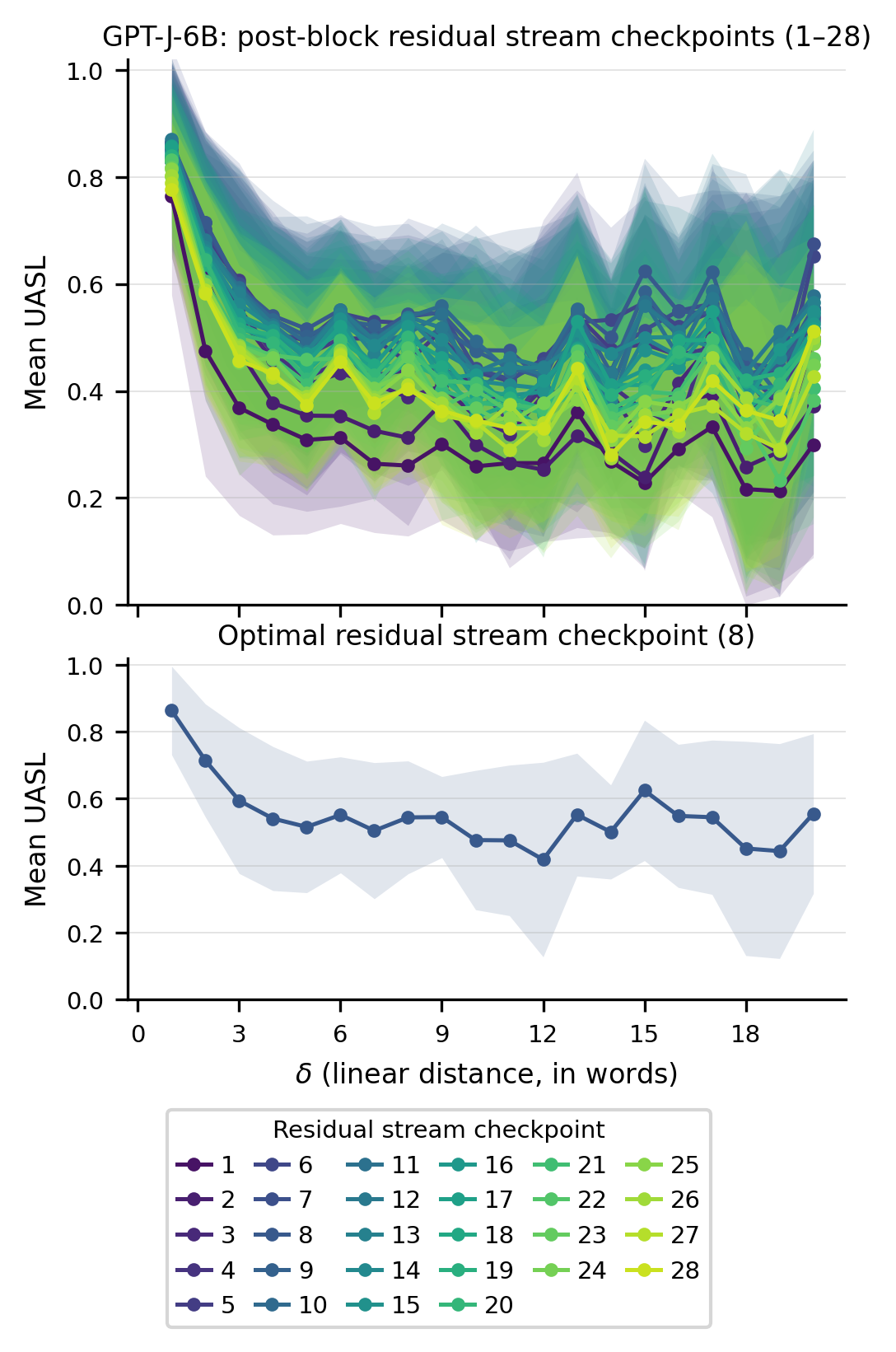}
\caption{Mean UASL shown as a function of the linear distance for different models. Shaded bands represent one standard deviation across relations. Clockwise from the top left: DeBERTa-v3-base, ModernBERT-base, GPT-J-6B, GPT-2-base. In each case the upper part of the figure shows every checkpoint, while the lower part focused on the best one.}
\label{fig:app-curves}
\end{figure*}

Figures \ref{fig:performance-as-a-function-of-distance} and \ref{fig:app-curves} illustrate the mean UASL curves as a function of linear distance, for each model at every post-block residual stream checkpoint. All models report similar performance decay as linear distance increases, further confirming the findings of \citet{Simon2025}.

\begin{figure}[b]
  \includegraphics[width=\columnwidth]{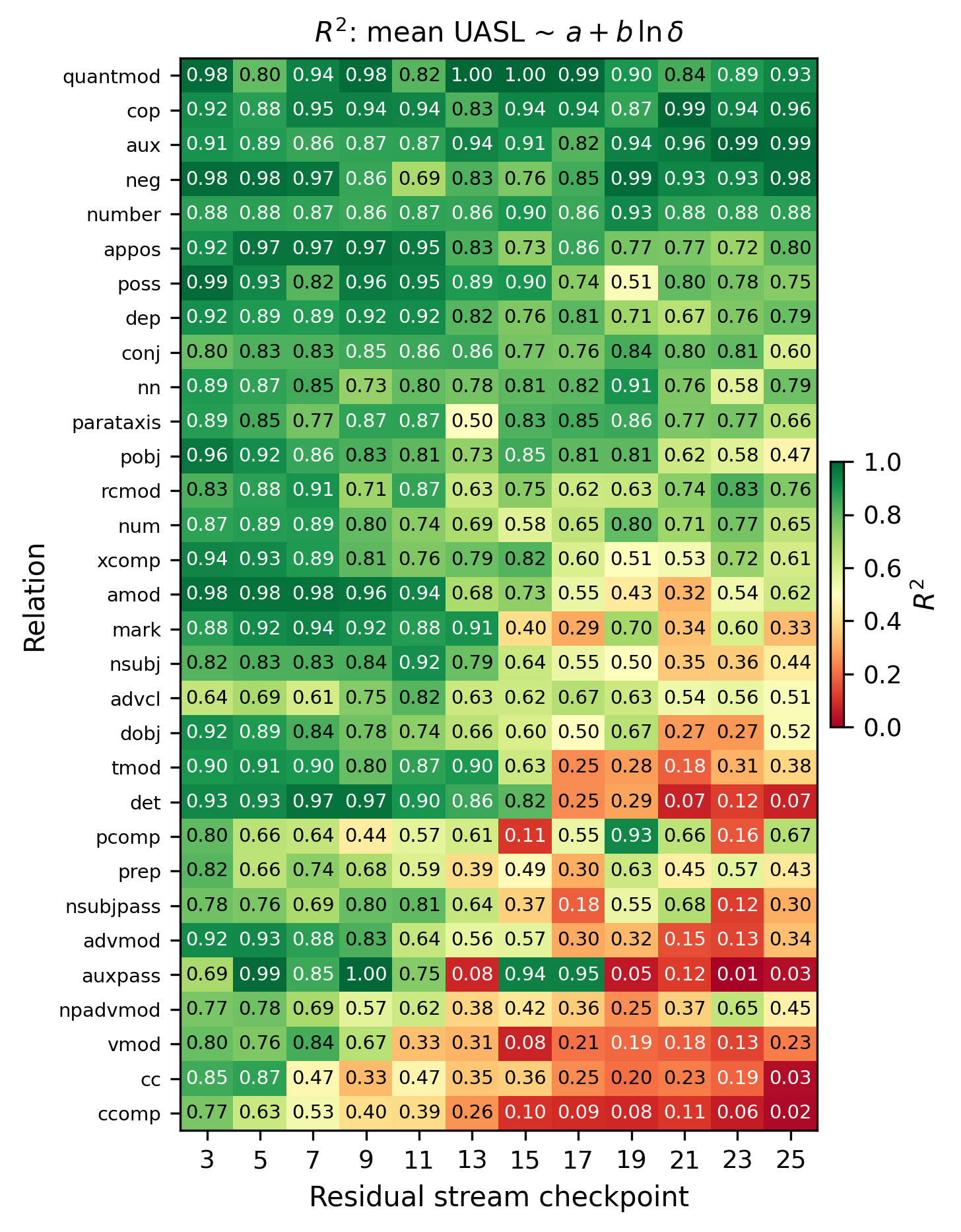}
  \caption{Fitness of a log-linear decay model across dependency relations for BERT-base, with residual stream evaluated after each full transformer block, and at residual stream checkpoint 16. The colour scale runs from red ($R^2 = 0$) through yellow to green ($R^2 = 1$). Omitted are relations for which a linear regression could not be fit. }
  \label{fig:R2-all-relations}
\end{figure}

\begin{figure*}[t]
  \centering
  \includegraphics[width=\textwidth]{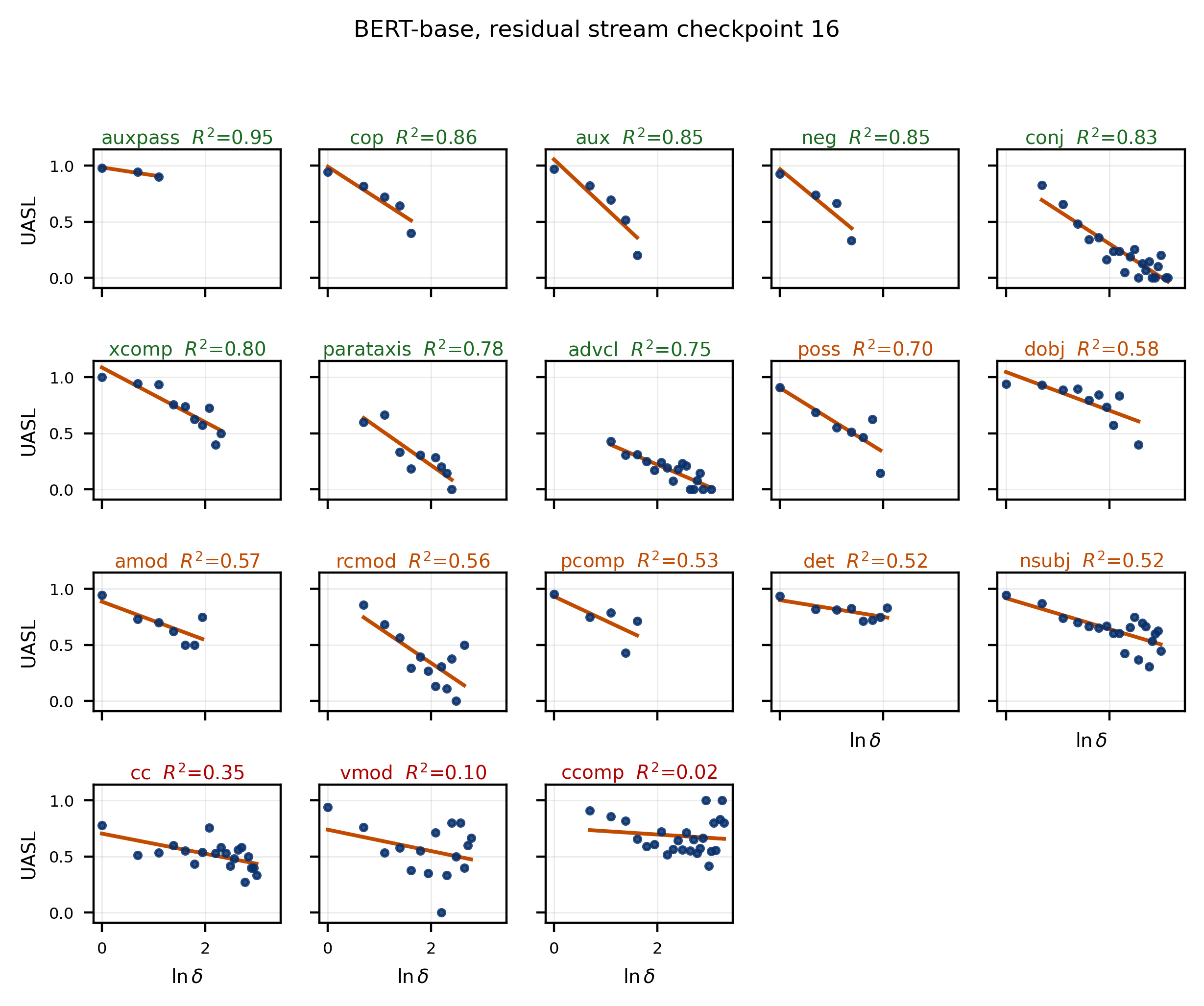}
  \caption{Individual dependency relation data by log linear distance, underlying Figure \ref{fig:R2-log-distance-model}. Datapoints are UASL values for one linear distance $\delta$ with at least five gold edges in the test set. Probing is done at residual stream checkpoint 16, where BERT-base attains its peak UUAS and which was used to fit the regression of \S\ref{EntropySec}. The line is the least-squares fit of Equation \ref{regrUASL}. A low $R^{2}$ need not indicate noise: \textit{ccomp} spans the widest range of linear distances and is close to flat, so its near-zero $R^{2}$ reflects the absence of decay rather than scatter about one. \textit{auxpass}, at the head of the ordering, is fitted on three points.}
  \label{fig:ulas-vs-log-distance}
\end{figure*}

    \begin{figure*}
\centering
\includegraphics[width=0.475\textwidth]{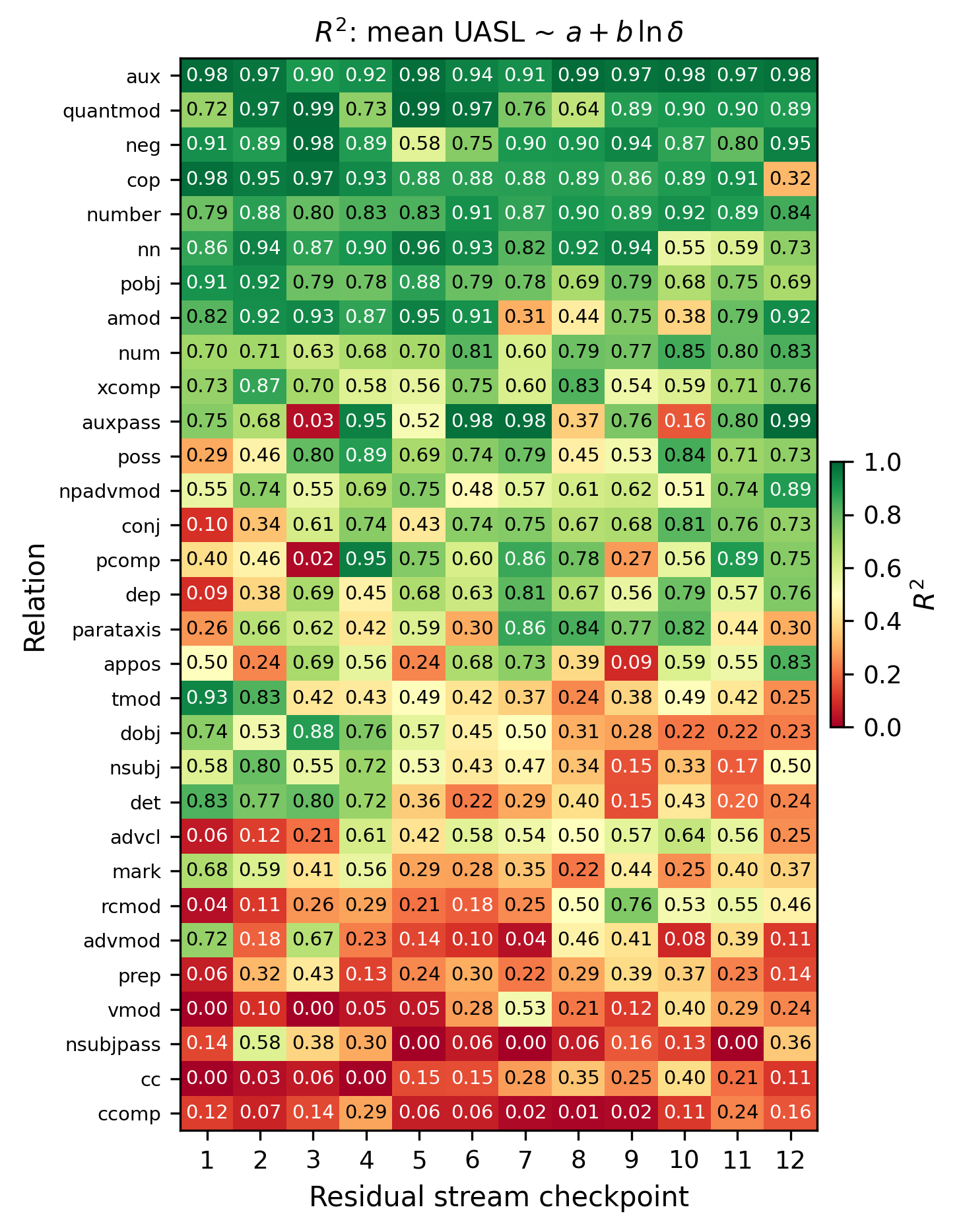}\hfill
\includegraphics[width=0.475\textwidth]{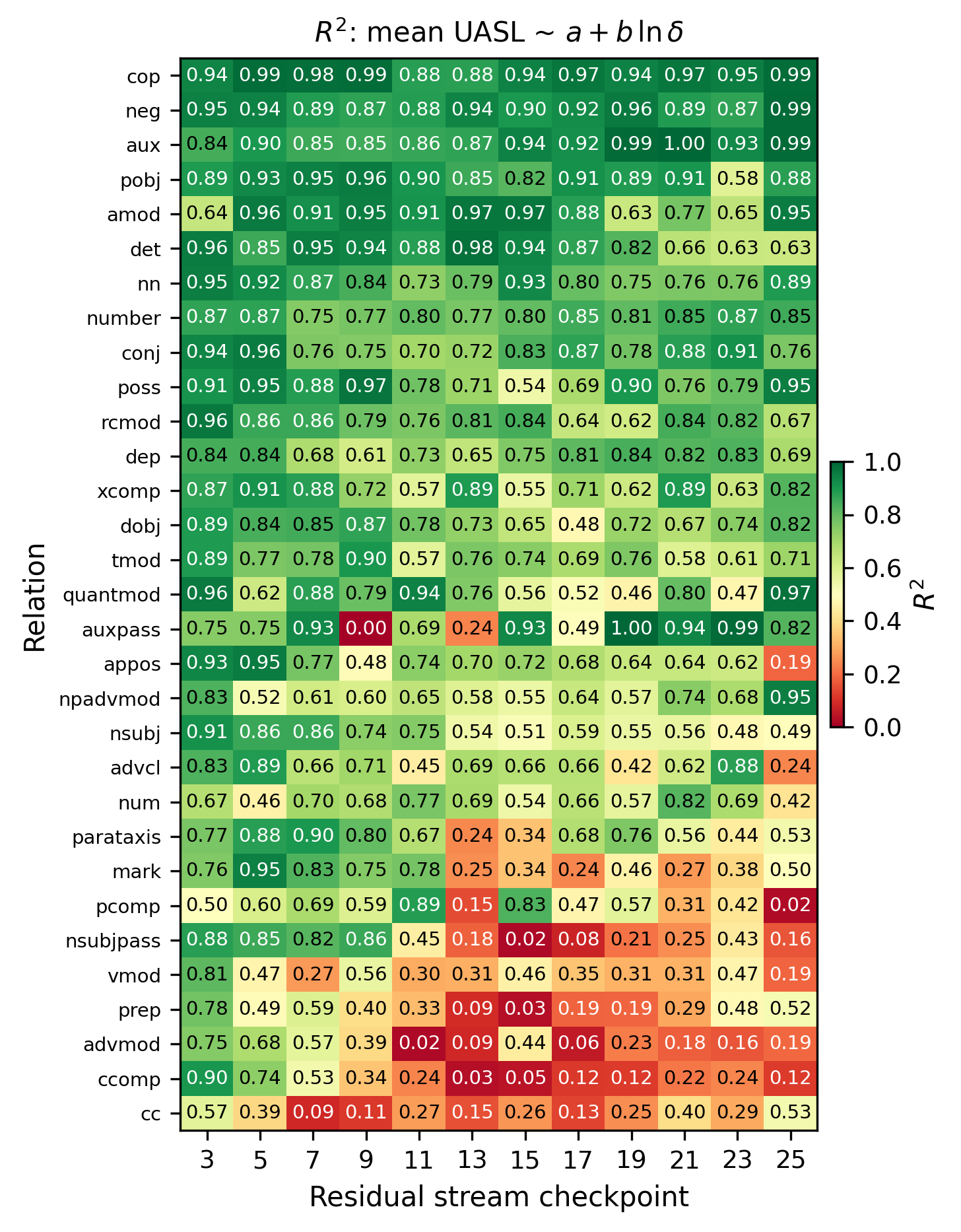}
\caption{Goodness of fit of the log-linear decay model, for DeBERTa-v3-base (Left) and GPT-2-base (Right), both models involving twelve transformer blocks.  In both cases a column is the residual stream after an entire transformer block; the two differ in checkpoint indices only because GPT-2 is stored with a post-attention checkpoint interleaved between consecutive post-MLP ones. Rows are the relations, ordered by mean $R^{2}$ across residual stream checkpoints.}
\label{fig:app-r2a}
\end{figure*}

\begin{figure*}
\centering
\includegraphics[width=0.73\textwidth]{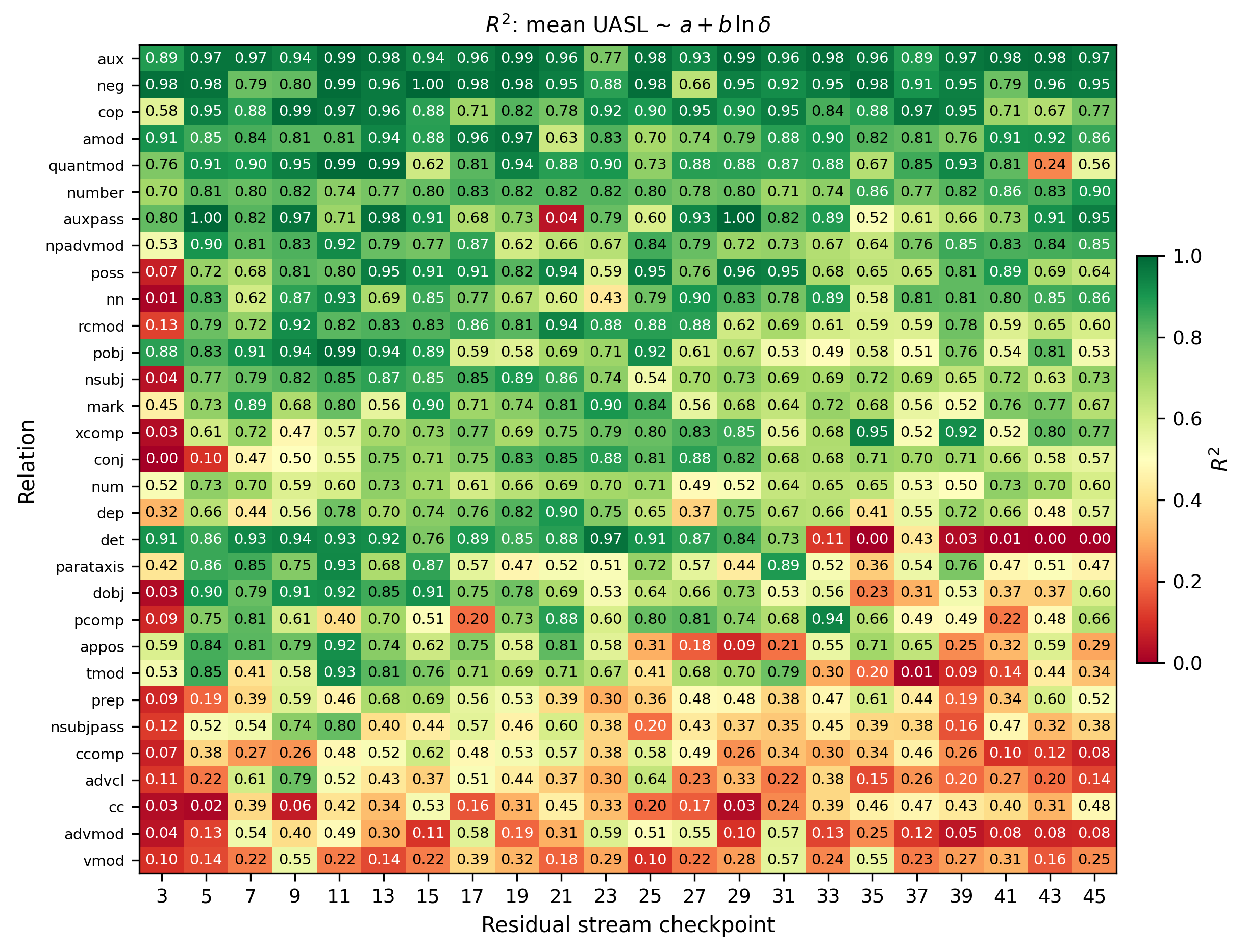}\\[0.8em]
\includegraphics[width=0.90\textwidth]{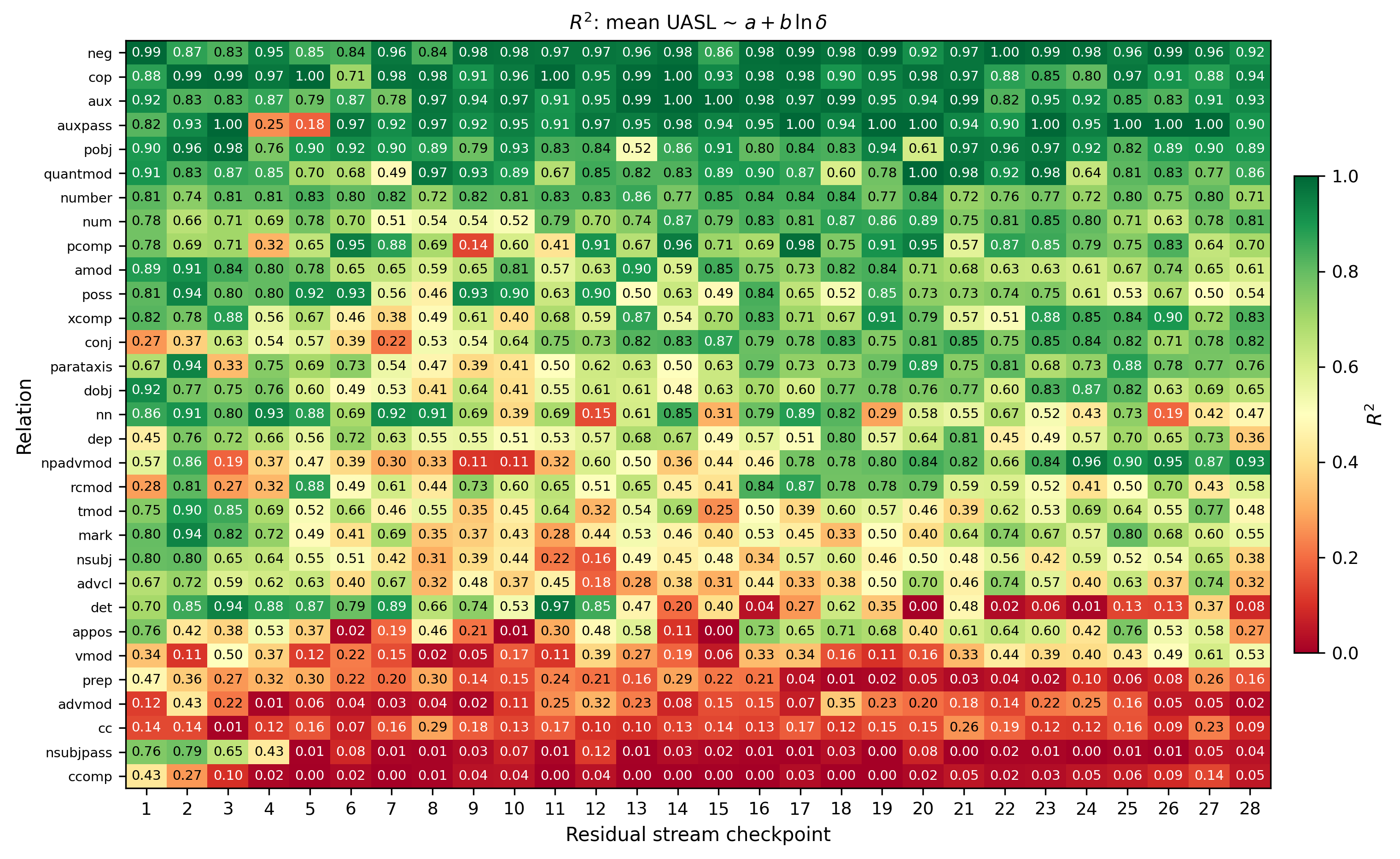}
\caption{Goodness of fit for the log-linear decay model in the case of ModernBERT-base (top) and GPT-J-6B (bottom), respectively with columns with residual stream checkpoints $3, 5, \ldots, 45$, after each of the 22 blocks, and residual stream checkpoints $1$--$28$, after each of 28 blocks. Rows are as in Figure \ref{fig:app-r2a}.}
\label{fig:app-r2b}
\end{figure*}

In turn, Figures \ref{fig:R2-all-relations}, \ref{fig:app-r2a} and \ref{fig:app-r2b} extend the $R^2$ heat maps  of the log-linear decay model (see Figure \ref{fig:R2-log-distance-model}) across all relations and all models. We  find that the median $R^2$ across (relation, residual stream checkpoint) pairs is 0.777 for BERT-base, 0.752 for GPT-2-base, 0.692 for ModernBERT-base, 0.648 for GPT-J-6B, and 0.588 for DeBERTa-v3-base. This confirms  \S\ref{sec:logdistsec}'s assessment of the log-linear model as a reasonable but imperfect metric that does not describe all dependency relations well for any of the models.%

\subsection{The range of a relation and the composite distance}
\label{sec:appendix-all-dendros}

\begin{table}[b]
\centering
\caption{For each dependency relation, the table displays the value of $\delta_{90}(r)$: the 90th percentile of its linear distances over the corpus, the linear distance being the number of words separating the head and the dependent (so that $1$ denotes adjacent words).}
\begin{tabular}{lr@{\hspace{1.25em}}|@{\hspace{1.25em}}lr}
\toprule
\textbf{Relation} & \textbf{Value} & \textbf{Relation} & \textbf{Value} \\
\midrule
advcl      & 17 & expl       & 4 \\
ccomp      & 17 & npadvmod   & 4 \\
parataxis  & 17 & pcomp      & 4 \\
conj       & 13 & pobj       & 4 \\
nsubjpass  & 12 & poss       & 4 \\
cc         & 11 & predet     & 4 \\
csubj      & 11 & det        & 3 \\
dep        & 11 & iobj       & 3 \\
vmod       & 10  & preconj    & 3 \\
discourse  & 9  & amod       & 2 \\
mark       & 9  & aux        & 2 \\
rcmod      & 9  & auxpass    & 2 \\
tmod       & 8  & neg        & 2 \\
appos      & 7  & nn         & 2 \\
nsubj      & 7  & num        & 2 \\
prep       & 7  & number     & 2 \\
advmod     & 5  & quantmod   & 2 \\
xcomp      & 5  & mwe        & 1 \\
acomp      & 4  & possessive & 1 \\
cop        & 4  & prt        & 1 \\
dobj       & 4  &            &   \\
\bottomrule
\end{tabular}
\label{tab:arc-len-90}
\end{table}

\S\ref{sec:distances} clusters relations using only $d_{\mathrm{UASL}}$, which compares the dependency relations' UASL against their linear distance curves. Here, we define an additional range component that quantifies the linear distances of dependency relations in the corpus, and a composite distance formula. We then report dendrograms calculated with these components.

To distinguish between local, short-range, and non-local, 
long-range dependencies, we computed first, for each relation, the 
linear distance distribution over the corpus and extracted its count-weighted 90th-percentile linear distance $\delta_{90}(r)$. 
These are computed using the full gold-arc counts.
Evaluating the 90th percentile is preferable over the maximum, 
to avoid excessive dependence on a single long arc outlier.
We find a significant range, with a maximum at {\rm advcl},
{\rm ccomp}, and {\rm parataxis} with $\delta_{90}(r) =17$ and a minimum at {\rm mwe}, {\rm possessive}, {\rm prt} with $\delta_{90}(r) =1$.

Then we measured differences in the range via the formula
\begin{equation}\label{eq:composite_distance}
\begin{array}{l}
    d_{\mathrm{range}}(r_1,r_2) =
  \displaystyle{\frac{|\delta_{90}(r_1) - \delta_{90}(r_2)|}{N_{\max}}}, \\[2mm]
  N_{\max} =   \max_{r}\, \delta_{90}(r)\, , 
  \end{array}
\end{equation}
where the $\delta_{90}(r)$ are the 90th percentile linear distances shown in Table \ref{tab:arc-len-90},
and then introduced the composite quantity
\begin{multline}
    d(r_1,r_2)=\\\sqrt{    \alpha\tilde{d}_{\mathrm{UASL}}(r_1,r_2)^{2}
    + (1-\alpha)\tilde{d}_{\mathrm{range}}(r_1,r_2)^{2} }
\end{multline}
where we have divided each $d_{\bullet}$ by its standard deviation to account for their different scales.

\begin{figure*}
    \centering
    \includegraphics[width=\linewidth]{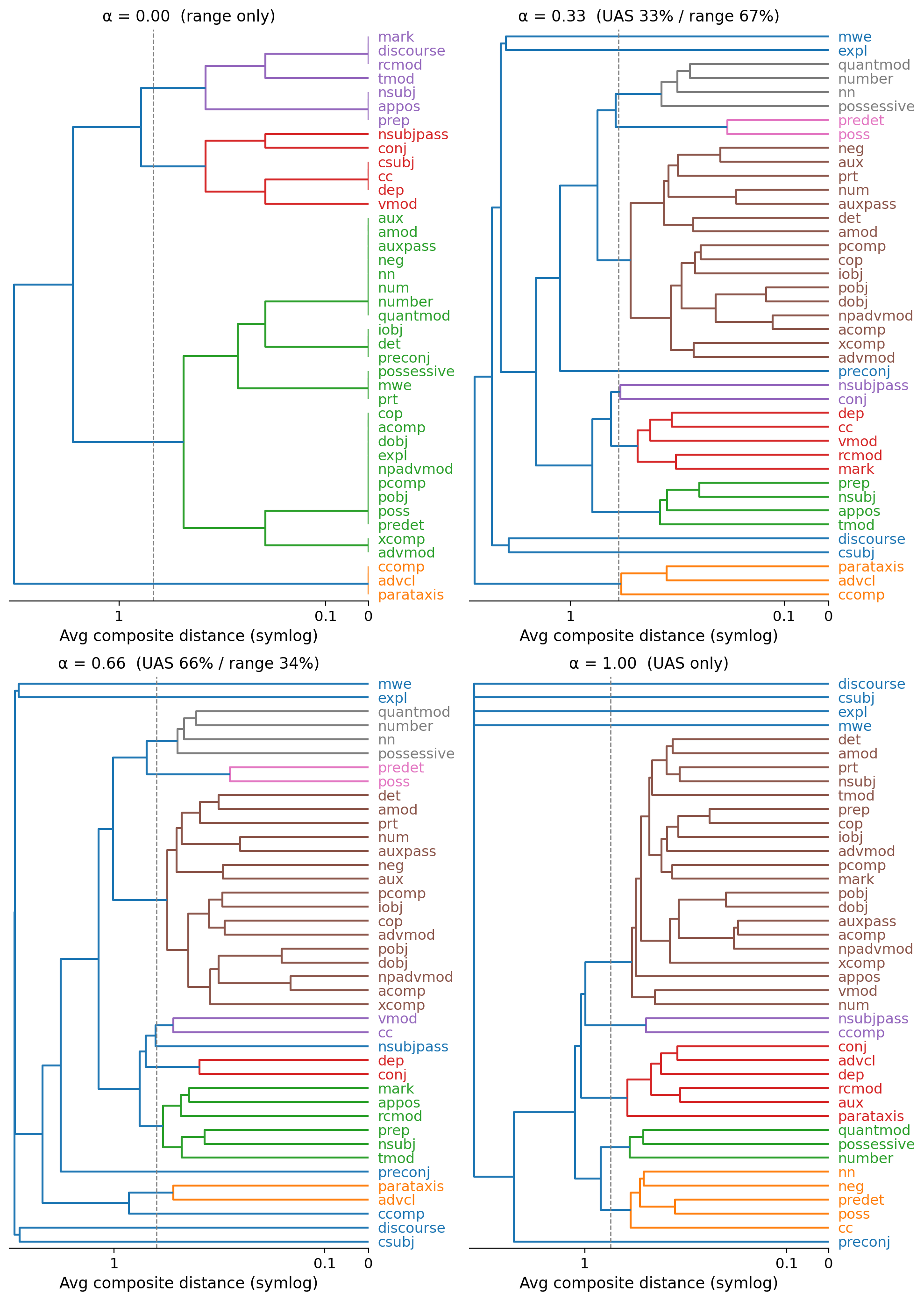}
    \caption{The four average-linkage dendrograms of the 41 dependency relations. $\tilde{d}_{\mathrm{range}}$ measures the differences between the relations' count-weighted 90th-percentile linear distance while $\tilde{d}_{\mathrm{UASL}}$ measures the probe's UASL curves as a function of linear distance. Dendrograms are computed with a variation of no arc-function UASL ($\alpha=0$, top left, range-only) to only UASL performance as a function of linear distance ($\alpha=1$, bottom right, UASL only). The horizontal axis uses a symmetric-log scale (linear below 0.3, logarithmic above), and the dashed gray vertical line marks 25\% of the tallest merge, after which leaf labels are colored to indicate their cluster affiliation.}
    \label{fig:dendrogram_relations}
\end{figure*}

In Figure \ref{fig:dendrogram_relations}, we report four dendrograms, each computed with different values of $\alpha$. Each dendrogram is reported along a symmetric-log scale horizontal axis (below 0.3, the average composite distance is linear; above this, the distance is logarithmic), and 25\% of the tallest merge height is marked with a dashed vertical line, after which leaf labels are colored to indicate their cluster affiliation.

The top left dendrogram ($\alpha=0$) is calculated using only the corpus's linear distances, which ends up clustering the three longest-range relations (\textit{advcl/ccomp/parataxis} where $\delta_{90} = 17$) from the rest of the dependency relations. Because this dendrogram uses only linear distances, intermediate-range dependency relations (\textit{mark/discourse/rcmod/tmod/nsubj/appos/prep}) cluster together as their linear distances can be seen to range from 7--9 in Table \ref{tab:arc-len-90}.

In the bottom right dendrogram ($\alpha=1$), dependency relations are clustered using $\tilde{d}_{\mathrm{UASL}}$, which is the measure developed and utilized throughout this paper as it contextualizes the probe's UASL curves as a function of linear distance.

\subsection{Wasserstein-1 distance}
\label{app:wasserstein}

\begin{figure*}[p]
    \centering
    \includegraphics[width=\linewidth]{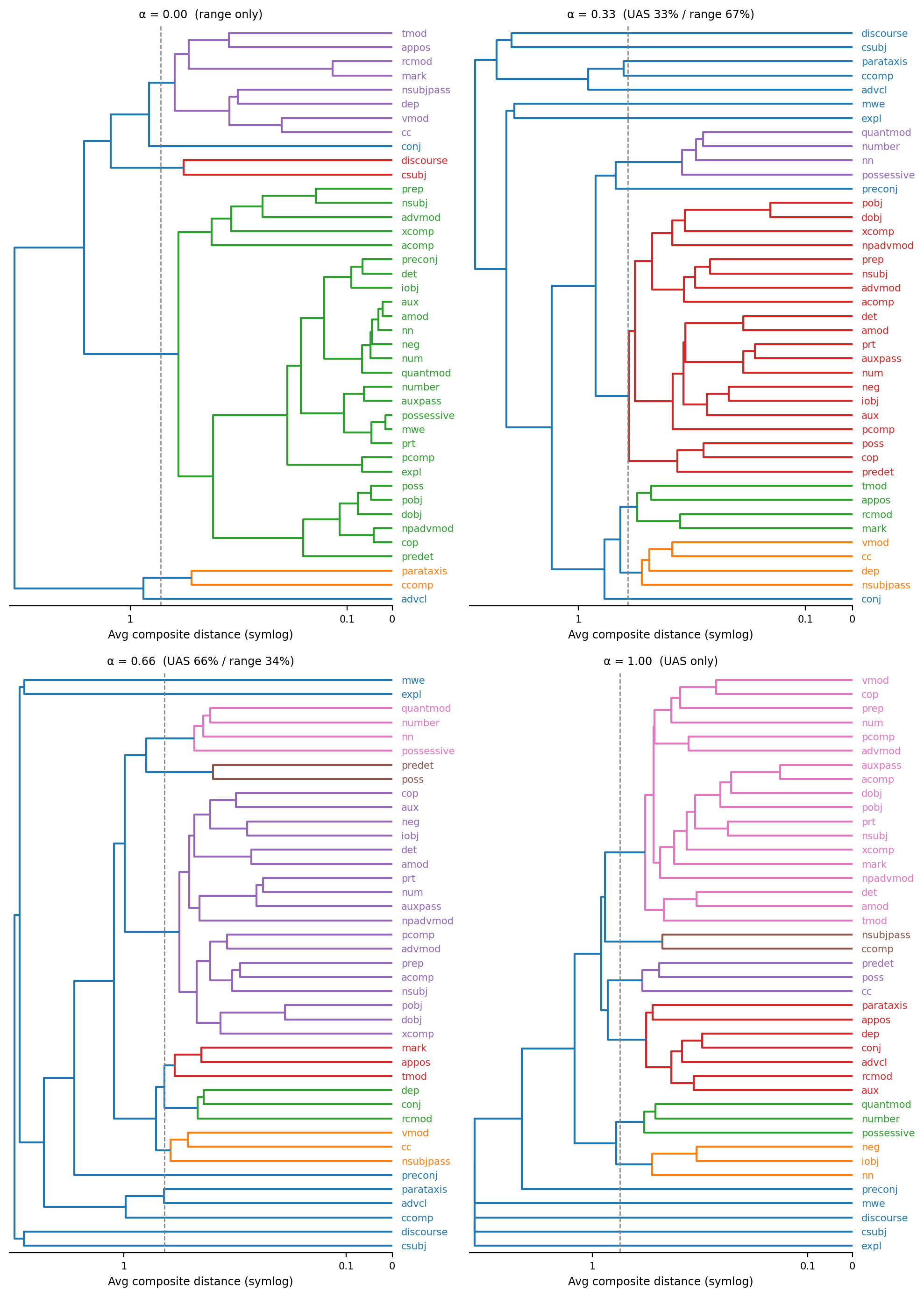}
    \caption{Like Figure \ref{fig:dendrogram_relations}, the above dendrograms of the 41 dependency relations measure the differences between the relations' count-weighted 90th-percentile linear distance. However, unlike Figure \ref{fig:dendrogram_relations}, the range component has been replaced with the Wasserstein-1 distance described in Equation \ref{eq:wasserstein_range}. All other components are the same as in Figure \ref{fig:dendrogram_relations}: $\alpha=1$ (bottom right) is a dendrogram computed only with respect to UASL, while $\alpha=0$ (top left) is computed only with the new Wasserstein-1 range component, and the horizontal axis and gray vertical line correspond to that in Figure \ref{fig:dendrogram_relations}.}
    \label{fig:dendrogram_w1}
\end{figure*}

An unsatisfactory aspect of the range component we discussed above is that relations with differently distributed linear distance can have distance zero, due to the use of the 90th-percentile linear distance (i.e. $d_{\rm range}$ is not a metric). To avoid this problem one can instead directly compare 
linear distance \emph{distributions}. Let 
$p_r(\delta)$ denote the relative frequency of linear distance $\delta$ among the gold arcs of relation $r$, and let $F_r$ be the
resulting distribution function. We consider the
Wasserstein-1 ($W_1$) distance
\begin{equation}\label{eq:wasserstein_range}
d^{W}_{\mathrm{range}}(r_1,r_2) = \sum_{\delta \ge 1} \left| F_{r_1}(\delta) - F_{r_2}(\delta) \right|.
\end{equation}
Unlike the percentile version previously discussed, this is a metric on the space of distributions. 
For example, \textit{cc} and \textit{csubj} both have $\delta_{90}=11$ but have $3.07$ words separation in the Wasserstein-1 metric, while \textit{advcl} and \textit{ccomp} both have $\delta_{90}=17$ but differ by $2.22$ words in Wasserstein-1 (see the $\alpha=0$ panel of Figure \ref{fig:dendrogram_w1}). Despite $W_1$ being more mathematically justified in this sense, the quantities $d_{\rm range}$ and $d_{\rm range}^W$ correlate at $\rho = 0.91$ across the 820 pairs of dependency relations. Moreover, the use of the $W_1$ distance does not make the resulting clustering significantly more interpretable: for intermediate values of $\alpha$ the clustering obtained using the $W_1$ metric appears more fragmented, with for example nine and eight singletons, respectively, for $\alpha = 0.33$ and $0.66$, against just five and six, and one sees a single very large cluster at 20 and 18 relations, instead of the argument-versus-function-word split that we observed. Figure \ref{fig:dendrogram_w1} shows four panels for comparison.

\subsection{Clustering of relations by UASL-vs-linear-distance profile}

\begin{figure*}[p]
\centering
\includegraphics[width=0.48\textwidth]{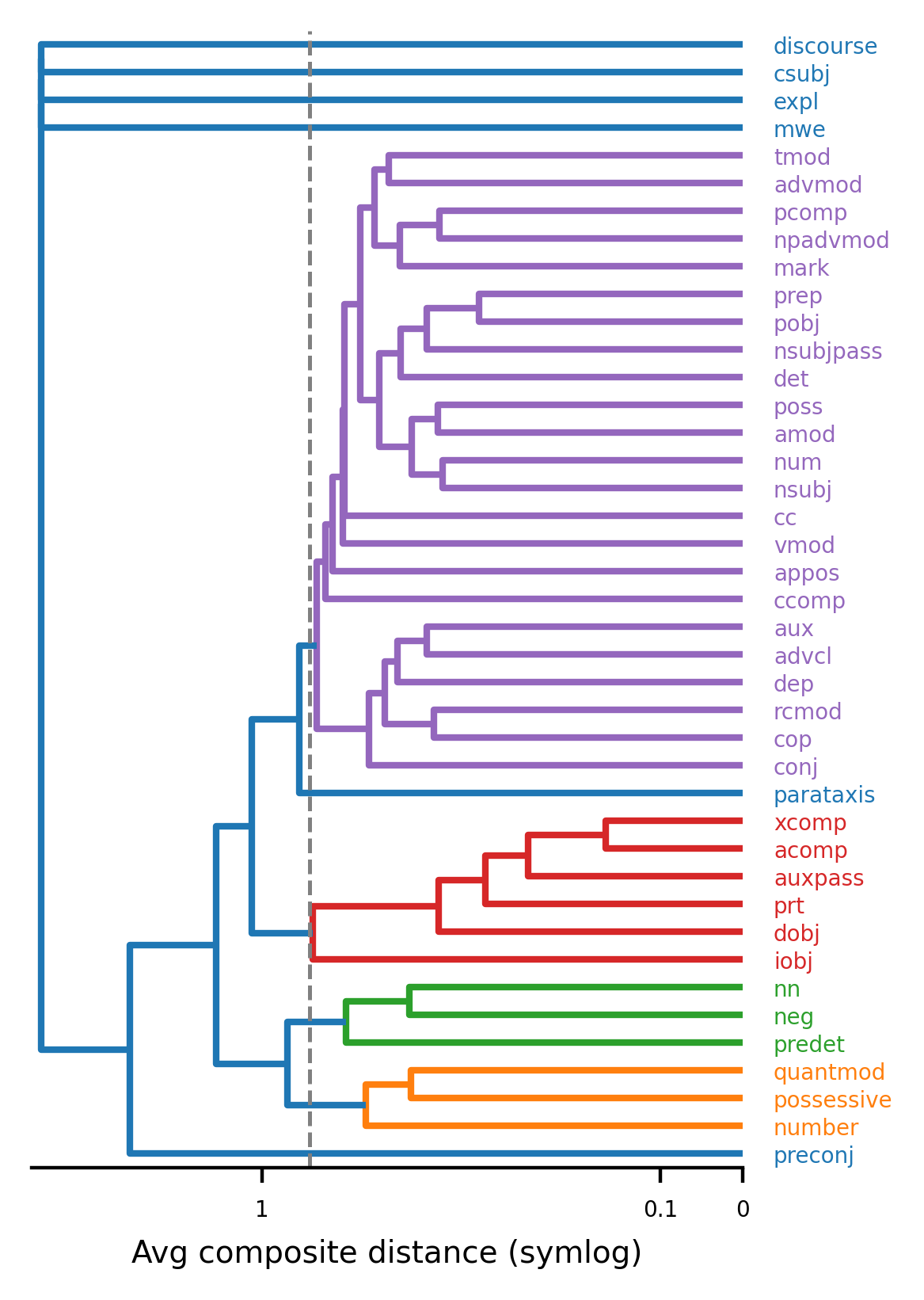}\hfill
\includegraphics[width=0.48\textwidth]{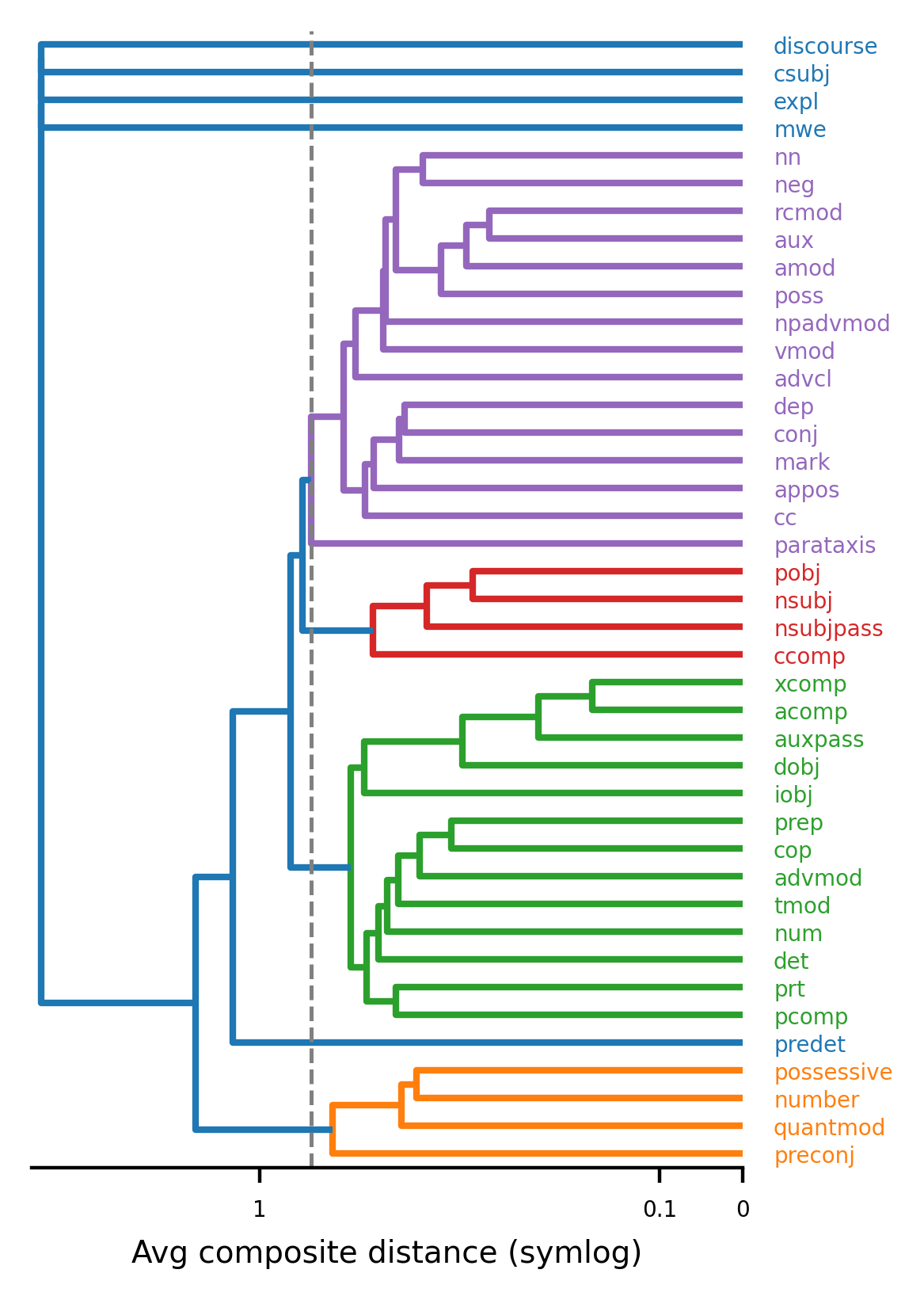}\\[0.5em]
\includegraphics[width=0.48\textwidth]{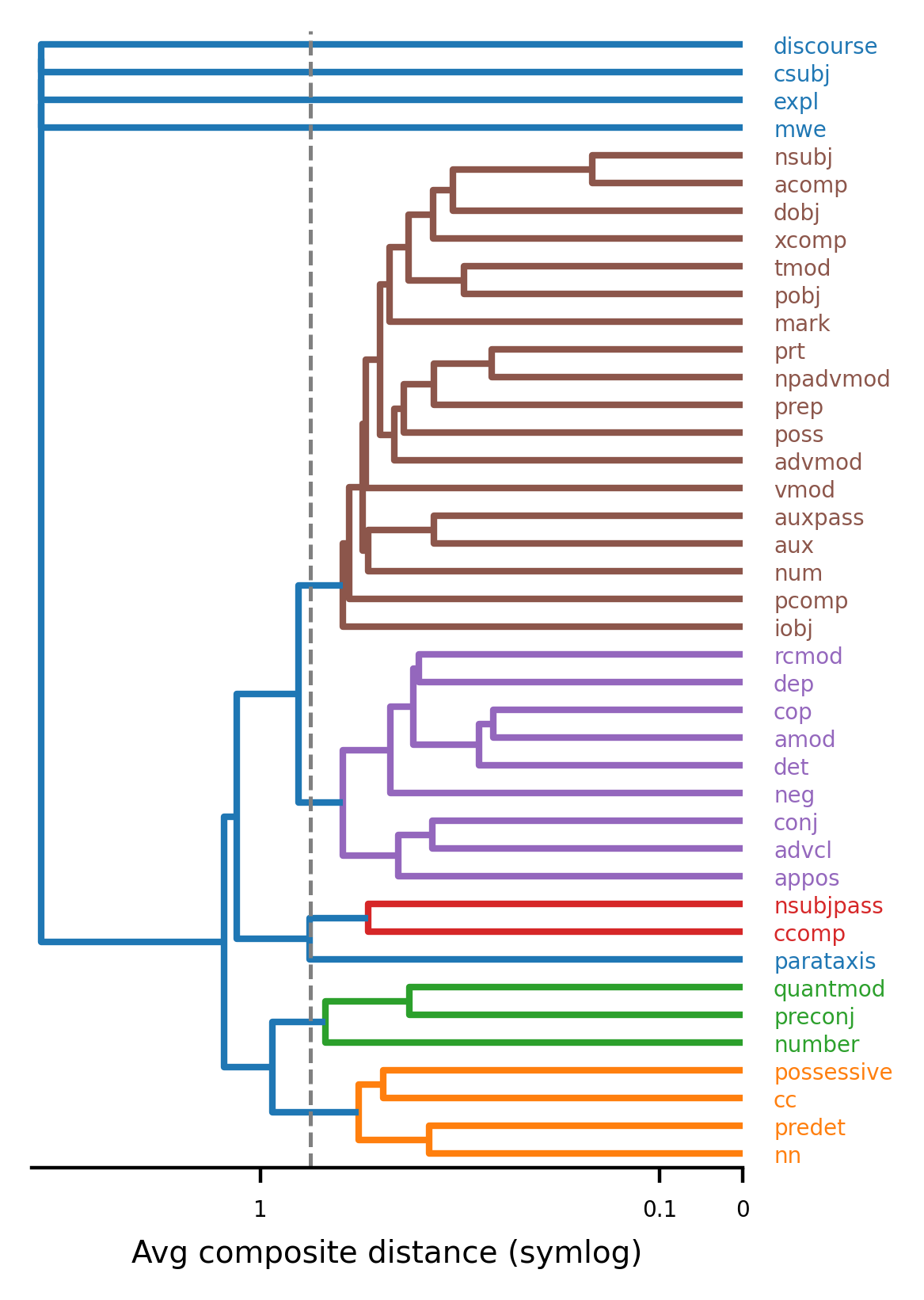}\hfill
\includegraphics[width=0.48\textwidth]{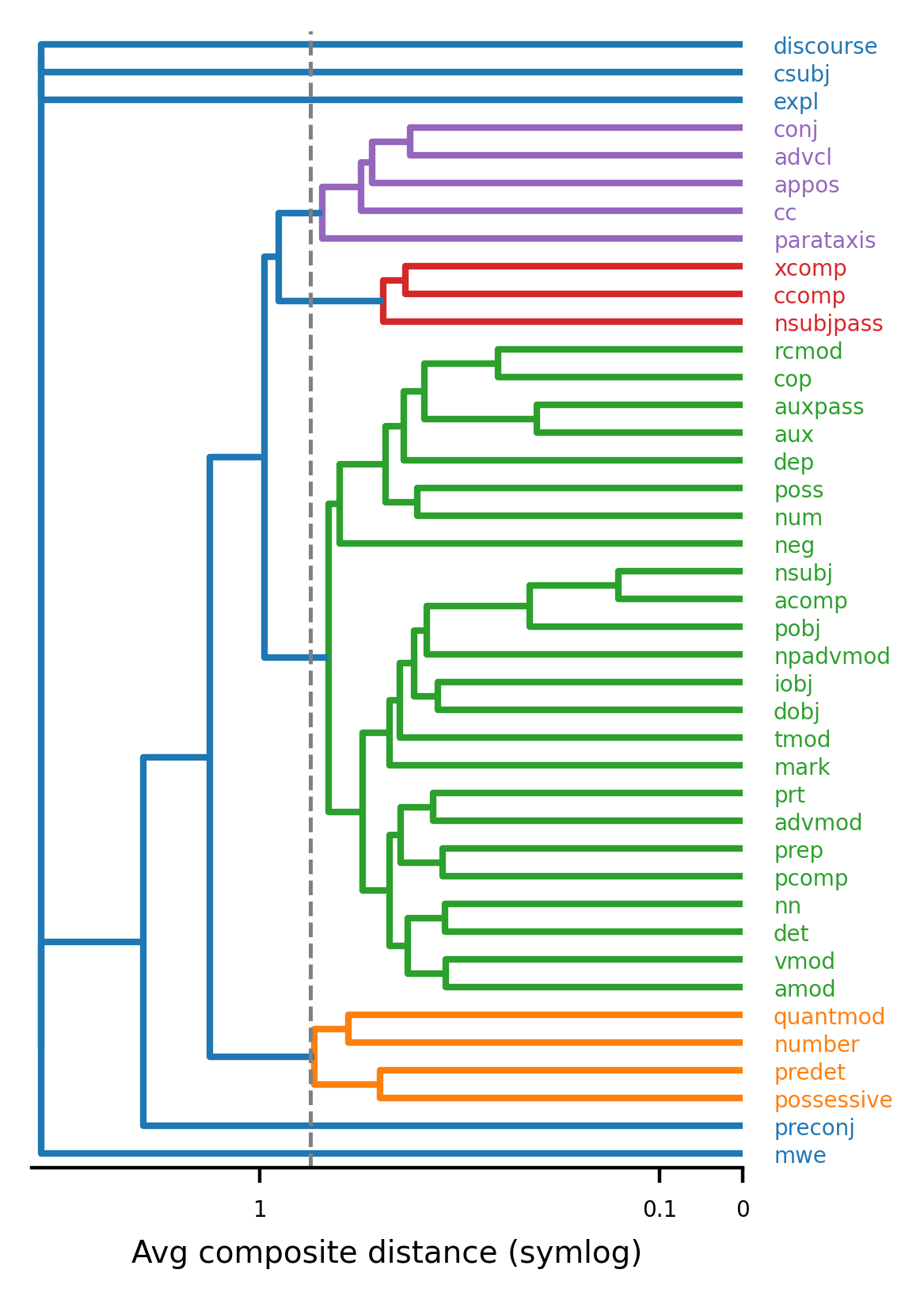}
\caption{Model comparison: DeBERTa-v3-base, ModernBERT-base, GPT-J-6B, GPT-2-base (clockwise from top-left). The average-linkage dendrograms of the 41 dependency relations, clustered by UASL, as a function of linear distance ($\alpha=1$ in the composite distance of Figure \ref{fig:dendrogram_relations}), measured over all residual stream checkpoints of the test set. The horizontal axis is a symmetric-log scale, linear below 0.3 and logarithmic above; the dashed grey vertical line marks 25\% of the tallest merge, past which leaf labels are coloured by cluster. The four panels share these settings, so they are stated here once rather than repeated as a title above each.}
\label{fig:app-dendro}
\end{figure*}

The UPGMA clustering depicted in Figure \ref{fig:UASL-only-dendrogram} is repeated in Figure \ref{fig:app-dendro} for each model. We again observe a broad clustering of function-word dependents (similar to the yellow-green cluster in BERT-base), though there is a bit more mixing of closed and open class.

\clearpage

\newpage

\section{Regression results across models}
\label{app:regression}

\begin{table*}[b]
\centering

\small
\setlength{\tabcolsep}{4.5pt}
\begin{tabular}{lrrrrrr}
\toprule
\textbf{Model} & \textbf{Peak UUAS} & \textbf{ckpt} & \textbf{$R^2$} & \textbf{head sim-entropy} & \textbf{mean log length} & \textbf{sd log length} \\
\midrule
BERT-base        & 0.815 & 16 & 0.736 & $-0.082^{***}$ & $-0.252^{***}$ & $+0.163^{*}$ \\
DeBERTa-v3-base  & 0.788 &  9 & 0.609 & $-0.067^{***}$ & $-0.226^{***}$ & $+0.186^{*}$ \\
ModernBERT-base  & 0.762 & 30 & 0.700 & $-0.056^{***}$ & $-0.216^{***}$ & $+0.103^{\phantom{*}}$ \\
GPT-2-base       & 0.776 & 16 & 0.744 & $-0.076^{***}$ & $-0.218^{***}$ & $+0.143^{*}$ \\
GPT-J-6B         & 0.783 &  8 & 0.746 & $-0.069^{***}$ & $-0.193^{***}$ & $+0.125^{*}$ \\
\bottomrule
\end{tabular}
\caption{Peak UUAS and WLS regression of \S\ref{EntropySec}, evaluated for each model at its own optimal residual stream checkpoint. The symbol $^{***}$ denotes $p<.001$, while $^{*}$ denotes $p<.05$. The dispersion term is marginal for ModernBERT-base ($p=.095$) and significant in all the other cases. All three effects discussed in the main text occur in each model with the same sign and with comparable magnitude.}
\label{tab:other-models}
\end{table*}

\begin{figure*}[t]
\centering
\includegraphics[width=0.92\textwidth]{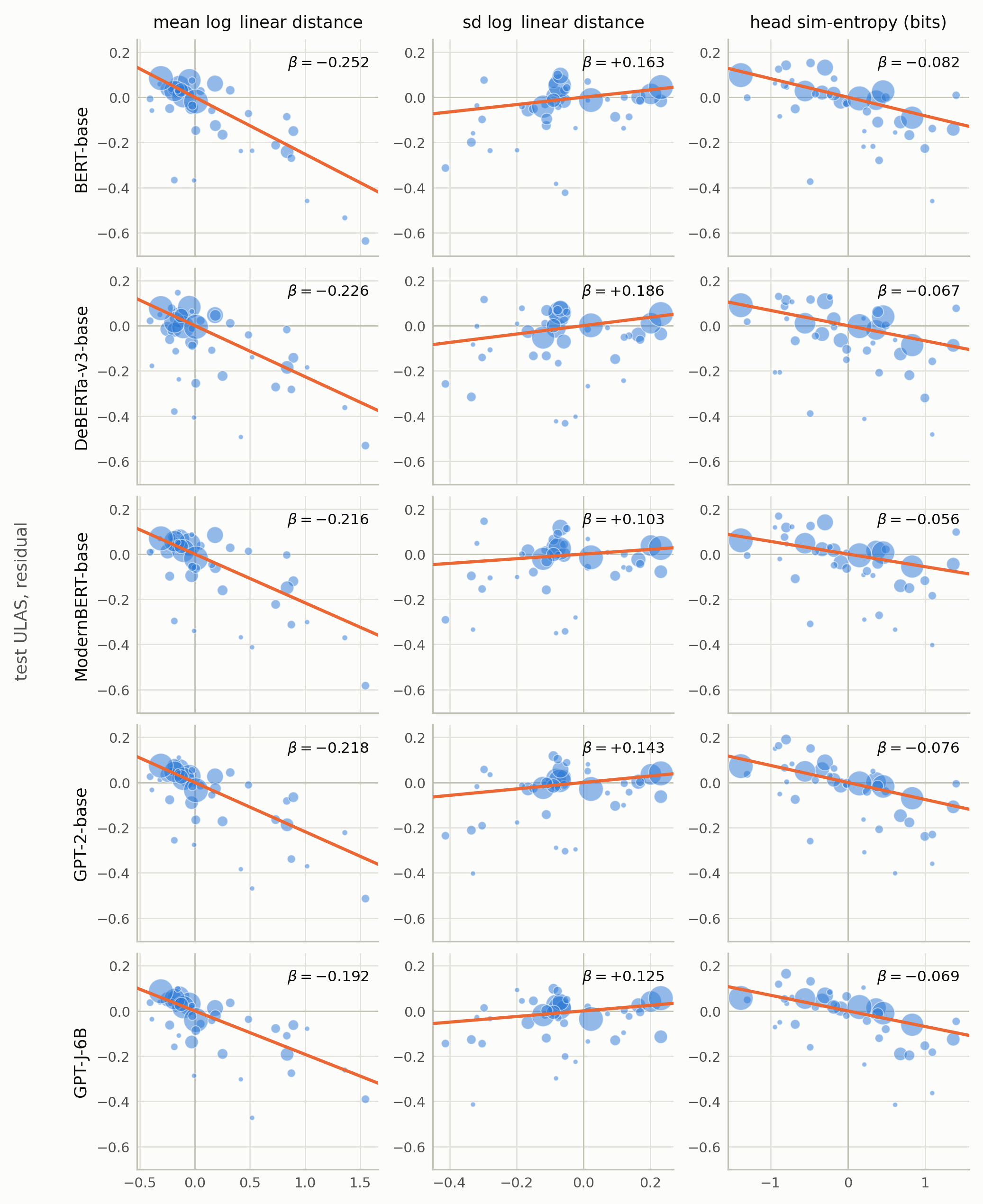}
\caption{Added-variable plots of UASL against each regression predictor, one row per model.  Point area is proportional to the relation's weight in the regression.}
\label{fig:predictor-grid}
\end{figure*}

We report the results in Table \ref{tab:other-models} where we find BERT-base to be the best model for syntactic tree reconstruction, beating out the far-larger GPT-J model by 3 UUAS points. More importantly, however, we observe that all models' regression coefficients display the same sign, and that all models have roughly the same magnitude and significance, indicating that our UASL analysis is not unique to BERT alone.

Figure \ref{fig:predictor-grid} shows the UASL with each of the three predictors of \S\ref{EntropySec}, one row per model. The predictors are computed once on the Penn Treebank, hence they remain the same for every model, so the rows only differ in the UASL values, and the locations of the same 42 relations just move vertically from row to row.

Note that the panels in this figure are added-variable plots: both UASL and the predictor shown are residualised on the other two predictors, so the weighted slope through each cloud is that predictor's coefficient in the full regression and the annotated $\beta$ values reproduce Table \ref{tab:other-models}. The partial view is necessary in the middle column, where $\mathrm{sd}(\log \delta)$ correlates $+0.56$ with $\mathrm{mean}(\log \delta)$: the raw scatter of UASL against dispersion slopes downward ($r$ between $-0.24$ and $-0.37$) although the partial effect is positive in every model.

Across the different models, the three effects are remarkably stable: the slope on the mean distance is consistently negative, that on the dispersion positive everywhere, and that on head similarity-corrected entropy is negative everywhere. Neither sign nor approximate magnitude depends on the architecture.

\subsection{Encoding linear distance in the regression}
\label{app:moments}

In \S\ref{EntropySec} we made three choices in analyzing the linear distance of a syntactic relation in terms of the mean and the standard deviation of the logarithm: we used the mean of the log rather than the log of the mean, the standard deviation instead of the variance, and we stopped at the second moment of the log-length distribution. We discuss these choices in more detail in this appendix, by testing them on the five different models, using the same data and weights.

\paragraph{Mean of the log.} If the UASL behaves like $a + b\log \delta$, then the mean of UASL is $a + b\,\mathrm{mean}(\log \delta)$, so this choice matches the model. We see empirically that replacing $\mathrm{mean}(\log \delta)$ with $\log(\mathrm{mean}\ \delta)$ costs $\sim 0.09$ in $R^{2}$ in all the models we are considering (BERT-base $0.694 \to 0.603$, DeBERTa-v3-base $0.546 \to 0.458$, ModernBERT-base $0.676 \to 0.591$, GPT-2-base $0.701 \to 0.609$, and GPT-J-6B $0.706 \to 0.610$), with $\Delta\mathrm{AIC}$ between $+7.4$ and $+11.8$. While in general $\log(\mathrm{mean}\ \delta)\geq \mathrm{mean}(\log \delta)$ (Jensen inequality), they correlate at $r=0.985$ over the 42 syntactic relations, with $\log(\mathrm{mean}\ \delta)$ losing on the high-frequency relations. 

\paragraph{Standard deviation.}
If we replace $\mathrm{sd}(\log \delta)$ with the variance
$\mathrm{var}(\log \delta)$ we see a lowering of $R^{2}$ in all five models (BERT-base $0.736 \to 0.722$; DeBERTa-v3-base $0.609 \to 0.585$; ModernBERT-base $0.700 \to 0.692$; GPT-2-base $0.744 \to 0.728$; GPT-J-6B $0.746 \to 0.733$), which results in worsening the AIC to $2.6$, with a drop in the statistical significance of the dispersion term ($p$ between $.056$ and $.179$, compared to $.017$ to $.095$ in the case of the standard deviation). This is the reason why we choose the
standard deviation, with its sign interpreted as in \S\ref{EntropySec}.

\paragraph{Second moment.} By considering the mean of log and the standard deviation, we have included the first and second moment of the log-length distribution. We checked that the third moment does not add any further useful information. Table \ref{tab:moment-ladder} shows the result of considering the three moments
\begin{align*}
M_1:\ & \mathrm{UASL} \sim H^{\mathrm{sim}}_{\mathrm{head}} + \mathrm{mean}(\log \delta) \\
M_2:\ & \quad \cdots + \mathrm{sd}(\log \delta) \\
M_3:\ & \quad \cdots + \mathrm{skew}(\log \delta)
\end{align*}
One sees that the dispersion term $M_2$ is statistically significant in four out of five models and improves the adjusted value of $R^{2}$ in all models. On the other hand, the skewness term $M_3$ is in every model not statistically significant, with the adjusted $R^2$ dropping in the case of
ModernBERT-base and GPT-J-6B, and the AIC worsening in three of the models. Adding the third moment also spoils the effect of the dispersion term, since $\mathrm{sd}$ and $\mathrm{skew}$ correlate at $-0.57$.

\begin{table}[t]
\centering

\begin{tabular}{lrrr}
\toprule
\textbf{Model} & \textbf{$M_1$} & \textbf{$M_2$} & \textbf{$M_3$} \\
\midrule
BERT-base       & 0.678 & \textbf{0.715} & 0.719 \\
DeBERTa-v3-base & 0.523 & \textbf{0.579} & 0.605 \\
ModernBERT-base & 0.660 & \textbf{0.676} & 0.668 \\
GPT-2-base      & 0.686 & \textbf{0.723} & 0.724 \\
GPT-J-6B        & 0.691 & \textbf{0.726} & 0.719 \\
\midrule
\multicolumn{4}{l}{\emph{added term, $p$-value}} \\
BERT-base       & --- & .019 & .223 \\
DeBERTa-v3-base & --- & .017 & .069 \\
ModernBERT-base & --- & .095 & .710 \\
GPT-2-base      & --- & .017 & .288 \\
GPT-J-6B        & --- & .019 & .751 \\
\bottomrule
\end{tabular}
\caption{Adjusted $R^{2}$ for the three moments, and $p$-value of the term added at each step (nested $F$-test): adding $M_2$ helps while adding $M_3$ does not.
}\label{tab:moment-ladder}
\end{table}

\clearpage
\section{A transformer pre-trained on permuted sentences}\label{sec:shuffle-model}

\subsection{Structural probes applied to a transformer pre-trained on permuted sentences}
\label{sec:permutate}

High-performing dependency relations often seem to connect two words in very close proximity to each other that, moreover, occur in a specified order (e.g. ``\textbf{as} \underline{well} as''). This becomes more evident if we consider structural probes trained on latent representations of a transformer that has been pre-trained with permuted sentences: RoBERTa-Shuffle-n1, the RoBERTa architecture \cite{liu-etal-2019-roberta} pre-trained by \citet{Sinha2021} on sentences whose word order has been randomised, and which is therefore sensitive only to sentence-wise distributional information. For this permuted model, both distance and order are affected. 

\begin{figure}[b!]
\includegraphics[width=1.07\columnwidth]{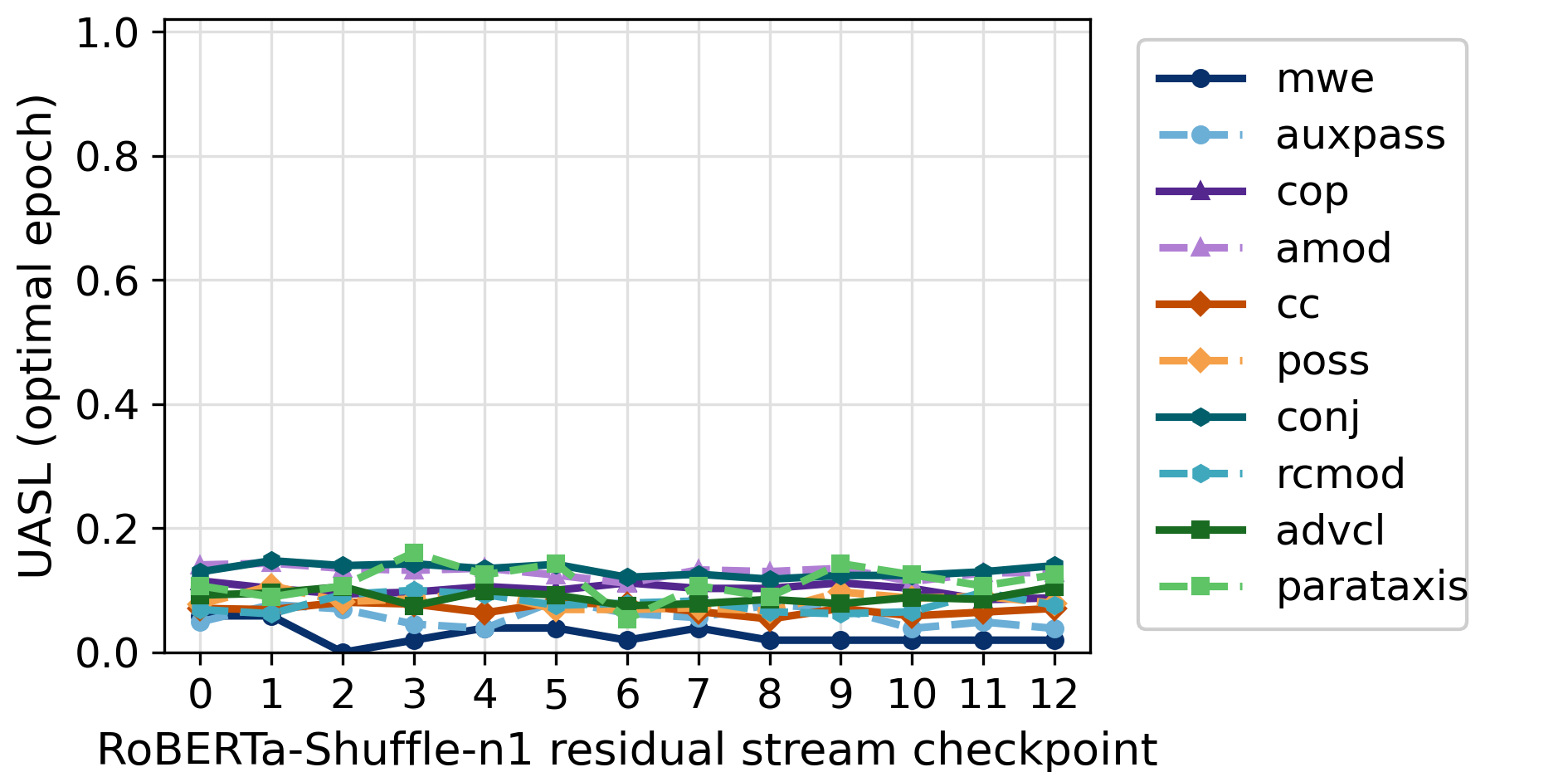}
  \caption{Dev set UASL curves of rank-64 structural probes trained on RoBERTa-Shuffle-n1's latent representations. Relations are those of Figure \ref{fig:dependencies2}. Residual stream checkpoints numbered as in \citet{hewitt2019structural}.}
  \label{fig:dependencies-shuffled}
\end{figure}

In the permuted case, as one would expect, there is a considerable drop in performance across all dependencies. However, we also find an inverted order in the resulting level of performance across different dependencies. For instance, the previously highest scoring {\em mwe} (e.g. ``\textbf{up} \underline{to} 40 cookies'') is among the lowest scoring, perhaps unsurprisingly given that it is generally a set phrase with a fixed {\em ordered} combination of {\em adjacent} words.

The high performer {\em auxpass} is among the lowest in 
the permuted case, showing that ordering and proximity 
between auxiliary and past participle ending is
crucial. After permutation, {\em aux} and {\em auxpass}
become comparable. An inversion of performance also occurs for {\em cc} (between the coordinated head and the conjunction) and
{\em conj} (between the coordinated head and the coordinated dependent). Notably, though {\em conj} performs worse than {\em cc} on BERT-base, it ranks above {\em cc} on the permuted model. Whereas {\em cc} relies on the coordinated head occurring directly before the dependent conjunction, {\em conj} shows an additional symmetry which might be preserved through some permutations.

\begin{figure}
\centering
\includegraphics[width=0.92\columnwidth]{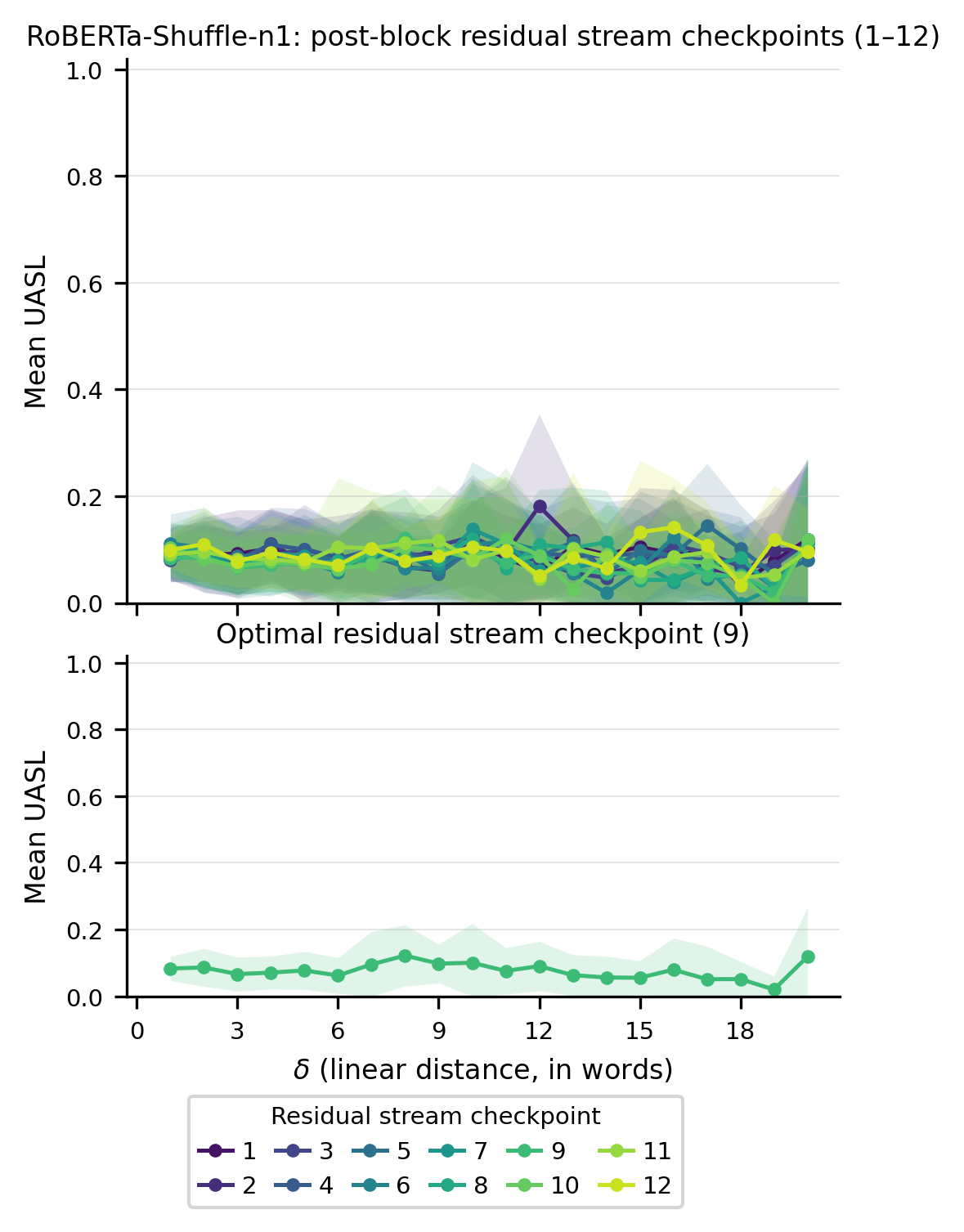}
\caption{Mean UASL as a function of linear distance for RoBERTa-Shuffle-n1, over all 12 post-block residual stream checkpoints (upper panel) and at the best one alone (lower panel). The shaded bands represent one standard deviation across relations. The curve is nearly flat: linear distance carries almost no information. %
}
\label{fig:app-shuffled-curves}
\end{figure}

\begin{figure}
\centering
\includegraphics[width=\columnwidth]{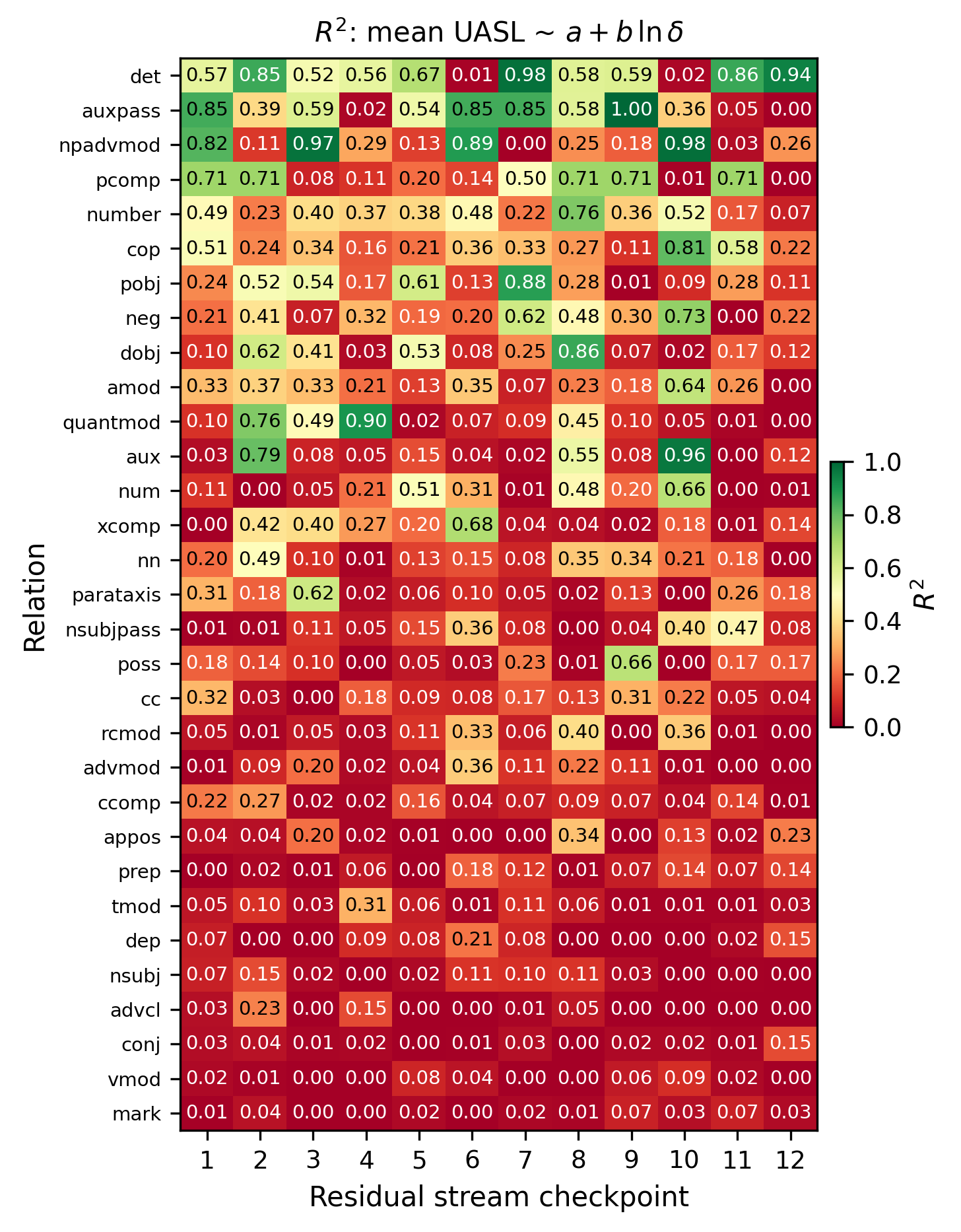}
\caption{Goodness of fit for the log-linear decay model for RoBERTa-Shuffle-n1. Columns are residual stream checkpoints $1$--$12$, after each transformer block and rows are syntactic relations ordered by mean $R^{2}$.  The median $R^{2}$ is $0.102$, against $0.777$ for BERT-base in Figure \ref{fig:R2-all-relations}.}
\label{fig:app-shuffled-r2}
\end{figure}

\subsection{Linear distance and lexical diversity for a model pre-trained on permuted sentences}
\label{app:shuffled}

We have shown in \S\ref{sec:logdistsec} and \S\ref{EntropySec},
using BERT-base, that the UASLs of syntactic relations decay 
approximately log-linearly in linear distance, and that the UASL performance can be predicted in terms of linear distance and lexical similarity-aware entropy of the head of the syntactic relation. We considered RoBERTa-Shuffle-n1 \citep{Sinha2021}, the model discussed in \S\ref{sec:permutate}, pre-trained on permuted sentences (maintaining the same probes, alignments, corpus splits, and predictors). The two properties mentioned are no longer detectable in this permuted case, indicating that they are neither artefacts of the UASL computation method nor properties intrinsic to the treebank data.

\paragraph{Overall performance.} Dev-set UUAS peaks at $0.093$ at checkpoint 9 of the residual stream, against $0.815$ for BERT-base, with very little variation (from $0.084$ to $0.093$) across the 13 checkpoints. Thus, the permuted model does not, at any depth, recover dependency trees, confirming what we see in the per-relation curves of Figure \ref{fig:dependencies-shuffled}.

\paragraph{Log-linear decay.} Figure \ref{fig:app-shuffled-r2} recomputes the goodness-of-fit heat map of \S\ref{sec:logdistsec}. The median $R^{2}$ over (relation, checkpoint) pairs is now $0.102$, compared with $0.777$ for BERT-base, and with $0.588$ for the weakest of the five models of Appendix \ref{app:other-models}. Figure \ref{fig:app-shuffled-curves} directly shows that the mean UASL-against-distance curves are close to flat in the permuted case.

\paragraph{The regression.} Table \ref{tab:shuffled-regression} evaluates the same regression of \S\ref{EntropySec} in the case of RoBERTa-Shuffle-n1. The three predictors account for $R^{2} = 0.195$ of the between-relation variance, against $0.736$ for BERT-base. The coefficient of the mean $\log$ linear distance falls from $-0.252$ to $-0.002$, and those of similarity-aware entropy and dispersion change sign (from $-0.082$ to $+0.014$ and from $+0.163$ to $-0.060$). The syntactic interpretation identified in \S\ref{EntropySec} is not detectable in a model pre-trained without word order (and without structural ordering, since all permutations are used, not just those reflecting different planarizations of a same abstract syntactic tree).

\begin{table}
\centering

\small
\setlength{\tabcolsep}{4.5pt}
\begin{tabular}{lrrrr}
\toprule
\textbf{Predictor} & \textbf{Coef} & \textbf{SE} & \textbf{p} & \textbf{BERT} \\
\midrule
const                & 0.100 & 0.018 & $<$.001 & 1.045 \\
head sim-entropy     & $+0.014$ & 0.006 & .031 & $-0.082$ \\
mean log dep length  & $-0.002$ & 0.014 & .870 & $-0.252$ \\
sd log dep length    & $-0.060$ & 0.031 & .058 & $+0.163$ \\
\bottomrule
\end{tabular}
\caption{The regression of Table \ref{tab:regression_results} on RoBERTa-Shuffle-n1, evaluated at its best residual stream checkpoint ($R^{2}=0.195$, adjusted $R^{2}=0.132$, $F(3,38)=3.07$, $p=.039$, $n=42$). The final column repeats the BERT-base coefficient for comparison. The diversity and the dispersion coefficients change sign, and the linear distance coefficient is reduced to less than $1\%$ of its magnitude.}
\label{tab:shuffled-regression}
\end{table}

\end{document}